%% file: ntrack.tex
\documentclass[lettersize,journal]{IEEEtran}
\usepackage{adjustbox}
\usepackage{algorithm}
\usepackage{algpseudocode}
\usepackage{amsfonts}
\usepackage{amsmath}
\usepackage{amssymb}
\usepackage{array}
\usepackage{booktabs}
\usepackage{bm}
\usepackage{breakcites}
\usepackage{cite}
\usepackage{color}
\usepackage{comment}
\usepackage{csvsimple}
\usepackage{float}
\usepackage{gensymb}
\usepackage{graphicx}
\usepackage{multirow}
\usepackage{pdfpages}
\usepackage{pgfplots}
\usepackage{textcomp}
\usepackage{tikz}
\usepackage{subcaption}
\usepackage{tabularx}
\usepackage[normalem]{ulem}
\usepackage{wrapfig}
\usepackage{xcolor}

\definecolor{mylime}{RGB}{200,245,100}
\definecolor{rlacolor}{RGB}{210,235,230}
\definecolor{powder}{RGB}{176,223,229}
\definecolor{desert}{RGB}{250,213,165}
\colorlet{spy_col}{green!80!yellow!40}

\usetikzlibrary{spy,shapes.misc, fit}
\usetikzlibrary{positioning}
\newcolumntype{Y}{>{\centering\arraybackslash}X}

\PassOptionsToPackage{hyphens}{url}\usepackage[hidelinks,colorlinks=true,citecolor=blue,linkcolor=blue]{hyperref}

\usepackage{orcidlink}

\graphicspath{{figures/}}
\newcommand\copyrighttext{%
  \footnotesize \textcopyright 2023 IEEE. Personal use of this material is
  permitted. Permission from IEEE must be obtained for all other uses, in any
  current or future media, including reprinting/republishing this material for
  advertising or promotional purposes, creating new collective works, for resale
  or redistribution to servers or lists, or reuse of any copyrighted component of
  this work in other works.}
\newcommand\copyrightnotice{%
\begin{tikzpicture}[remember picture,overlay]
\node[anchor=south,yshift=10pt] at (current page.south) {\fbox{\parbox{\dimexpr\textwidth-\fboxsep-\fboxrule\relax}{\copyrighttext}}};
\end{tikzpicture}%
}

\title{NTrack: A Multiple-Object Tracker and Dataset for\\ 
Infield Cotton Boll Counting}

\author{Md Ahmed Al Muzaddid\orcidlink{0000-0002-7916-714X} and William J. Beksi\orcidlink{0000-0001-5377-2627}, \textit{Member, IEEE}
\thanks{Manuscript received 20 September 2023; accepted 30 November 2023. This article
was recommended for publication by Associate Editor K. Zhu and Editor G.
Fortino upon evaluation of the reviewers' comments. \textit{(Corresponding
author: William J. Beksi).}

The authors are with the Department of Computer Science and Engineering, The
University of Texas at Arlington, Arlington, TX 76019, USA. (e-mail:
mdahmedal.muzaddid@mavs.uta.edu, william.beksi@uta.edu).

Color versions of one or more figures in this article are available at
https://doi.org/10.1109/TASE.2023.3342791.

Digital Object Identifier 10.1109/TASE.2023.3342791
\vspace{4mm}
}
}

\begin{document}

\markboth{IEEE TRANSACTIONS ON AUTOMATION SCIENCE AND ENGINEERING}%
{}

\maketitle
\copyrightnotice
\pagestyle{plain}

\begin{abstract}
In agriculture, automating the accurate tracking of fruits, vegetables, and
fiber is a very tough problem. The issue becomes extremely challenging in
dynamic field environments. Yet, this information is critical for making
day-to-day agricultural decisions, assisting breeding programs, and much more.
To tackle this dilemma, we introduce \textit{NTrack}, a novel multiple object
tracking framework based on the linear relationship between the locations of
\textit{neighboring} tracks. \textit{NTrack} computes dense optical flow and
utilizes particle filtering to guide each tracker. Correspondences between
detections and tracks are found through data association via direct observations
and indirect cues, which are then combined to obtain an updated observation. Our
modular multiple object tracking system is independent of the underlying
detection method, thus allowing for the interchangeable use of any off-the-shelf
object detector. We show the efficacy of our approach on the task of tracking
and counting infield cotton bolls. Experimental results show that our system
exceeds contemporary tracking and cotton boll-based counting methods by a large
margin. Furthermore, we publicly release the \textit{first} annotated cotton
boll video dataset to the research community.
\end{abstract}

\def\abstractname{Note to Practitioners}
\begin{abstract}
This work is motivated by the need to provide highly-accurate estimates of the
total number of cotton bolls across an entire farm. We provide a multiple object
tracking framework for automating the counting process using a dynamic motion
model that can reidentify severely occluded objects. This information is
immensely beneficial to agronomists and breeders. For example, it can allow them
to accelerate the selection of genotypes and identify cotton cultivars that
exhibit tolerance to adverse environmental conditions (e.g., drought, poor soil
quality, etc.) via yield prediction. Since the performance of our tracker is
tied to the accuracy of the object detector, the ability to swap detectors is
important for enhancing the usability of the system. Our dataset structure was
modeled after similar multiple object tracking datasets for the purpose of
increasing adoption among practitioners. To implement our system, it is assumed
that the image/video capture device is of high resolution and data is acquired
under ideal lighting conditions. The framework can be run on any
commercial-off-the-shelf platform (e.g., ground-based robots, unmanned aerial
vehicles, etc.), with sufficient computing and memory resources, using a
vision-based sensor.
\end{abstract}

\begin{IEEEkeywords}
Agricultural automation,
computer vision for automation,
visual tracking.
\end{IEEEkeywords}
\vfill

\section*{Multimedia Material}
The source code, dataset, and documentation associated with this project can be
found at
\href{https://robotic-vision-lab.github.io/ntrack}{https://robotic-vision-lab.github.io/ntrack}.

\section{Introduction}
\label{sec:introduction}
\IEEEPARstart{C}{otton} (Gossypium hirsutum L.) is a vital source of natural
fiber. It accounts for nearly 25\% of total world textile fiber use
\cite{usda-ers-1}. Not only is cotton one of the most important textile fibers
in the world, but it has also become a substantial source of food and feed for
humans and livestock by providing cottonseed oil and hulls \cite{national2023}.
Improving the production of cotton is essential to fulfilling the fiber, food,
and feed requirements of the Earth's increasing population
\cite{reynolds2016physiological}. In addition, cotton is a high-valued crop
that requires a significant number of inputs such as preparing seed beds,
planting, reducing competition from insects and weeds, applying harvest aids,
and harvesting. Therefore, among stakeholders, the high return value and cost
of cotton production provides incentives for embracing technologies to improve
profitability by reducing expenses and boosting yields
\cite{zhou2015precision}.

\subsection{Significance of Counting Cotton Bolls}
Infield cotton boll counting is \textit{key} to predicting fiber yield as well
as providing a better understanding of the physiological and genetic mechanisms
of the crop's growth and development.  Repeated counting can contribute to the
calculation of the growth rate in the flowering and boll maturity stages, and
allow for selecting genotypes that more effectively utilize their energy to
form cotton products. The standard approach to obtain yield information is by
manual field sampling, which is \textit{tedious}, \textit{labor intensive}, and
\textit{expensive}. Constrained by these limitations, sampling is done over a
few separate crops and the measurements are extrapolated over an entire farm.
However, inherent human bias and sparsity in the measurements can result in
\textit{inaccurate} yield estimation. Improving yield is a primary objective in
cotton management projects. Physiological factors and environmental variables
all have an effect on flower and boll retention. Understanding these processes
is \textit{crucial} not only to enhancing cotton yield, but also to advancing
the study of plant phenotyping, i.e., the process of measuring and analyzing
observable plant characteristics.

\begin{figure}
\centering
\begin{tikzpicture}
  \node[inner sep=0pt] (field) at (-1,0)
    {\includegraphics[width=.15\textwidth]{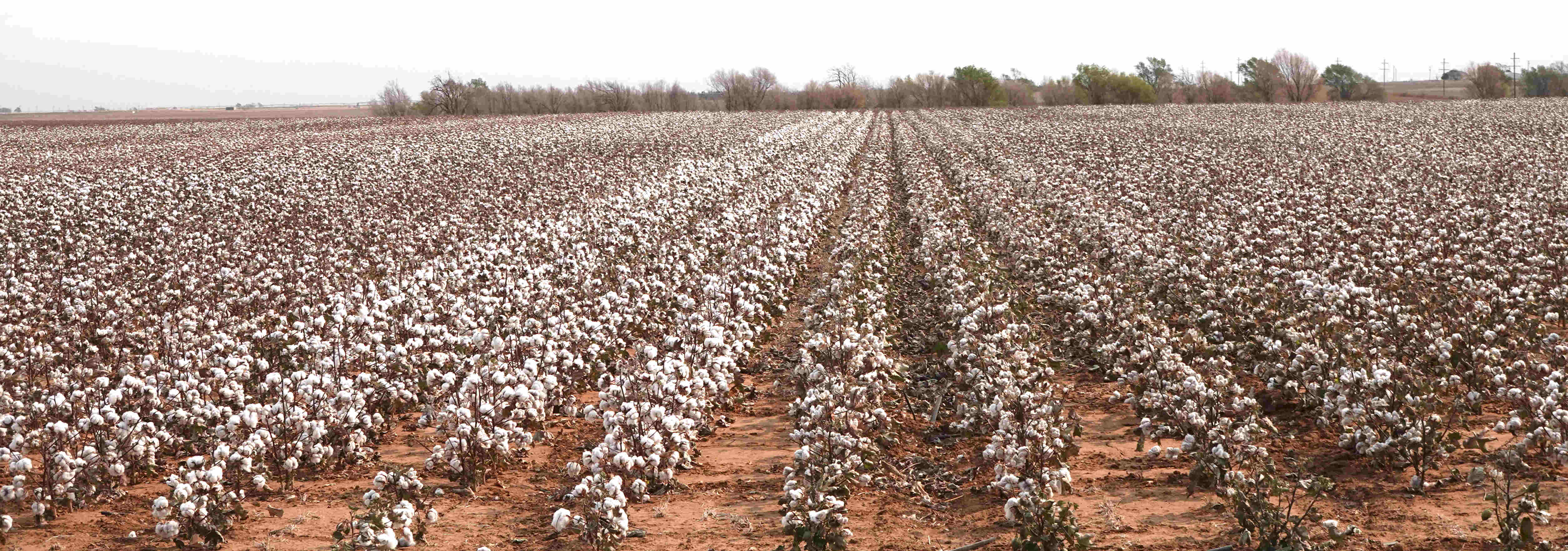}};
  \node[inner sep=0pt] (row) at (2,0)
    {\includegraphics[width=.05\textwidth]{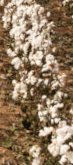}};
  \node[inner sep=0pt] (annotation) at (4.5,-2.5)
    {\includegraphics[width=.10\textwidth]{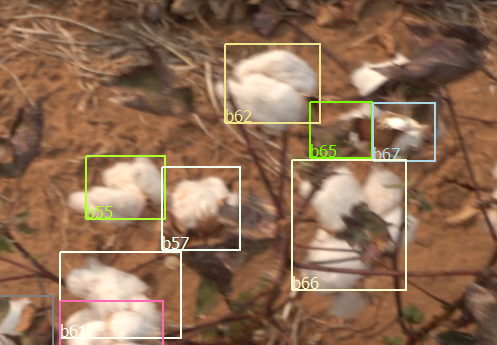}};
  \node (detector) at (4.5,0) [draw,thick,minimum width=1.5cm,minimum height=1.5cm, fill=desert, opacity=.9] {Detector};
  \node (tracker) at (2,-2.5) [draw,thick,minimum width=1.5cm,minimum height=1.5cm,fill=powder] {Tracker};
  \node (count) at (-1,-2.5) [thick, circle] {Count};
    \draw[->,thick] (field.east) -- (row.west) node [midway,above] {a};
    \draw[->,thick] (row.east) -- (detector.west) node [midway,above] {b};
    \draw[->,thick] (row.south) -- (tracker.north) node [midway,right] {e};
    \draw[->,thick] (annotation.west) -- (tracker.east) node [midway,above] {d};
    \draw[->,thick]  (detector.south) -- (annotation.north) node [midway,right] {c};
    \draw[->,thick] (tracker.west) -- (count.east) node [midway,above] {f};
\end{tikzpicture}
\caption{An overview of NTrack, a multiple object tracking system for infield
cotton boll counting. The steps from data collection to cotton boll count
estimation are the following: a - capture video, b - extract video frames, c -
detect cotton bolls, d - input cotton boll detections to the tracker, e - input
video to the tracker, f - output the number of cotton bolls.}
\label{fig:overview}
\end{figure}
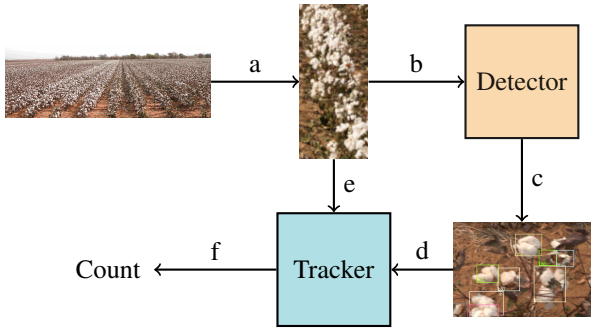

\subsection{Object Detection and Tracking in Agriculture}
The last few decades have seen remarkable advances in vision-based object
tracking. \hspace{1mm} This research has mainly been
\newpage 
\hspace{-4.5mm} applied to tracking pedestrians
\cite{yu2016poi,wojke2017simple,wang2020towards} and vehicles
\cite{koller1994robust,betke2000real}, with benefits to many other important
applications such as surveillance, traffic safety, autonomous driving, and
more.  Nevertheless, vision-based tracking has not evolved proportionally for
the agricultural domain. There is a \textit{dire} need for tracking algorithms
and datasets to support operations in agriculture. For instance, plant
detection and tracking may be utilized to optimize the usage of water,
fertilizer, or other chemicals through variable-rate applications. Accurately
counting the number of leaves, flowers, fruits, etc., is \textit{vital} for
yield prediction and plant phenotyping. Non-destructive, high-throughput data
acquisition platforms (e.g., unmanned aerial vehicles, ground-based robots,
etc.) are enabling experts to analyze more of this data for optimization.

\subsection{Challenges in Detecting and Tracking Cotton Bolls}
Detecting cotton bolls is a \textit{hard} problem and tracking cannot be done if
the object detector fails. Bolls are often clustered together and they may be
split into two or more disjoint regions by branches, foliage, or other
occlusions. In contrast to fruits with rigid shapes (e.g., apples, oranges,
etc.), cotton bolls have complex structures and varied sizes
\cite{mauney1986vegetative}. Hence, the ability to detect and provide a
highly-accurate count of the total number of cotton bolls in a given field is an
\textit{open problem} that we address with our framework,
Fig.~\ref{fig:overview}.

\begin{figure}
\centering
\includegraphics[width=\columnwidth]{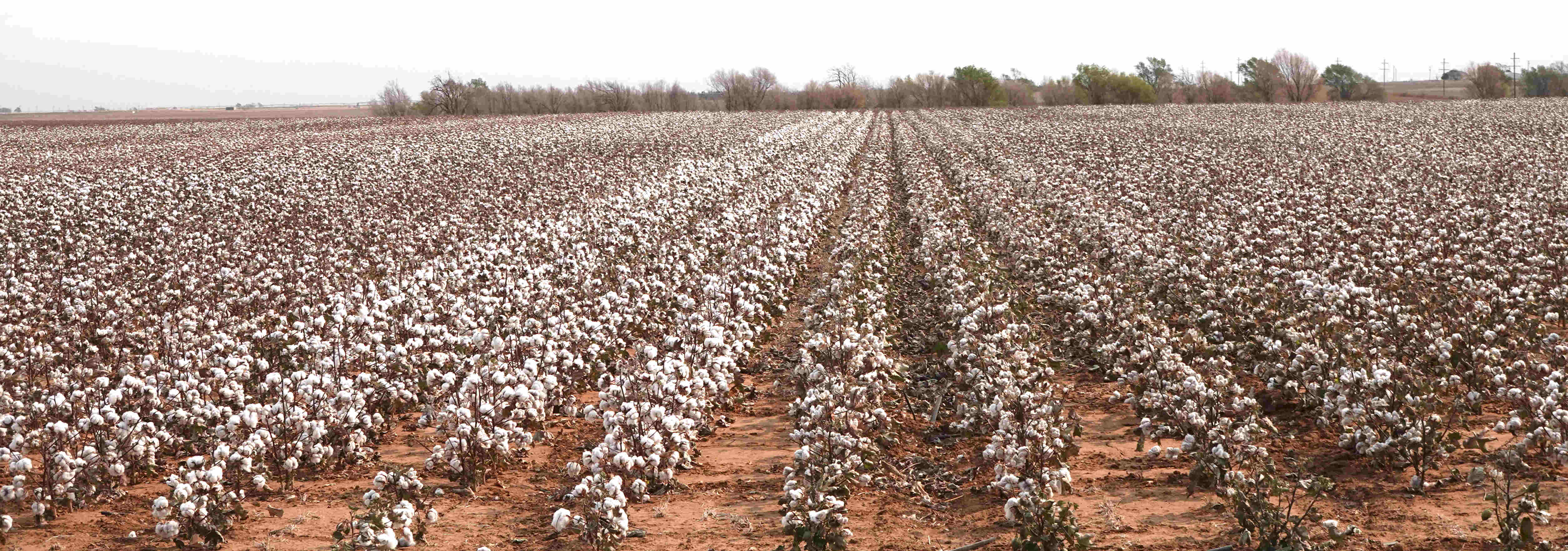}
\caption{The cotton crop research field used for data collection.}
\label{fig:research_field}
\end{figure}

The lack of high-quality datasets is a \textit{significant} roadblock for
meaningful progress in precision plant phenotyping. Annotated video datasets
for cotton crops are \textit{nonexistent}. With the aim of establishing a
robust system for tracking and counting cotton bolls, we created
\textbf{TexCot22} \cite{muzaddid2023texcot22}, an infield cotton boll video
dataset. Each tracking sequence was collected from unique rows of an outdoor
cotton crop research plot located in the High Plains region of Texas,
Fig.~\ref{fig:research_field}. The research plot is comprised of multiple
varieties of cotton. Specifically, each row contains the same cultivar of
cotton while different rows have unrelated cultivars. This results in a diverse
variety of cotton boll shapes and sizes. Additionally, we captured video
sequences from rows treated with contrasting levels of irrigation, which has a
direct impact on the size and the shape of bolls.

In summary, our contributions are the following.
\begin{itemize}
  \item We propose a new tracking system that infers an occluded object's
  location based on the visibility of its neighbors, and allows the tracker to
  maintain identity even under the existence of heavy occlusions.
  \item We identify limitations in appearance-based re-identification
  techniques by outperforming the state of the art in many metrics.
  \item We present the first publicly available infield cotton boll video
  dataset for advancing vision-based research in precision plant phenotyping.
\end{itemize}

The remainder of the paper is structured as follows. We explain the object
detection and tracking problem, and summarize related research in
Section~\ref{sec:related_work}. In Section~\ref{sec:approach}, our detection and
tracking system is presented. The details of our infield cotton boll video
dataset are provided in Section~\ref{sec:cotton_boll_dataset}.
Section~\ref{sec:experiments} outlines and discusses the results of our
experimental evaluation. We conclude in Section~\ref{sec:conclusion} and provide
directions for future work.

\section{Related Work}
\label{sec:related_work}
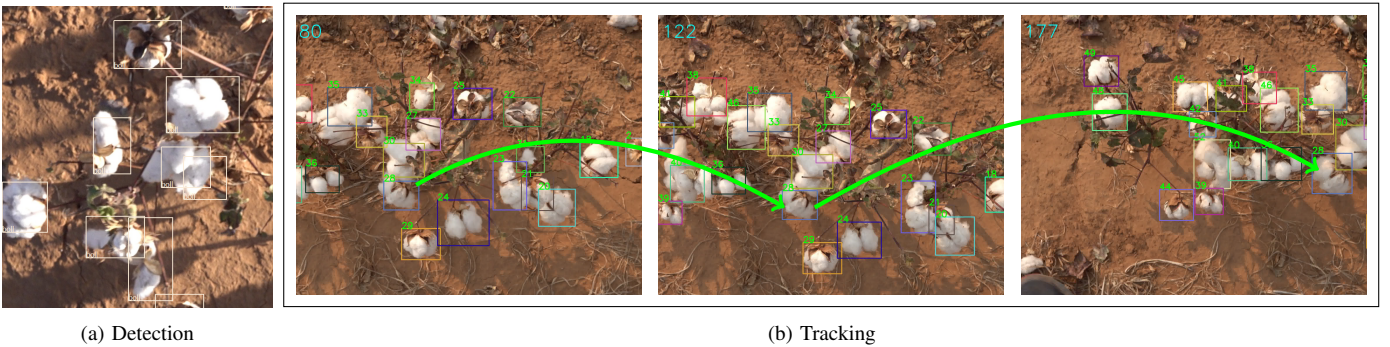
\begin{figure*}
\centering
\begin{subfigure}[b]{0.20\textwidth}
  \centering
  \includegraphics[trim=50 0 100 0, clip, width=.99\textwidth]{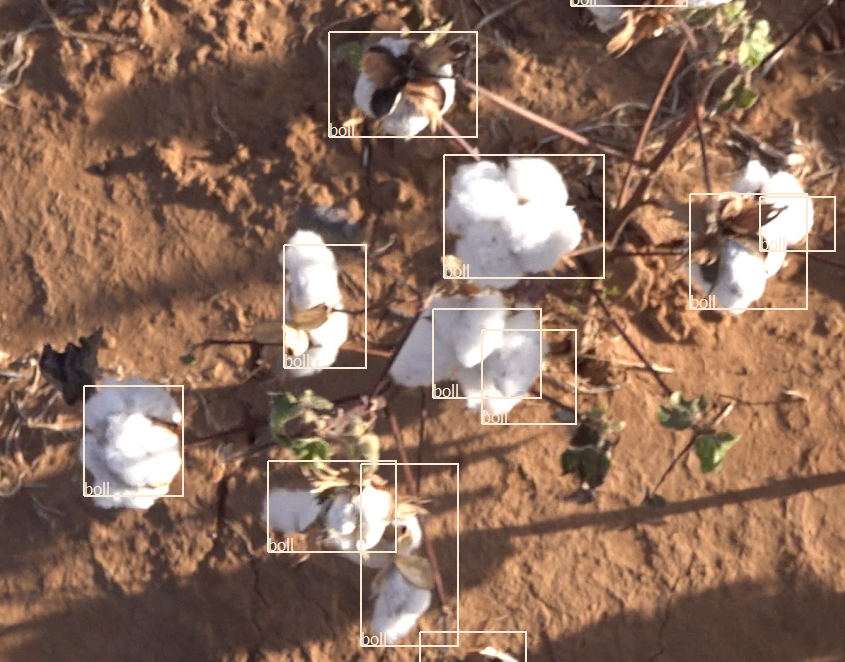}
  \caption{Detection}
\end{subfigure}
\begin{subfigure}[b]{0.79\textwidth}
  \begin{tikzpicture}
    \node (im1)[outer sep=0pt] {\includegraphics[trim=675  200 800 50, clip, width=.32\textwidth]{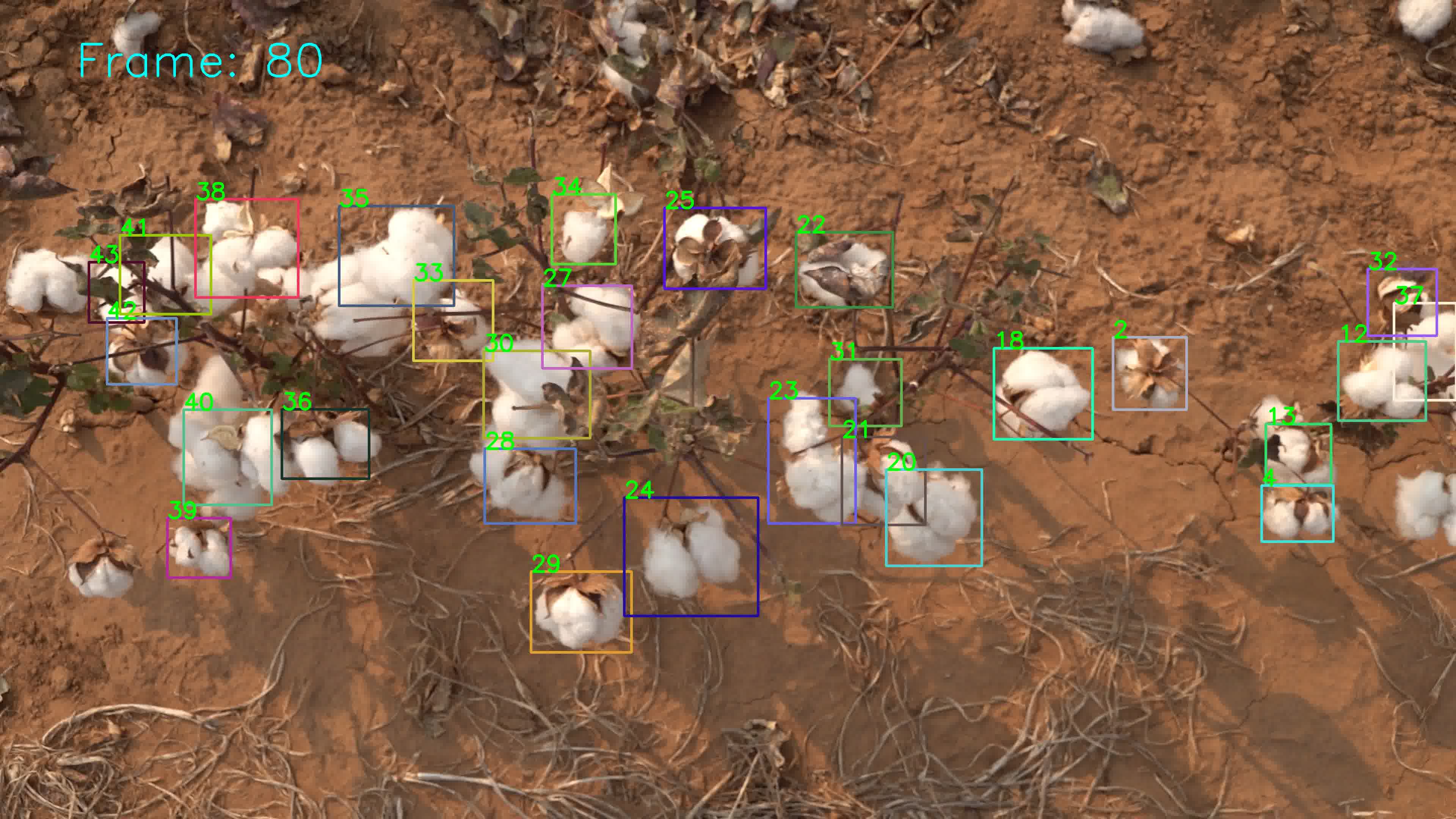}};
    \node (im2)[right = .0cm of im1.east, outer sep=0pt] {\includegraphics[trim=675  200 800 50, clip, width=.32\textwidth]{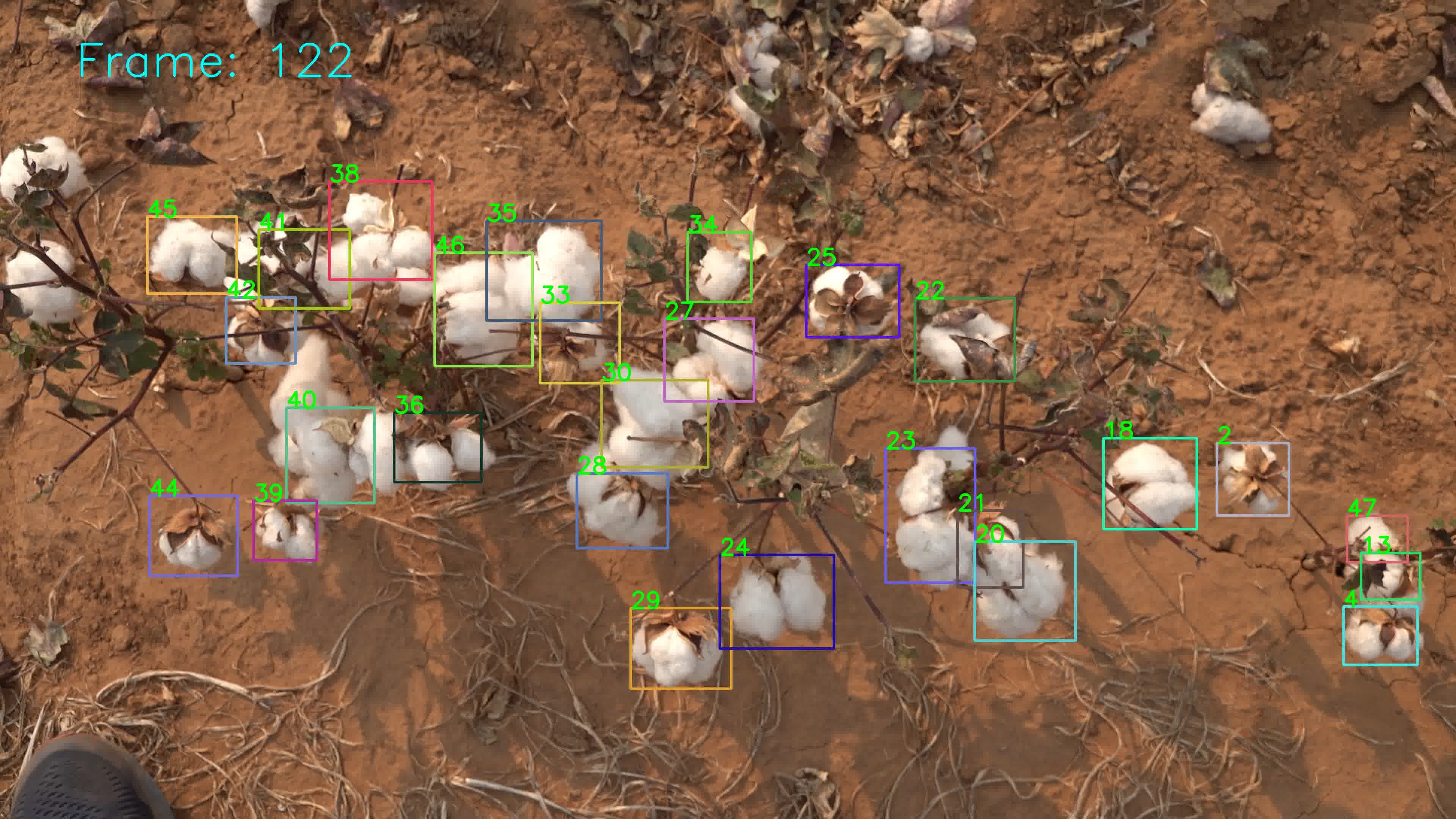}};
    \node (im3) [right = .0cm of im2.east, outer sep=0pt]{\includegraphics[trim=675  200 800 50, clip, width=.32\textwidth]{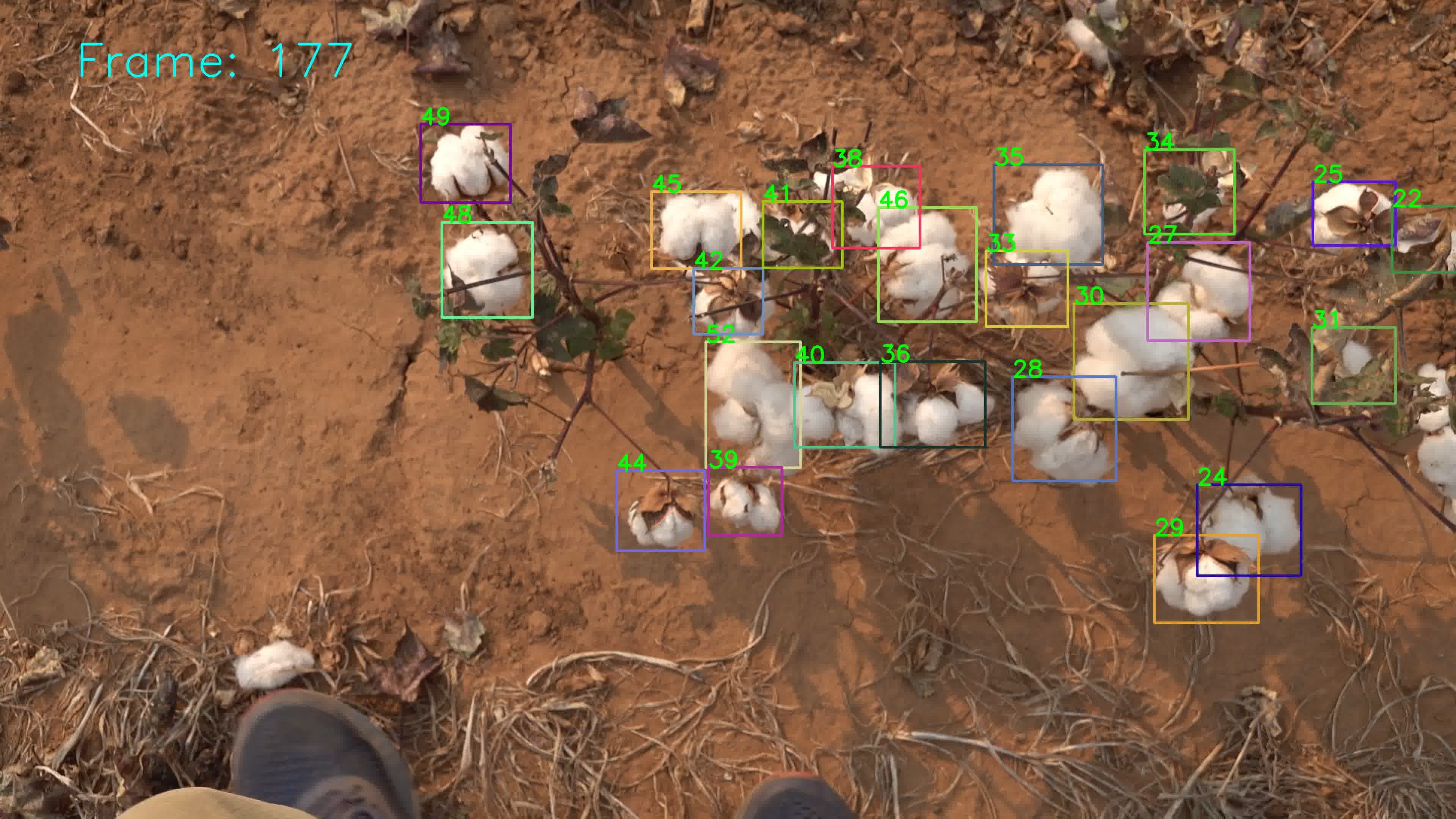}};
    \node[ draw=black, fit=(im1) (im2) (im3) ,inner sep=.5mm](FIt2) {};
    \draw[green,ultra thick,->] (-.7, -.4) to [out=30,in=150] (4.2, -.7);
    \draw[green,ultra thick,->] (4.6, -.7) to [out=30,in=150] (11.3, -0.2);
  \end{tikzpicture}
  \caption{Tracking}
  \label{fig:tracking}
\end{subfigure}
\caption{Multiple object tracking involves (a) detecting the location of
objects and placing bounding boxes around them; and (b) associating a unique
identification to each object, which is then used to track the trajectories of
the objects as they pass through the video sequence.}
\label{fig:detection_vs_tracking}
\end{figure*}

Detecting and tracking objects is one of the most fundamental tasks in computer
vision. Object detection is the process of identifying a target object in an
image or a single video frame. Specifically, it is the task of detecting
\textit{instances} of objects that belong to a certain class. Object tracking
seeks to predict the positions, and other pertinent information, of moving
objects in a video. Single-object tracking involves the following steps: (i)
detect the location of the object, (ii) assign a unique identification to the
object, and (iii) track the object as it moves through the video sequence while
storing the relevant information. In multiple object tracking (MOT)
\cite{luo2021multiple}, given an input video the task is partitioned into
locating multiple objects, maintaining their identities, and yielding their
individual trajectories. The overall process is shown in
Fig.~\ref{fig:detection_vs_tracking}.

\subsection{Multiple Object Tracking}
MOT approaches can be categorized into \textit{online} and \textit{offline}
(batch) tracking. While offline tracking optimizes output based on past and
future observations, online tracking only has access to information up until the
current frame. Offline methods formalize tracking as an association problem and
are solved by using global optimization techniques. For example, early work by
Zhang et al. \cite{zhang2008global} used a probabilistic model to associate
detections with tracks by solving an augmented min-cost flow problem.  A
hierarchical association framework that allows for the integration of affinity
measures or optimization methods was put forth by Huang et al.
\cite{huang2008robust}.  Yang et al. \cite{yang2011learning} attempted to find
the best global associations and transformed the task of finding these
associations into an energy minimization problem.

With limited information, online methods tend to be implemented in a greedy
fashion. Currently, the dominant approach for online tracking follows a
tracking-by-detection paradigm (e.g.,
\cite{wojke2017simple,bergmann2019tracking,braso2020learning,zhang2021fairmot}).
When performing tracking-by-detection, an object detector first localizes all
objects of interest via a set of bounding boxes. Then, the tracking system
associates bounding boxes with preexisting tracks based on motion, appearance,
spatial, or other affinities. As a result, object detection accuracy plays a
decisive role in the final tracking performance. However, many methods only
consider detection boxes whose scores are higher than a set threshold. Objects
with low detection scores (e.g., due to occlusions) are discarded, which
results in sub-optimal results. Trackers such as ByteTrack
\cite{zhang2022bytetrack} try to solve this problem by considering nearly every
detection in their association method.

\subsection{Motion Models}
Various motion models have been used to estimate spatial affinity including the
bounding box intersection over union and Euclidean distance metrics. For
instance, to estimate a pedestrian's velocity, Leal-Taix{\'e} et al.
\cite{leal2014learning} introduced an interaction feature encoded from image
features based on the pedestrian's environment. Bewley et al.
\cite{bewley2016simple} used a Kalman filter-based motion model to predict track
positions while detections are associated via the Hungarian algorithm
\cite{kuhn1955hungarian}.

To overcome the linearity and Gaussian restriction of the Kalman filter, Okuma
et al. \cite{okuma2004boosted} utilized a particle filter for tracking. Li et
al. \cite{li2008tracking} used a cascade particle filter, which consists of
multiple stages of importance sampling to track objects in low frame rate
videos. Optical flow was exploited by Choi \cite{choi2015near} to encode the
relative motion pattern for estimating the likelihood of matching detections.
Yoon et al. \cite{yoon2015bayesian} considered time-varying multiple relative
motion models to represent motion context and facilitate data association.
Although these approaches exploit relative motion, their constant velocity
motion model is not applicable for many scenarios including ours. 

Mauri et al. \cite{mauri2020deep} discussed the use of depth, predicted from
monocular images using self-supervised depth estimation
\cite{godard2019digging}, to estimate vehicular motion. Unfortunately, coarse
depth estimation based on such techniques cannot be used in our situation.
These models are not able to distinguish between two cotton bolls at slightly
different depths. In our approach, we use an optical flow-based dynamic motion
model along with a relative location estimator to update object locations.

\subsection{Appearance Models}
Similar to motion models, appearance models are extensively used in MOT. An
appearance model extracts re-identification features from image regions
corresponding to each bounding box. Appearance-based affinity metrics have
proven to be very informative in the presence of long-occlusion intervals.
Examples include POI \cite{yu2016poi}, DeepSORT \cite{wojke2017simple}, and
Tracktor \cite{bergmann2019tracking}, which extract appearance features using
deep convolutional neural networks (CNNs). Recent transformer-based approaches
(e.g., TrackFormer \cite{meinhardt2022trackformer}) also use implicit
appearance cues to reidentify objects. More traditional methods have been used
too. For instance, region covariance matrices \cite{porikli2006covariance},
color histograms \cite{mitzel2011real}, and gradient-based representations
\cite{choi2012unified}, may be leveraged to find appearance-based similarity.
In contrast to these works, we propose a relative location-based
re-identification technique that can handle significant changes in appearance
due to long-spanning occlusions.

\begin{figure*}
\centering
\scalebox{1.3}{
\hspace{6mm}
\begin{subfigure}{.99\linewidth}
  \input{figures/model.tikz}
\end{subfigure}%
}
\caption{The NTrack pipeline. The \textit{Association} module finds
correspondences between the predicted state of $\mathcal{T}$ tracks (top, green
squares). $\mathcal{D}^{f_l}$ is the set of detection bounding boxes (bottom,
green squares) detected in frame $F_{l}$. The dormant (unmatched) tracks
$\mathcal{T}_d$ (orange squares) are analyzed by the \textit{Relative Location
Analyzer} (RLA) and updated accordingly. In the RLA module, the dormant tracks'
locations (orange circles) are estimated based on the location of their nearest
neighbors (green circles). Unmatched detections $\mathcal{D}_u$ (bottom,
light-green squares) are passed through the \textit{Initialize} module as
candidates for a new track.}
\label{fig:architecture_overview}
\end{figure*}
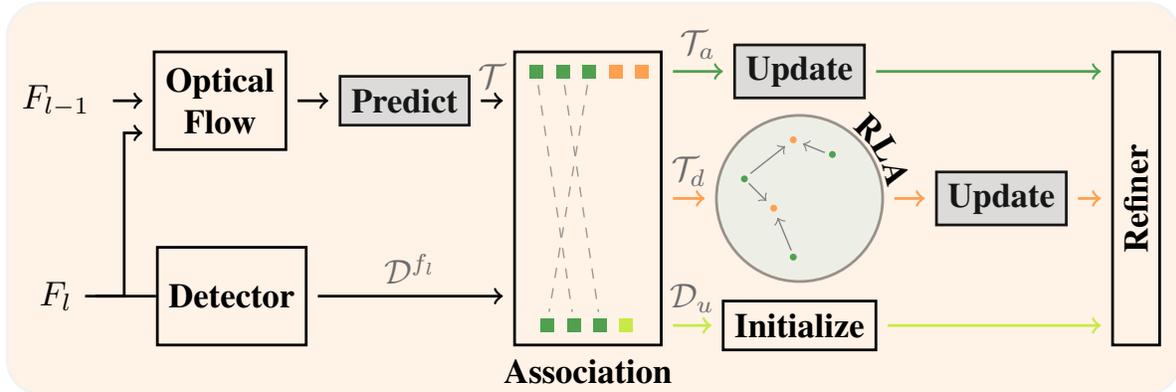

\subsection{Multiple Object Tracking in Agriculture}
Precisely detecting and tracking objects is imperative for automated
fruit/vegetable/fiber counting. In work by Hung et al. \cite{hung2015feature},
apple trees were sampled every 0.5 meters to estimate yield by counting the
apples in each non-overlapping image. Chen et al. \cite{chen2017counting}
implemented a two-stage method to count apples and oranges from images. An
incremental structure from motion technique was proposed by Roy and Isler
\cite{roy2016surveying} to register and localize apples in 3D, which is later
used for extracting the count. H{\"a}ni et al. \cite{hani2018apple} first
identified apple clusters in images based on color, and then used a CNN to
classify the clusters where each class represents the number of apples in the
cluster. Afonso et al. \cite{afonso2020tomato} provided a detection-based
approach to enumerate tomatoes in images captured from a greenhouse environment.
A DeepSORT inspired tracking method to count fruit in a controlled setting was
used by Kirk et al. \cite{kirk2021robust}. All of the aforementioned techniques
tally fruits composed of rigid shapes, which are simpler to separate into
clusters thus making the counting task easier. On the other hand, counting
irregular-shaped cotton bolls under dynamic field conditions is far more
challenging.

\subsection{Cotton Boll Counting}
To acquire information for yield prediction, recent research has developed
algorithms to segment and count cotton bolls. Sun et al. \cite{sun2019image}
introduced a counting method based on the geometric features of cotton bolls.
This was done by applying color and spatial features to segment bolls from the
background, followed by geometric feature-based algorithms to estimate the boll
count. Tedesco-Oliveira et al. \cite{tedesco2020convolutional} used different
deep learning-based object detection models on a cotton crop dataset consisting
of 948 images for training, 236 images for validation, and 205 images for
testing. By comparison, our dataset is \textit{significantly larger} (see
Section~\ref{sec:cotton_boll_dataset}) and we employ a model that is capable of
detecting objects of various sizes. Sun et al. \cite{sun2020three} take a 3D
approach to counting cotton bolls by using point clouds reconstructed from
multi-view images via structure from motion under field conditions.

All of the preceding methods investigate the counting of cotton bolls from
images. Image-based counting requires automated or manual sampling. For
example, Tedesco-Oliveira et al. \cite{tedesco2020convolutional} selected 25
dissimilar images so that there were no overlapping images in each video
sequence. Such sampling-based methods demand additional pre-processing and
post-processing steps.  Conversely, our approach not only eliminates this
overhead without sacrificing accuracy, but it also achieves better accuracy
than previous works. Moreover, we argue that counting from videos has the
potential to locate additional cotton bolls since the counting sequence enables
an automated system to find bolls that may otherwise be occluded in a single
image. To the best of our knowledge, our work is the first to engage in the
cotton boll counting task \textit{directly from infield video sequences}.

\section{Approach}
\label{sec:approach}
The architectural overview of NTrack is depicted in
Fig.~\ref{fig:architecture_overview}. Similar to other tracking-by-detection
systems, the performance of NTrack highly depends upon the detection accuracy.
We assume that the detection bounding boxes in every frame are estimated prior
to tracking. The tracking procedure is independent of the underlying detection
algorithm, which allows NTrack's \textit{Detector} module to use any
state-of-the-art object detection method to identify cotton bolls. 

Multiple tracks, each linked to a unique cotton boll and steered/guided by a
particle filter, are responsible for estimating the location of an associated
boll. The job of the \textit{Association} module is to find the correspondences
between detections and tracks based on a matching criteria.  Given a track's
past trajectory along with its neighbors' trajectories, the \textit{Relative
Location Analyzer} (RLA) module can estimate the current location of the track.
A track is suppressed by the \textit{Refiner} module if the track's location
goes out of the frame border or it cannot be matched against a detection in
recent frames.

We denote a set of detections by $\mathcal{D} = \{r_i\}$, where $r_i$ is the
detection response represented as the tuple $(x_i, y_i, s_i, w_i, f_i)$. Within
the tuple, $(x_i,y_i)$ is the center, $s_i$ is the size, and $w_i$ is the width
of the detected bounding box. The frame/time at which an object is detected is
defined by $f_i$. Note that we use $f_i$ to denote both the frame and time
interchangeably. $\mathcal{D}^{f_l} \subset \mathcal{D}$ is the set of bounding
boxes detected in frame $f_l$. We define a track $\tau_k \in \mathcal{T}$ over
multiples frames by a set of bounding boxes associated with a particular
object, where $\mathcal{T}$ represents the set of all tracks in the system.
Ideally, each object is associated with a single track. In reality, maintaining
a perfect one-to-one mapping between an object and a track is impossible due to
occlusions, misdetections, and ambiguous associations. Therefore, a track is
\textit{active} in a given frame if the system can successfully associate the
track's predicted bounding box with a detected bounding box, otherwise it is
\textit{dormant}.


For each new frame $F_l$, dense optical flow is computed by the \textit{Optical
Flow} module with respect to the previous frame $F_{l-1}$ based on the
Gunnar-Farneback algorithm \cite{farneback2003two}. Optical flow vectors are
outputted for each pixel of the frame. Next, informed by these flow vectors,
all tracks predict their new location in frame $F_l$ via the \textit{Predict}
module. Each track employs a particle filter for the prediction and update
step. The particle filters track the bounding box locations (centers).
Nonetheless, since the camera motion and movement of the objects are irregular,
neither a constant velocity nor constant acceleration model-based state
estimation performs as expected. Hence, we apply a dynamic flow velocity model
for the bounding box location estimation. Unlike location, the scale and width
of the bounding box changes gradually and is therefore tracked by a Kalman
filter using a constant velocity model.

With the detection bounding boxes provided by the tracks, the
\textit{Association} module attempts to make a unique correspondence between
two sets of bounding boxes. It produces the following outputs: (i) a set of
active tracks $\mathcal{T}_a$ successfully matched against detections
$\mathcal{D}_m$, (ii) a set of dormant tracks $\mathcal{T}_d =
\mathcal{T}\setminus \mathcal{T}_a$, and (iii) a set of unmatched detections
$\mathcal{D}_u= \mathcal{D}^{f_l} \setminus \mathcal{D}_m$. The particle
filters associated with the active tracks $\mathcal{T}_a$ update their state
using the information from matched detections $\mathcal{D}_{m}$ through the
Association module's \textit{Update} module. The dormant track
$\mathcal{T}_d$'s states are analyzed by the RLA and updated via its
\textit{Update} module. If any of the remaining unmatched detections
$\mathcal{D}_u$ exceeded a detection confidence threshold, then they are
initialized as new tracks (\textit{Initialize} module). Lastly, the
\textit{Refiner} module removes any track that was dormant for the last 100
frames.

\subsection{Object Location Prediction}
Instead of using well-known motion models to predict the next state $(x_i, y_i,
s_i, w_i)$ of the object, we use optical flow-based motion estimation. Due to
outdoor environmental conditions (e.g., wind), the branches of a cotton plant
can sway back and forth. Moreover, the camera movement over the terrain is
erratic, which causes irregular motion dynamics. Even under these adverse
conditions, we are still able to find reliable locations of the tracked objects
in image coordinates through optical flow. In the prediction step, the particles
of the particle filter are moved according to the estimated flow velocities.
Since we have multiple targets per frame, we opt for a robust to noise dense
optical flow algorithm instead of sparse flow, which would require estimating
flow separately for each object.

\subsection{Object Location Update}
During the update phase, a track could be in an active or dormant state based
on the association outputs. Thus, we use one of two different routines to
update a track. If the track is active in frame $f_l$, then we update the
weight of the particles based on direct observation (i.e., the detected object
bounding box from the detector). Otherwise, we use the relative locations with
respect to the neighboring tracks as indirect observations to update the
dormant track. More specifically, the relative locations are used to calculate
the bounding box center, which is then used as a proxy for the direct
observation in order to update the particle filter.

\begin{figure*}
\centering
\begin{subfigure}[b]{0.49\textwidth}
  \begin{tikzpicture}[spy using outlines={chamfered rectangle, magnification=3, width=3cm, height=2cm,green, connect spies}]
    \node {\includegraphics[width=.99\textwidth]{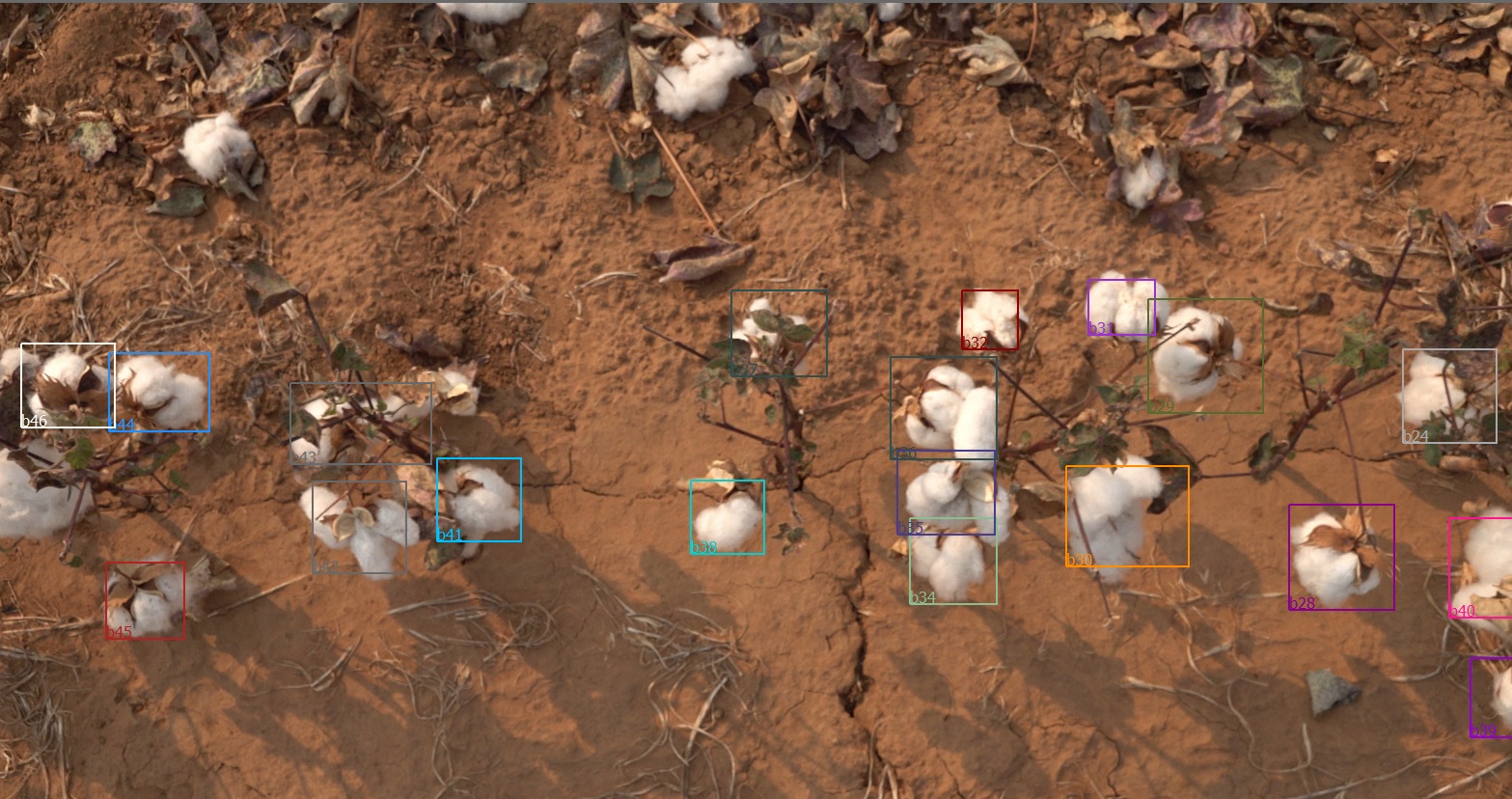}};
    \spy[every spy on node/.append style={ultra thick}] on (-3.7,.09)  in node [right] at  (-1,1.75);
  \end{tikzpicture}
\end{subfigure}
\begin{subfigure}[b]{0.49\textwidth}
  \begin{tikzpicture}[spy using outlines={chamfered rectangle, magnification=3, width=3cm, height=2cm,green, connect spies}]
    \node {\includegraphics[width=.99\textwidth]{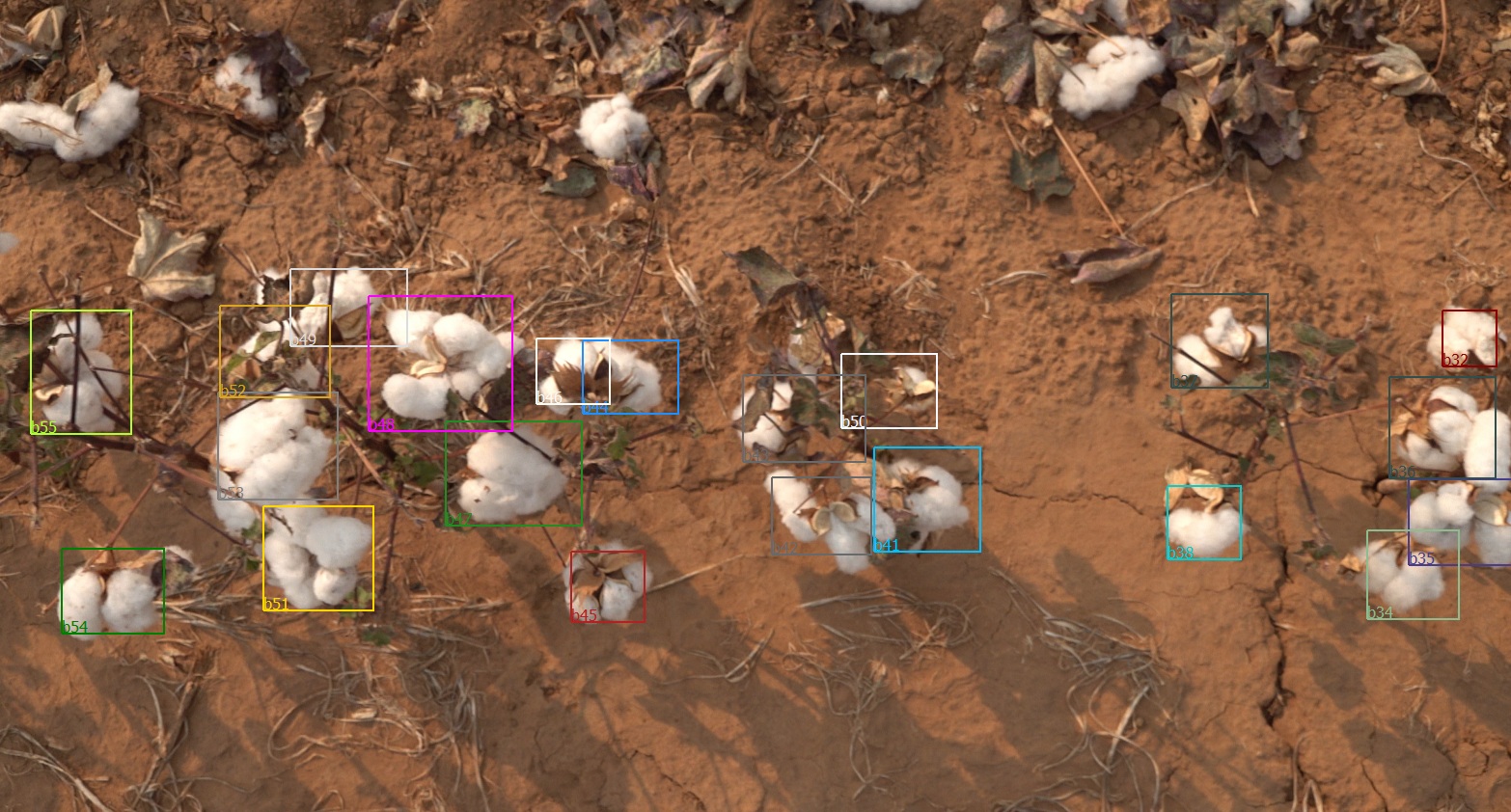}};
    \spy[every spy on node/.append style={ultra thick}] on (-.85,.18) in node [right] at  (-1,1.75);
  \end{tikzpicture}
\end{subfigure}
\begin{subfigure}[b]{0.49\textwidth}
  \begin{tikzpicture}[spy using outlines={chamfered rectangle, magnification=3, width=3cm, height=2cm,green, connect spies}]
    \node {\includegraphics[width=.99\textwidth]{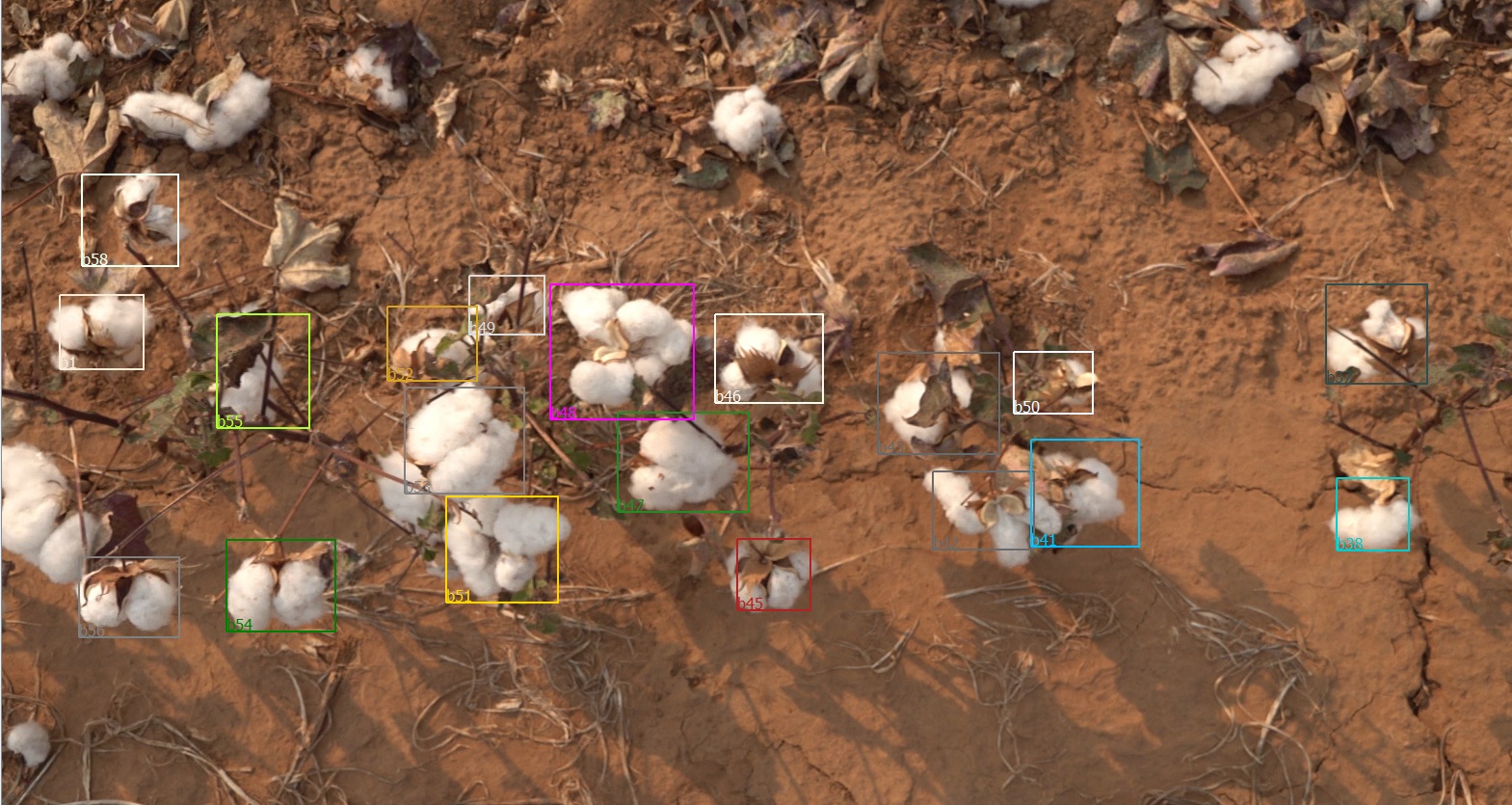}};
    \spy [every spy on node/.append style={ultra thick}] on (0.05,.25) in node [right] at (-1,1.75);
  \end{tikzpicture}
\end{subfigure}
\begin{subfigure}[b]{0.49\textwidth}
\centering
\includegraphics[width=.99\textwidth]{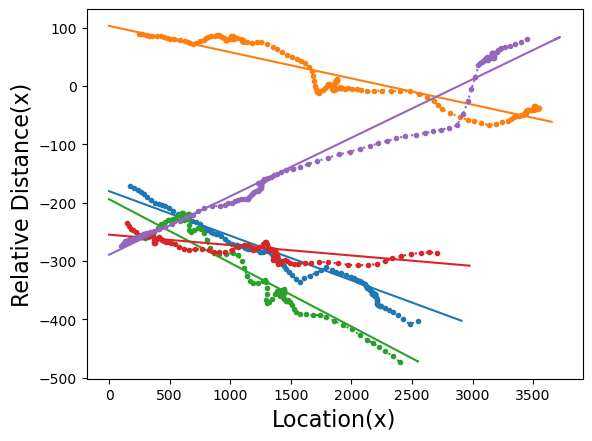}
\end{subfigure}
\caption{The relative distance between neighboring cotton bolls changes
gradually due to the shift in perspective. At frame 1142, the distance between
\textit{b46} and \textit{b44} is large (top left). At frame 1203, the distance
between the bolls reduces (top right). At frame 1212, both bolls overlap, i.e.,
the distance is zero (bottom left). The plot shows the relative distances
between pairs of cotton bolls versus their locations (bottom right). The
distance and location are measured in image coordinates along $x$ (width)
direction. Each dotted line shows the distance between a pair of cotton bolls.
The straight lines are linear fits, for reference, to the respective dotted
lines.}
\label{fig:relative_distance}
\end{figure*}

\subsection{Relative Object Location}
Cotton plants in the field may move unpredictably, which makes the tracking
task much more difficult. However, the locations of the cotton bolls are not
totally independent of each other. In fact, they are highly correlated with
neighboring bolls, Fig.~\ref{fig:relative_distance}.  More formally, we denote
the location $(x, y)$ of a track $\tau_i$ in the image coordinate at time $f_l$
as $\rho^{f_l}_{\tau_i} \in \mathbb{R}^2$. The locations along the image width
($x$) and height ($y$) are independent, and the estimated location based on the
neighbor track $\tau_j$ is defined as $\rho_{\tau_i, \tau_j}^{f_l} \in
\mathbb{R}^2$. Similarly, we define $d_{\tau_i,\tau_j}^{f_l} \in \mathbb{R}^2$
as the relative distance of track $\tau_i$, with respect to neighbor $\tau_j$,
recorded at time $f_l$. For simplicity, in the following derivation we assume
$\rho^{f_l}_{\tau_i}$, $\rho_{\tau_i, \tau_j}^{f_l}$, etc., are scalars and
that they represent only the $x$ component of the location. Based on the
observed correlation (Fig.~\ref{fig:relative_distance}), we assume that the
relative distance between two neighboring tracks changes linearly with the
track's location. Concretely, let $d_{\tau_i,\tau_j}^{f_l} = c_1 \cdot
\rho^{f_l}_{\tau_j} + c_0$ where $c_1$ and $c_0$ are constants. We derive the
linear relationship between the locations of two neighboring tracks as
\begin{align}
   d_{\tau_i,\tau_j}^{f_l} &= c_1 \cdot \rho^{f_l}_{\tau_j} + c_0, \notag\\
   d_{\tau_i,\tau_j}^{f_l} + \rho^{f_l}_{\tau_j} &= c_1 \cdot \rho^{f_l}_{\tau_j} + c_0 + \rho^{f_l}_{\tau_j},\\
   \rho_{\tau_i, \tau_j}^{f_l} &= (c_1 +1) \cdot \rho^{f_l}_{\tau_j} + c_0, \notag
\end{align}
which suggests that the location of neighboring tracks can be a good estimator
of a dormant track's location.

Based on this insight, we calculate track $\tau_i$'s location at time $f_l$
based on a particular neighbor $\tau_j$ as follows.  Let $\rho_{\tau_j}^{f_l}$
and $\rho_{\tau_i, \tau_j}^{f_l}$ be continuous random variables.
The linear relationship between a neighboring track's location implies that we
can assume the mean $E[\rho_{\tau_i, \tau_j}^{f_l}]$ is linear in
$\rho_{\tau_j}^{f_l}$. At the same time, assuming the variance
$var[\rho_{\tau_i, \tau_j}^{f_l}]$ is constant over $f_l$, we can model
$\rho_{\tau_i, \tau_j}^{f_l}$ as a Gaussian random variable
\begin{equation}
 \rho_{\tau_i, \tau_j}^{f_l} \sim \mathcal{N}(\mu_{\tau_i, \tau_j}^{f_l}, \sigma^2_{\tau_i,\tau_j}),
\end{equation}
where
\begin{align}
\begin{pmatrix}
  \rho_{\tau_i, \tau_j}^{f_1} \\ \rho_{\tau_i, \tau_j}^{f_2} \\ \vdots \\ \rho_{\tau_i, \tau_j}^{f_m}
\end{pmatrix}
  &=
\begin{pmatrix}
  1 & \rho^{f_1}_{\tau_j} \\
  1 & \rho^{f_2}_{\tau_j} \\
  \vdots  & \vdots \\
  1 & \rho^{f_m}_{\tau_j}
\end{pmatrix}
\begin{pmatrix}
  c_0 \\ c_1
\end{pmatrix}
  +
\begin{pmatrix}
  e_1 \\ e_2 \\ \vdots \\ e_m
\end{pmatrix},
\label{eq:least_squares}\\
\mu_{\tau_i, \tau_j}^{f_l} &= E[\rho_{\tau_i, \tau_j}^{f_l}] = [1 \;\rho^{f_l}_{\tau_j}] \cdot
[c_0 \; c_1]^\top,\\
\sigma^2_{\tau_i, \tau_j} &= var[\rho_{\tau_i, \tau_j}^{f_l}] = \frac{\sum_{i}^{}e_i^2}{m}.
\end{align}
We assume that there are $m$ frames prior to frame $f_l$ in which tracks
$\tau_i$ and $\tau_j$ were active simultaneously and the relative distances
between them were recorded. $c_0$ and $c_1$ are the least-squares solutions to
\eqref{eq:least_squares}. Finally, the estimated locations
$\rho_{\tau_i,\tau_j}^{f_l}\,|\,\tau_j\in kn(\tau_i)$ based on $k$ neighbors
are combined as follows (for simplicity we drop the superscript $f_l$):
\begin{align}
  \label{eqn:product_of_gaussians}
  \rho_{\tau_i} &\sim \prod_{\scriptscriptstyle \tau_j \in kn(\tau_i)} \mathcal{N}(\mu_{\tau_i,\tau_j}, \sigma^2_{\tau_i,\tau_j}),\\
  \label{eqn:simplified_single_gaussian}
  &\sim \beta \mathcal{N}(\mu_{\tau_i},\sigma^2_{\tau_i}),
\end{align}
where $\beta$ is a scaling factor and
\begin{align}
  \sigma_{\tau_i} &= \left(  \sum_{\tau_j \in kn(\tau_i) }^{} \sigma^{-2}_{\tau_i, \tau_j}\right)^{-1/2},\\
  \mu_{\tau_i}    &= \sigma^2_{\tau_i}  \sum_{\tau_j \in kn(\tau_i)}^{}  \sigma^{-2}_{\tau_i, \tau_j} \mu_{\tau_i,\tau_j}.
  \label{eqn:final_meu_sigma}
\end{align}
In \eqref{eqn:product_of_gaussians}, $kn(\tau_i)$ is the set of $\tau_i$'s
neighbors with cardinality $k$, which is inspired by the Kalman filter's
approach to combine the prior with an observation for an improved estimate.
The simplification from \eqref{eqn:product_of_gaussians} to
\eqref{eqn:simplified_single_gaussian} is due to \cite{bromiley2003products}.

The neighbors $kn(\cdot)$ are selected based on the $k$-nearest neighbors
algorithm. We measure the distances between the bounding box centers using
their Euclidean norm. For a particular track $\tau_i$, we can end up with more
than $k$ neighbor locations in total, even if we record just $k$ neighboring
locations in each frame. In addition, we want to prioritize a neighbor that was
simultaneously active with $\tau_i$ in more frames in the recent past.
Therefore, to choose $k$ neighbors among all the recorded neighbors we rank
each neighbor,
\begin{equation}
  R^{f_l}_{\tau_i}(\tau_j) = \sum_{f_m\in \mathcal{A}^{f_l}_{\tau_i,\tau_j}}f_m,
\end{equation}
where $\mathcal{A}^{f_l}_{\tau_i,\tau_j}$ is the set of timestamps prior to
$f_l$ in which tracks $\tau_i$ and $\tau_j$ were active simultaneously. For
example, suppose tracks $\tau_i$ and $\tau_{n_1}$ are simultaneously active at
times 2, 3, and 5. Then, $\mathcal{A}^{f_l}_{\tau_i,\tau_{n_1}} = \{2,3,5\}$ and
similarly $\mathcal{A}^{f_l}_{\tau_i,\tau_{n_2}} = \{5,6\}$. In this case, we
prioritize the neighbor $n_2$ according to the rank, which is higher than $n_1$
since $R^{f_l}_{\tau_i}(\tau_{n_2})= 5+6 \; > \; R^{f_l}_{\tau_i}(\tau_{n_1}) =
2+3+5$.

\section{Cotton Boll Dataset}
\label{sec:cotton_boll_dataset}
\subsection{Video Acquisition Details}
\begin{figure*}
\centering
\subfloat[]{
  \includegraphics[width=0.47\columnwidth,height=33mm]{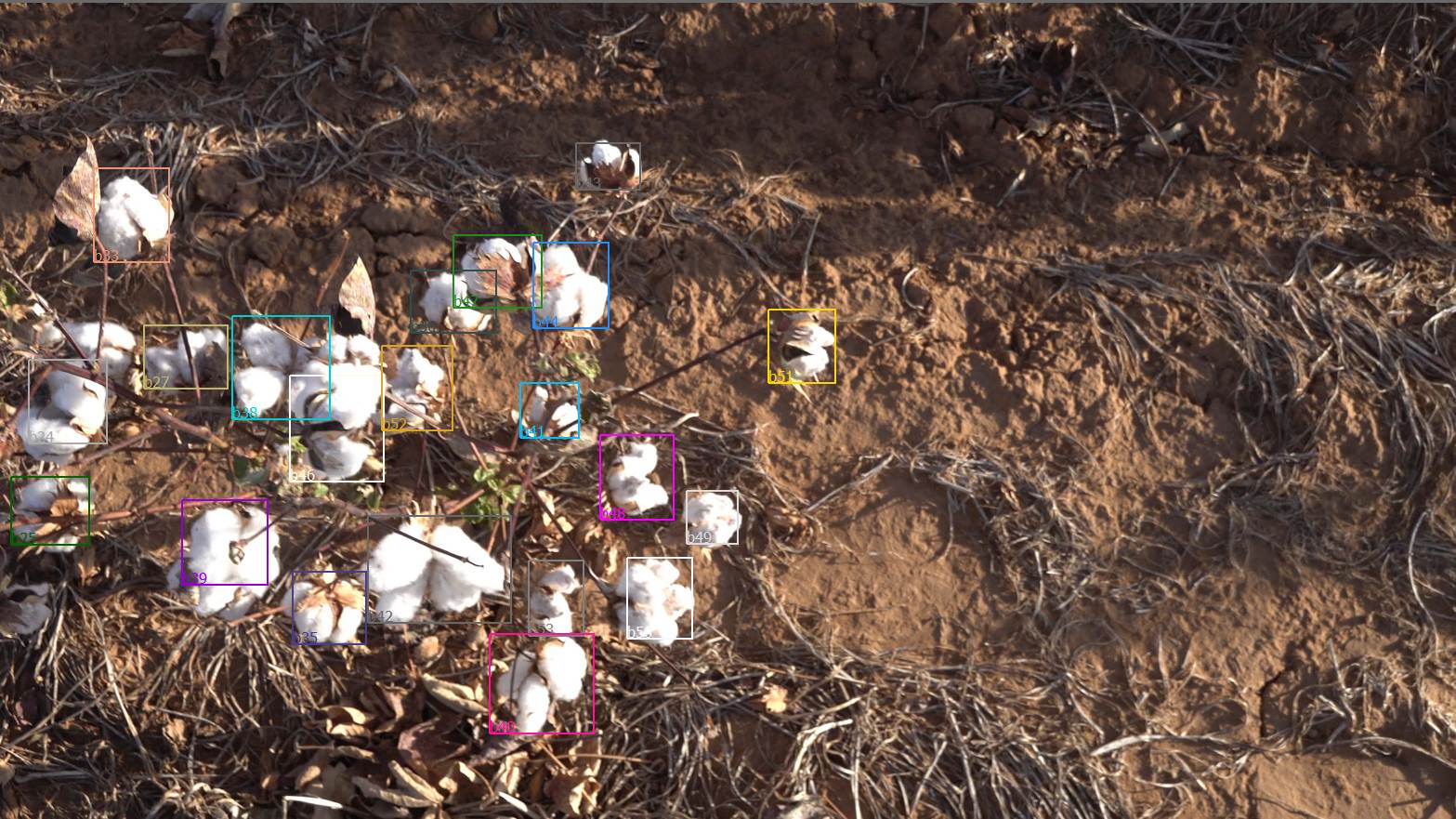}
  \label{fig:cotton_boll_dataset_1}
} \hspace{-4mm}
\subfloat[]{
  \includegraphics[width=0.47\columnwidth,height=33mm]{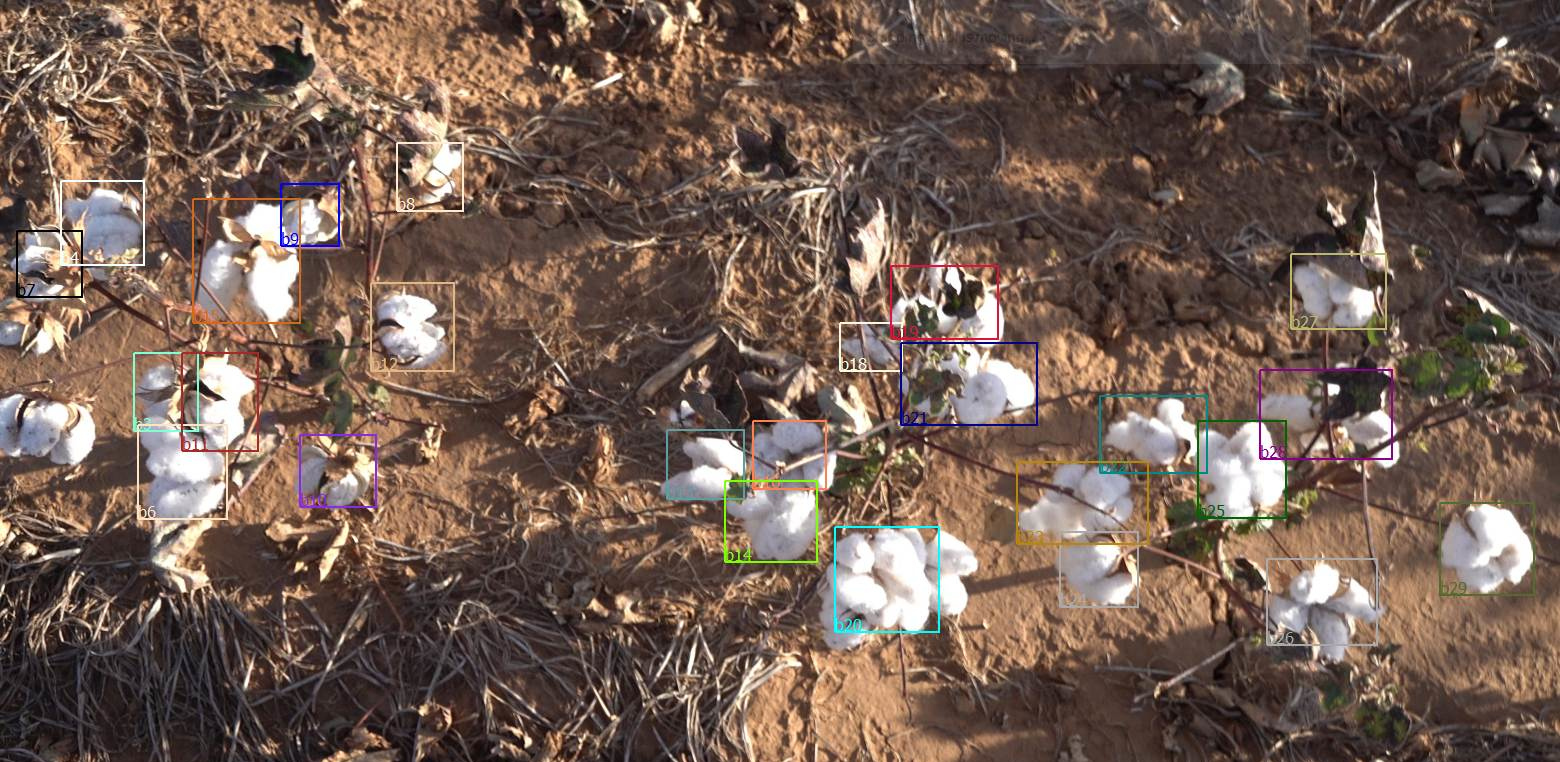}
  \label{fig:cotton_boll_dataset_2}
} \hspace{-4mm}
\subfloat[]{
  \includegraphics[width=0.47\columnwidth,height=33mm]{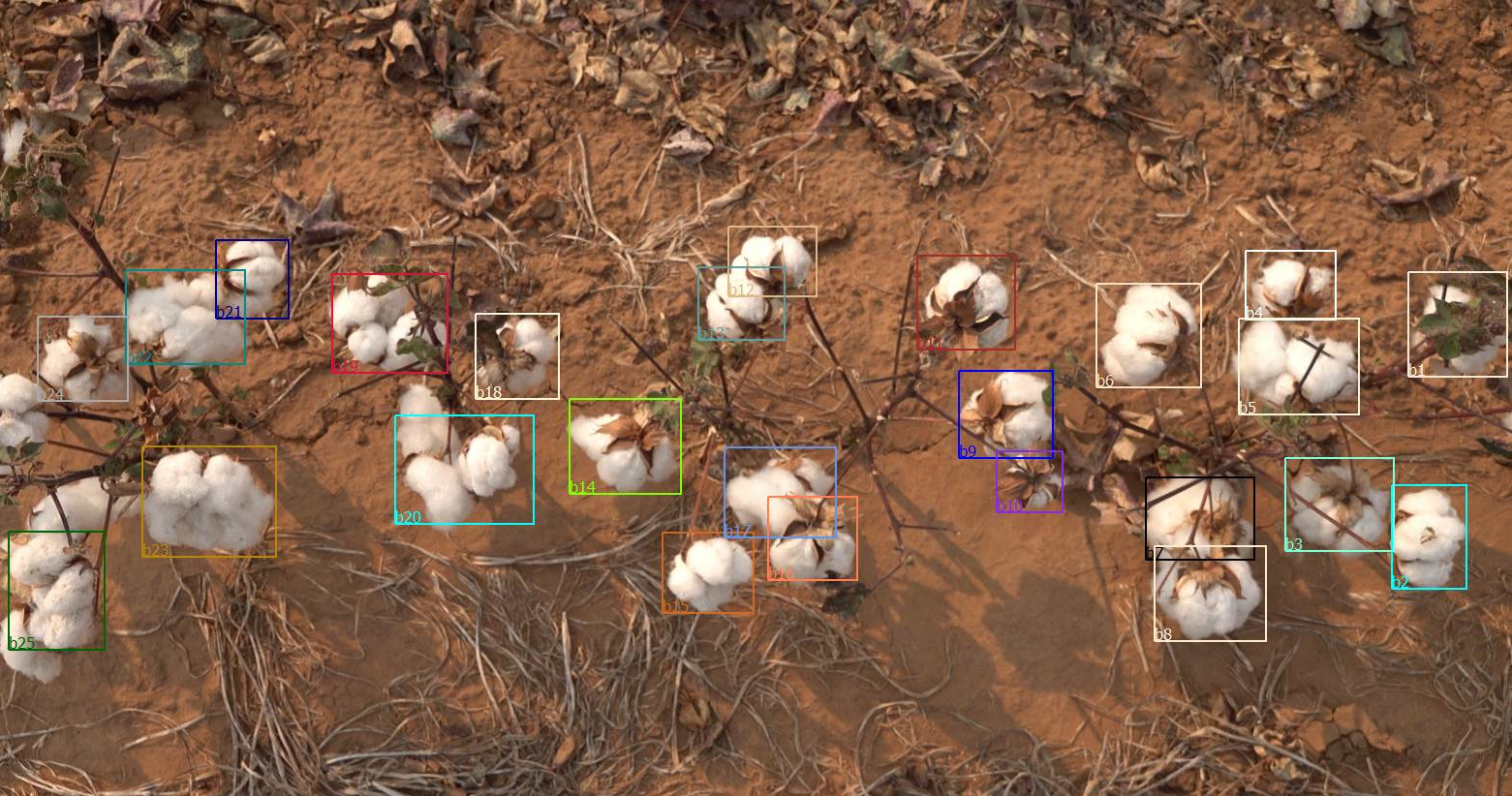}
  \label{fig:cotton_boll_dataset_3}
} \hspace{-4mm}
\subfloat[]{
  \includegraphics[width=0.47\columnwidth,height=33mm]{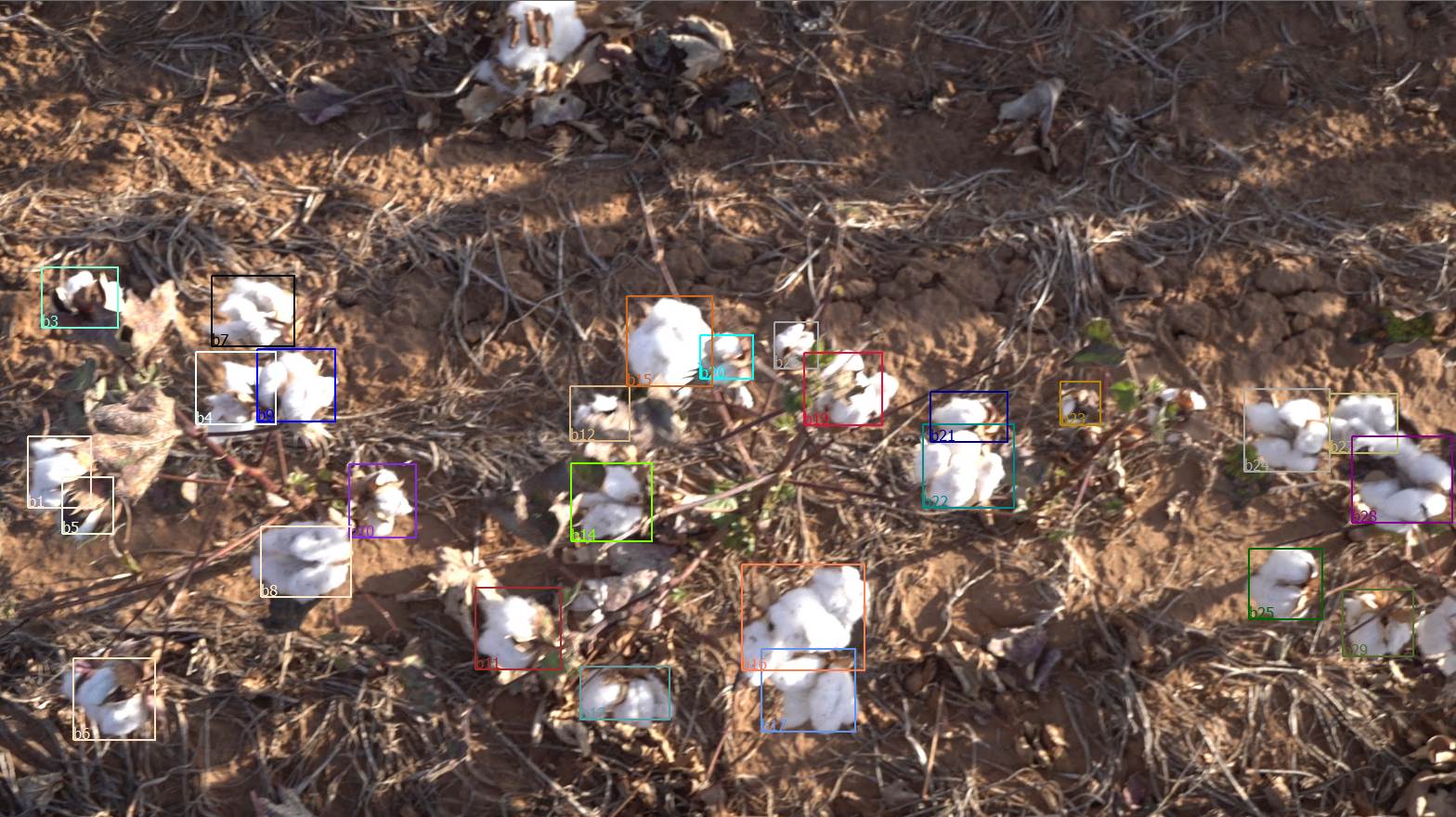}
  \label{fig:cotton_boll_dataset_4}
} \hspace{-4mm}
\subfloat[]{
  \includegraphics[width=0.47\columnwidth,height=33mm]{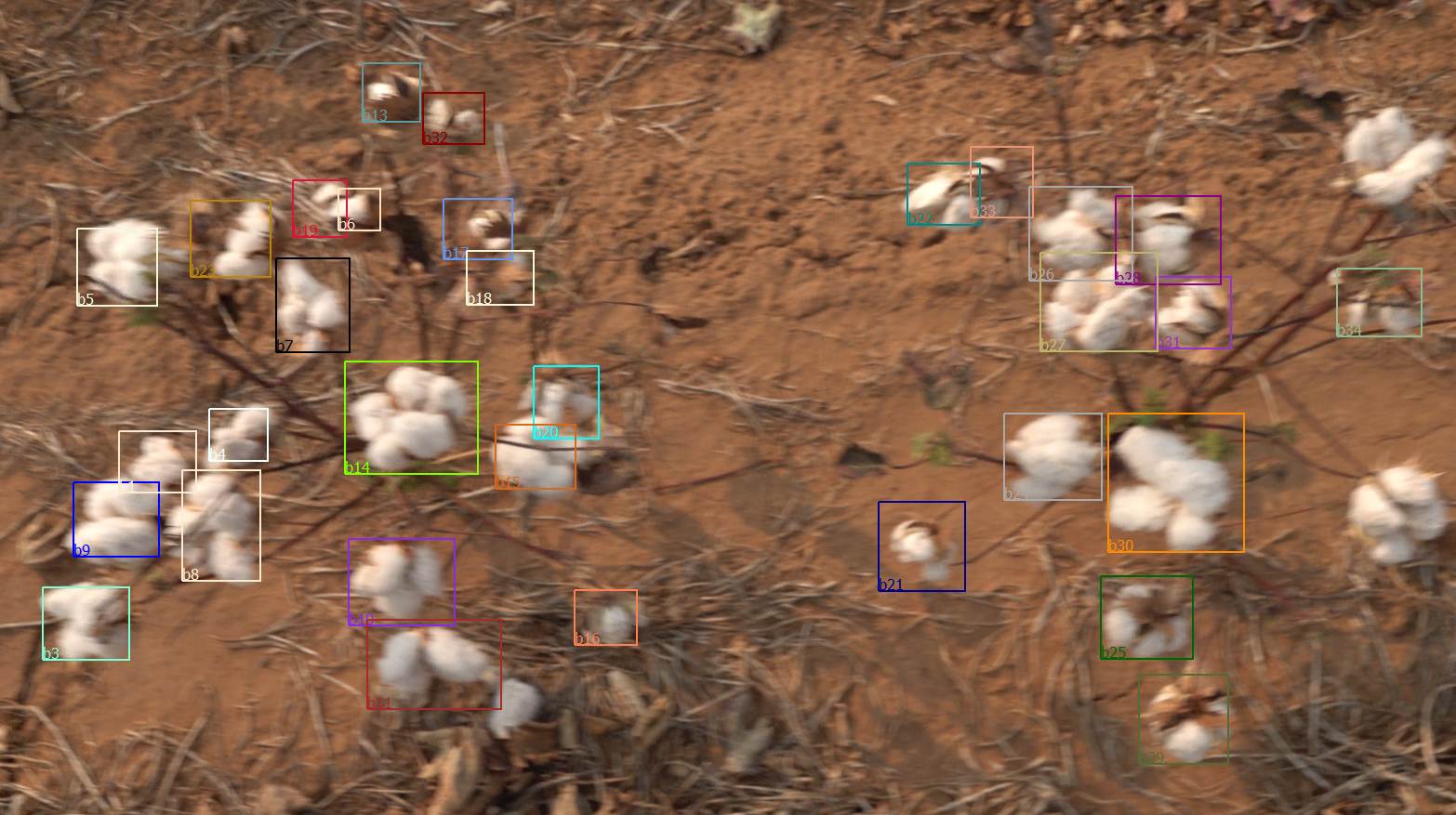}
  \label{fig:cotton_boll_dataset_5}
} \hspace{-4mm}
\subfloat[]{
  \includegraphics[width=0.47\columnwidth,height=33mm]{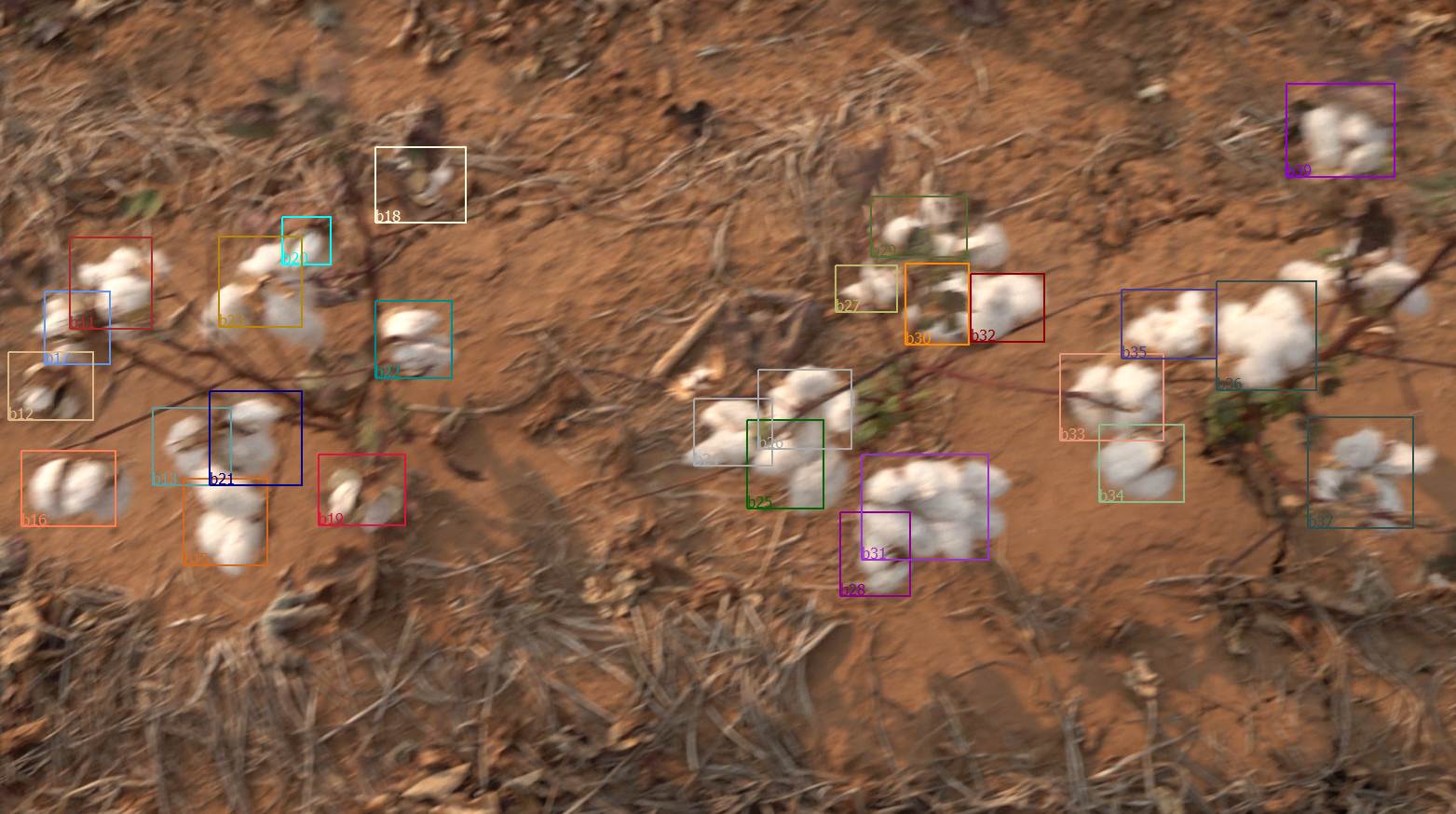}
  \label{fig:cotton_boll_dataset_6}
} \hspace{-4mm}
\subfloat[]{
  \includegraphics[width=0.47\columnwidth,height=33mm]{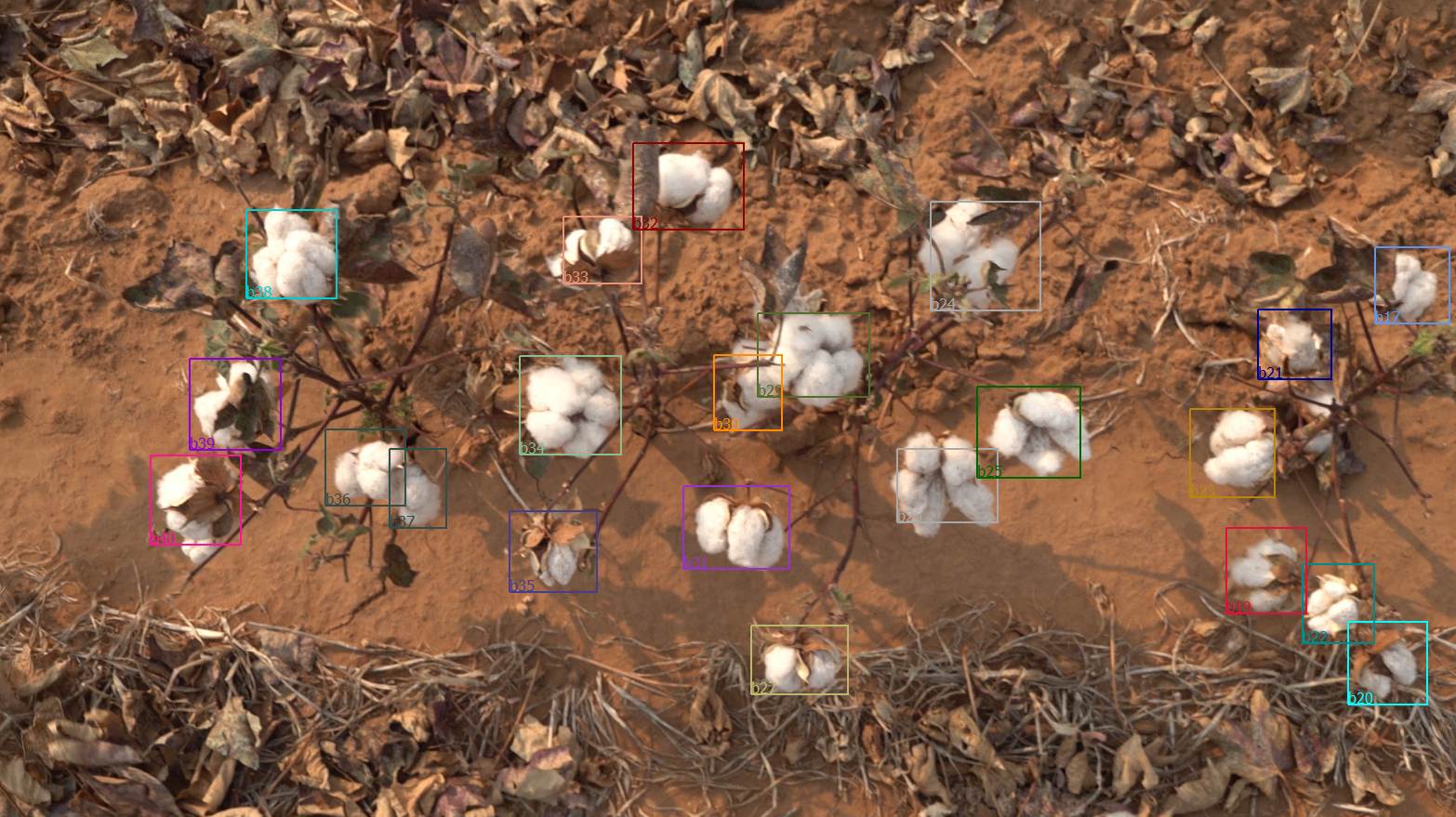}
  \label{fig:cotton_boll_dataset_7}
} \hspace{-4mm}
\subfloat[]{
  \includegraphics[width=0.47\columnwidth,height=33mm]{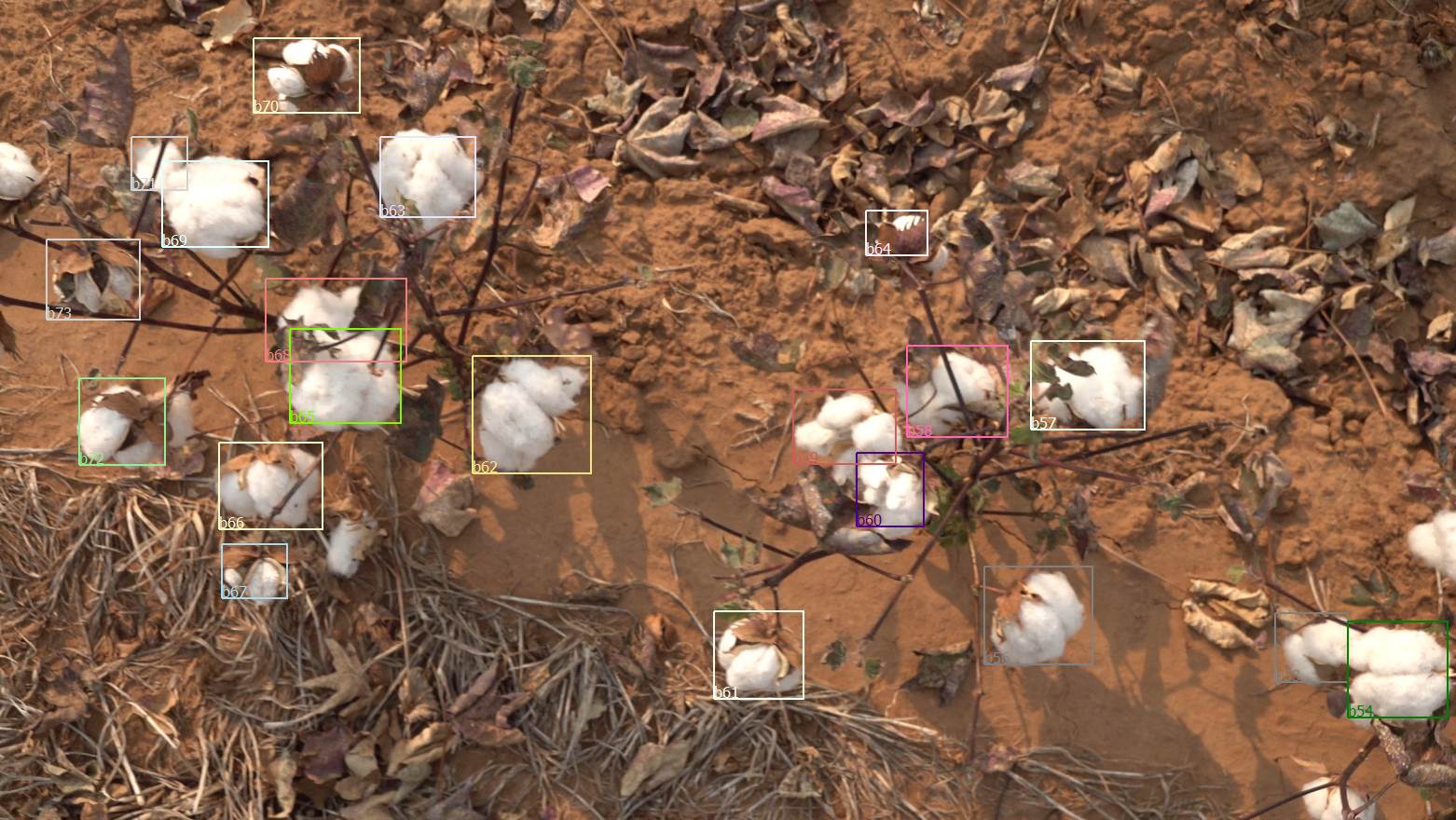}
  \label{fig:cotton_boll_dataset_8}
}
\caption{Examples (a-h) of annotated cotton boll field images with complex
backgrounds from the \textbf{TexCot22} \cite{muzaddid2023texcot22} dataset.}
\label{fig:cotton_boll_dataset}
\end{figure*}

To construct the \textbf{TexCot22} \cite{muzaddid2023texcot22} dataset we
captured multiple video sequences for training and testing. Similar to other
tracking datasets, each tracking sequence is 10 to 20 seconds in length. The
dataset consists of a total of 30 sequences of which 17 are for training and
the remaining 13 are for testing.

The video sequences were captured at 4K resolution and at distinct frame rates
(e.g., 10, 15, 30). There are typically 2 to 10 cotton bolls per cluster. The
average width and height of an annotated bounding box is approximately $230
\times 210$ pixels. 

To make the dataset robust to environmental conditions, we recorded the field
videos at separate times of day to account for varying lighting conditions. In
total, there are roughly $30\times300$ frames with 150,000 labeled instances. On
average there are 70 unique cotton bolls in each sequence. 

The directory structure of the dataset is similar to MOT17
\cite{milan2016mot16}. The ground truth and the detection files are also
available in MOT17 format. Hence, any tracking method that runs on MOT17 can
readily utilize \textbf{TexCot22} without any additional modifications. Example
ground-truth images from the dataset are displayed in
Fig.~\ref{fig:cotton_boll_dataset}.

\subsection{Annotation Rules}
\begin{table}
\centering
\begin{tabular}{  c|  p{6cm}   }
\toprule
 & \textbf{Rule}\\
\midrule
\textbf{What}      & Any open cotton bolls on plants excluding bolls that have fallen to the ground.\\\hline
\textbf{When}      & Start when the enclosing bounding box enters the frame.
                     Remove as soon as the bounding box goes beyond the frame border.\\\hline
\textbf{How}       & Annotations are not pixel perfect. The majority of pixels
                     belonging to a cotton boll should be contained by the bounding box.\\\hline
\textbf{Occlusion} & Annotate a cotton boll as long as it is partially visible and
                     distinguishable from the neighboring bolls. In the case of a long occlusion
                     interval, the same ID is assigned to the occluded boll as long as it is
                     identifiable.\\
\bottomrule
\end{tabular}
\caption{The annotation rules used for constructing the \textbf{TexCot22}
\cite{muzaddid2023texcot22} dataset.}
\label{tab:annotation_rules}
\end{table}

We followed a set of rules to exhaustively annotate all cotton bolls in each
sequence with bounding boxes. The bounding boxes around the cotton bolls are
very tight, however there may exist some pixels outside of the bounding box
that are part of the boll. A compiled a set of annotation rules is provided in
Table~\ref{tab:annotation_rules}.

\section{Experiments}
\label{sec:experiments}
\subsection{Implementation Details}
To make a fair comparison among existing tracking methods, we generated the
same set of detection bounding boxes utilizing a Cascade R-CNN
\cite{cai2018cascade} model with a ResNet-50 backbone. Using the
\textbf{TexCot22} training data, the model was trained for 100 epochs. The
training took place on a CentOS 7.6.1810 machine using an Intel Xeon E5-2620
2.10 GHz CPU, 132 GB of memory, and an NVIDIA GeForce GTX 1080 Ti GPU. The
detection accuracy of the trained model was 97\% on the test data. We also
tried the detector provided by the official Tracktor
\cite{bergmann2019tracking} GitHub repository. However, our detector achieved
the best results among all the trackers.

For calculating dense optical flow, we made use of the OpenCV
\cite{bradski2000opencv} implementation of the Gunnar-Farneback algorithm. The
estimated flow velocity of a bounding box is computed by averaging the
velocities of the center pixels (e.g., a $3\times3$ window at the bounding box
center). The problem of assigning the set of detected bounding boxes to the set
of predicted bounding boxes is solved via ByteTrack's \cite{zhang2022bytetrack}
association procedure.

To estimate the relative location based on the $k$-nearest neighbors, we
empirically opted for a neighbor size of three. If we considered too many
neighbors, then the RLA module gave a coarse relative location. Conversely, a
single neighbor often provided a noisy estimation. The relative locations along
the $x$ (width) and $y$ (height) directions were calculated independently using
\eqref{eqn:simplified_single_gaussian}. 

\begin{table*}
\centering
\begin{adjustbox}{max width=\textwidth}
\begin{tabular}{|l|c|c|c|c|c|c|c|c|}
\hline
\textbf{Method} & \textbf{IDP\textcolor{teal}{$\uparrow$}} & \textbf{IDR\textcolor{teal}{$\uparrow$}}
& \textbf{IDF$_1$\textcolor{teal}{$\uparrow$}} & \textbf{HOTA\textcolor{teal}{$\uparrow$}}
& \textbf{MOTA\textcolor{teal}{$\uparrow$}}    & \textbf{MOTP\textcolor{teal}{$\uparrow$}}
& \textbf{IDsw\textcolor{teal}{$\downarrow$}}  & \textbf{Frag\textcolor{teal}{$\downarrow$}}\\ \hline
\textbf{DeepSORT}      & 82.50\%          & 81.97\%          & 82.24\%          & 66.47\%          & 84.80\%          & 80.07\%          & 1751         & 633 \\ \hline
\textbf{Tracktor}      & 82.47\%          & 81.01\%          & 81.74\%          & 66.28\%          & 86.56\%          & 79.03\%          & 2070         & 787 \\ \hline
\textbf{ByteTrack}     & 90.88\%          & 88.76\%          & 89.80\%          & 71.19\%          & 88.60\%          & 80.32\%          & 1193         & 564 \\ \hline
\textbf{TrackFormer}   & 89.94\%          & 70.40\%          & 78.98\%          & 54.90\%          & 69.58\%          & 70.86\%          & \textbf{652} & \textbf{311} \\ \hline
\textbf{NTrack (ours)} & \textbf{93.28}\% & \textbf{91.70}\% & \textbf{92.49}\% & \textbf{73.56}\% & \textbf{89.25}\% & \textbf{81.49}\% & 1062         & 508 \\ \hline
\end{tabular}%
\end{adjustbox}
\caption{A comparison of NTrack against DeepSORT \cite{wojke2017simple},
Tracktor \cite{bergmann2019tracking}, ByteTrack \cite{zhang2022bytetrack}, and
TrackFormer \cite{meinhardt2022trackformer}. The arrow directions indicate the
optimal metric values.}
\label{tab:metric_comparison}
\end{table*}

\subsection{Cotton Boll Tracking Evaluation}
We evaluated multiple MOT metrics including higher order tracking accuracy
(HOTA) \cite{luiten2021hota}, identity-aware \cite{ristani2016performance}, and
those defined by CLEAR MOT \cite{bernardin2008evaluating}. Association and
localization are two major criteria for deciding tracking performance.  While
measures such as MOTA (accuracy) and MOTP (precision) emphasize localization,
the IDP (identification precision), IDR (identification recall), IDF$_1$
(identification \textit{F}$_1$ score), and IDsw (identity switches) put more
weight on maintaining true identity. The Frag (fragmentation) metric is the
number of times an object is lost, but then redetected in a future frame thus
fragmenting the track. The evaluation was performed against the following
state-of-the-art tracking methods: DeepSORT \cite{wojke2017simple}, Tracktor
\cite{bergmann2019tracking}, ByteTrack \cite{zhang2022bytetrack}, and
Trackformer \cite{meinhardt2022trackformer}. As shown in
Table~\ref{tab:metric_comparison}, NTrack outperforms these systems by a
significant margin in the majority of the metrics. 

Our primary goal was to design a tracking system that can count cotton bolls
with high accuracy. Thus, by design NTrack should perform better in ID
preserving performance metrics. Nevertheless, our tracker outperforms other
methods in localization measures as well. In addition, NTrack demonstrates its
overall superiority by exceeding others in the HOTA measure, which explicitly
balances the effect of performing accurate detection, association, and
localization into a single unified metric. To avoid data labeling
inconsistencies near the frame border, we considered the ground truth and
hypothesis bounding boxes that do not overlap with the frame margin. More
specifically, we consider 200 pixels on both sides of a frame as the margin.

\subsection{Cotton Boll Counting Evaluation}
\begin{table}
\centering
\begin{adjustbox}{max width=\textwidth}
\begin{tabular}{|l|c|c|}
\hline
\textbf{Method} & \textbf{MAPE\textcolor{teal}{$\downarrow$}} &\textbf{RMSE\textcolor{teal}{$\downarrow$}} \\\hline
\textbf{Deep learning\cite{tedesco2020convolutional}} & 9.00\%          & 9.00 (best case) \\ \hline
\textbf{Geometric-feature-based\cite{sun2019image}}   & 15.04\%         & 7.40 \\ \hline
\textbf{3D point cloud-based\cite{sun2020three}}      & 10.00\%         & 16.87 \\ \hline
\textbf{NTrack (ours)}                                & \textbf{4.00}\% & \textbf{4.73}\\ \hline
\end{tabular}%
\end{adjustbox}
\caption{Cotton boll counting error.}
\label{tab:counting_results_vs_counting}
\end{table}

Determining the number of unique cotton bolls in a given video sequence was a
prime requirement in designing NTrack. Therefore, when creating the tracker we
focused on maintaining the true identity of the bolls. As shown in
Table~\ref{tab:counting_results_vs_counting}, the counting results show that
NTrack performs exceptionally well at the counting task when compared to other
cotton boll counting methods. These results are based on the mean absolute
percentage error (MAPE) and root mean square error (RMSE), which are defined as
\begin{align}
  \text{MAPE} &= \frac{1}{n} \sum_{i=1}^{n}\frac{|C_{gt} - C_h|} {C_{gt}},
  \label{eqn:mape}\\
  \text{RMSE} &=\sqrt{\frac{1}{n}\sum_{i=1}^{n}{(C_{gt} - C_h)^2}}.
  \label{eqn:rmse}
\end{align}
In \eqref{eqn:mape} and \eqref{eqn:rmse}, $C_{gt}$ and $C_h$ are the number of
unique cotton bolls detected manually (i.e., the ground truth) and by NTrack in
the $i$th video sequence, respectively, and $n$ is the number of video
sequences. The results reported for the other methods in
Table~\ref{tab:counting_results_vs_counting} are taken from their respective
papers since the datasets and source code are not publicly available. 

\begin{table*}
\begin{tabularx}{\textwidth}{c Y| *{2}{Y}|*{2}{Y}|*{2}{Y}|*{2}{Y}|*{2}{Y}}
\toprule
\textbf{Sequence} & \textbf{GT}
 & \multicolumn{2}{c}{\textbf{NTrack (ours)}}
 & \multicolumn{2}{c}{\textbf{DeepSORT}}
 & \multicolumn{2}{c}{\textbf{Tracktor}}
 & \multicolumn{2}{c}{\textbf{ByteTrack}}
& \multicolumn{2}{c}{\textbf{TrackFormer}}\\
\cmidrule(lr){3-4} \cmidrule(l){5-6} \cmidrule(l){7-8} \cmidrule(l){9-10} \cmidrule(l){11-12}
  & & Count & Error \%   & Count & Error \% & Count & Error \% & Count & Error \% & Count & Error \% \\
\midrule
    vid09\_01  & 66      & 66 &  0       & 115 &  74              & 220 & 233        & 77 &  17       & 58 &  12    \\
    vid09\_02  & 72      & 80 & 11       & 146 & 103              & 234 & 225        & 90 &  25       & 74 &  3    \\
    vid09\_03  & 72      & 67 &  4       & 105 &  46              & 208 & 189          & 78 &  8        & 62 &  14     \\\hline

    vid14\_01  & 95      & 102 & 7       & 233 & 145              & 396 & 317         & 123 &  29      & 103 &  8   \\\hline

    vid23\_01  & 60      & 56 &  7       & 136 & 127              & 187 & 212         & 93 &  55       & 57 &  5   \\
    vid23\_02  & 50      & 46 &  8       & 64  &  28              & 136 & 172          & 52 &  4        & 49 &  2     \\
    vid23\_03  & 96      & 95 &  1       & 140 &  46              & 223 & 132         & 101 &  5       & 88 &  8    \\\hline

    vid25\_01  & 68      & 67 &  1       & 80  &  18              & 119 & 75           & 72 &  5        & 64 &  6    \\
    vid25\_02  & 73      & 72 &  1       & 103 &  41              & 184 & 152         & 88 &  21       & 57 &  22   \\
    vid25\_03  & 61      & 60 &  2       & 69  &  10              & 111 & 10           & 62 &  2        & 62 &  2    \\\hline

    vid26\_01  & 51      & 51 &  0       & 86  &  69              & 134 & 163         & 60 &  18       & 53 &  4   \\
    vid26\_02  & 66      & 67 &  1       & 79  &  20              & 120 & 82           & 69 &  5        & 59 &  11    \\
    vid26\_03  & 66      & 69 &  5       & 82  &  24              & 126 & 91           & 71 &  8        & 71 &  8    \\
\bottomrule
    Mean Error   &       &    &  \textbf{4}&   &  55              &   &163             &  &  15         &  &  8               \\
\bottomrule
\end{tabularx}
\caption{The accuracy of NTrack against DeepSORT \cite{wojke2017simple},
Tracktor \cite{bergmann2019tracking}, ByteTrack \cite{zhang2022bytetrack}, and
TrackFormer \cite{meinhardt2022trackformer} is shown by comparing the ground
truth (GT) cotton boll counts with the estimated counts.}
\label{tab:counting_results_vs_tracking}
\end{table*}

Table~\ref{tab:counting_results_vs_tracking} shows the performance of NTrack
against state-of-the-art tracking techniques. The results in
Table~\ref{tab:counting_results_vs_tracking} were obtained from experiments
done with the same test dataset (\textbf{TexCot22}) and protocol. NTrack
outperformed all of the other methods. It is interesting to note that
TrackFormer, which is based on a transformer architecture, was able to track
cotton bolls more consistently than NTrack in terms of ID switching and
fragmentation (Table~\ref{tab:metric_comparison}). Nevertheless, TrackFormer's
counting error (8\%) is twice that of NTrack (4\%).

\begin{table}[]
\centering
\begin{tabular}{|l|l|l|l|l|}
\hline
       & ByteTrack & DeepSORT  & Tracktor & TrackFormer \\ \hline
NTrack & 0.003945  & 0.001769 & 0.000027 & 0.033470   \\ \hline
\end{tabular}
\caption{The hypothesis testing results on the mean cotton boll counting error.
For each column, we report the test result between NTrack and the competing
method in terms of the p-value.}
\label{tab:hypothesis_testing}
\end{table}

To demonstrate that the difference in counting performance reported in
Table~\ref{tab:counting_results_vs_tracking} is statistically significant, we
conducted hypothesis testing. Specifically, we performed one-sided paired
t-tests against the other tracking techniques. The null hypothesis of the test
is the assumption that the mean counting error of NTrack is greater than or
equal to other methods, i.e.,
\begin{equation}
  H_0: \mu_{n} \geq \mu_{x}.
  \label{eq:null_hypothesis}
\end{equation}
Conversely, the alternative hypothesis is the assumption that the mean counting
error of NTrack is less than the other methods, i.e.,
\begin{equation}
  H_a: \mu_{n} < \mu_{x}. 
  \label{eq:alternative_hypothesis}
\end{equation}
In \eqref{eq:null_hypothesis} and \eqref{eq:alternative_hypothesis}, $\mu_{n}$
and $\mu_{x}$ are the mean counting errors of NTrack and the other techniques
(e.g., DeepSORT, Tracktor, ByteTrack, and TrackFormer), respectively. The test
results (p-values) in Table~\ref{tab:hypothesis_testing} demonstrate that we can
reject the null hypothesis at a significance level of $\alpha = 0.05$ in favor
of the alternative hypothesis. In other words, with 95\% confidence we can
conclude that the test data provides sufficient evidence to support the
observation that NTrack's mean counting error is lower than the other methods.

\subsection{Qualitative Evaluation}
\begin{figure*}
\centering
\begin{subfigure}[b]{0.32\textwidth}
  \begin{tikzpicture}[spy using outlines={ circle, magnification=2.5, size=2.7cm, height=2cm, spy_col, connect spies}]
    \node (image) {\includegraphics[trim=625  200 650 50,clip, width=.99\textwidth]{qualitative_comp/sample1_ours_f80}};
    \spy[every spy on node/.append style={ultra thick}] on (-1.65,.32)  in node [right] at  (0,-0.7);
  \end{tikzpicture}
\end{subfigure}
\begin{subfigure}[b]{0.32\textwidth}
  \begin{tikzpicture}[spy using outlines={ circle, magnification=2.5, size=2.7cm, height=2cm, spy_col, connect spies}]
    \node {\includegraphics[trim=625  200 650 50,clip, width=.99\textwidth]{qualitative_comp/sample1_ours_f122}};
    \spy[every spy on node/.append style={ultra thick}] on (-.85,.18) in node [right] at  (.2,-0.7);
  \end{tikzpicture}
\end{subfigure}
\begin{subfigure}[b]{0.32\textwidth}
  \begin{tikzpicture}[spy using outlines={circle, magnification=2.2, size=2.7cm, height=2cm, spy_col, connect spies}]
    \node {\includegraphics[trim=625  200 650 50,clip, width=.99\textwidth]{qualitative_comp/sample1_ours_f177}};
    \spy [every spy on node/.append style={ultra thick}] on (1.8,.55) in node [right] at (-1,-0.7);
  \end{tikzpicture}
\end{subfigure}

\begin{subfigure}[b]{0.32\textwidth}
  \begin{tikzpicture}[spy using outlines={ circle, magnification=2.5, size=2.7cm, height=2cm, spy_col, connect spies}]
    \node (image) {\includegraphics[trim=625  200 650 50,clip, width=.99\textwidth]{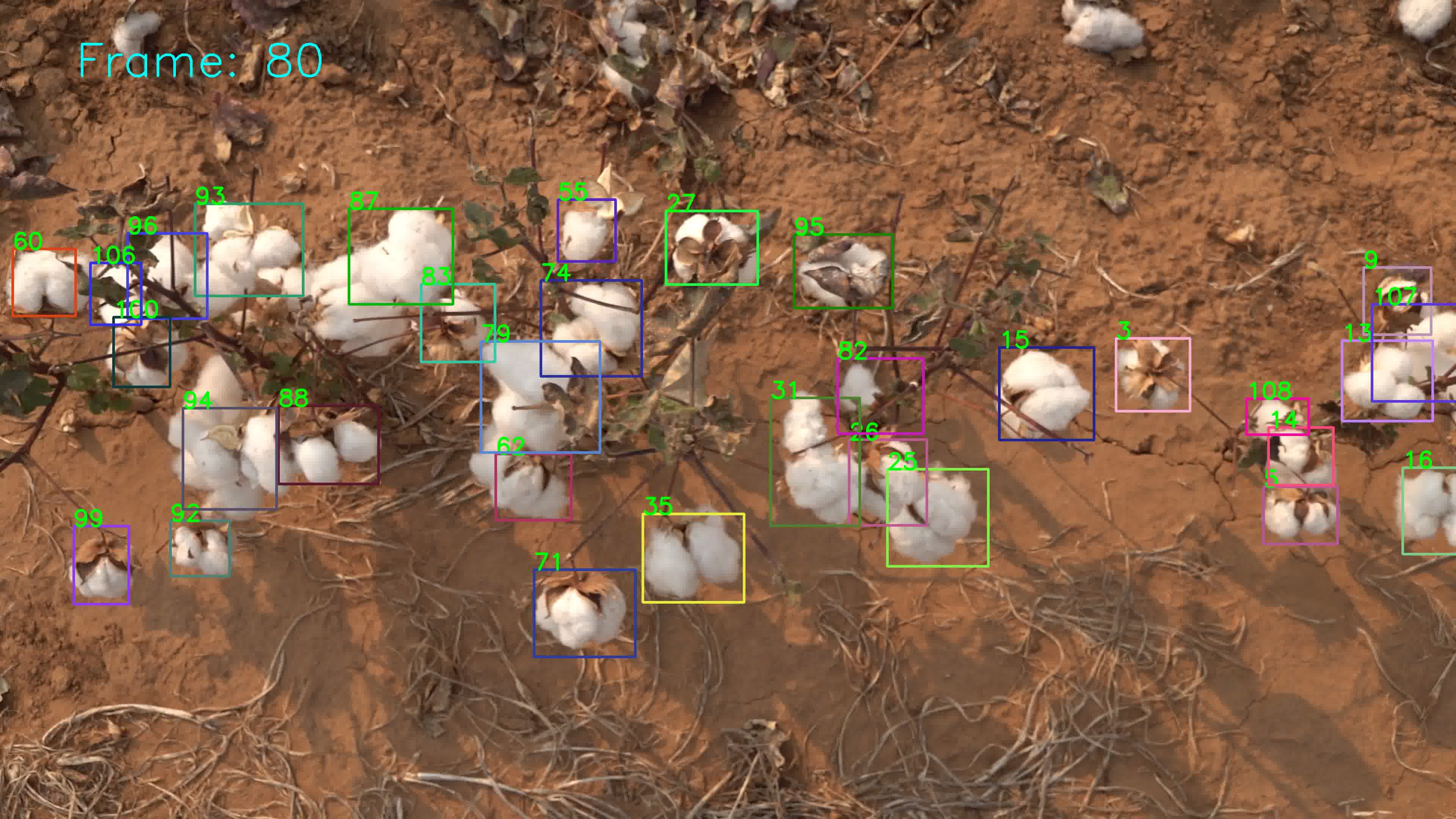}};
    \spy[every spy on node/.append style={ultra thick}] on (-1.65,.32)  in node [right] at  (0,-0.7);
  \end{tikzpicture}
\end{subfigure}
\begin{subfigure}[b]{0.32\textwidth}
  \begin{tikzpicture}[spy using outlines={ circle, magnification=2.5, size=2.7cm, height=2cm, spy_col, connect spies}]
    \node {\includegraphics[trim=625  200 650 50,clip, width=.99\textwidth]{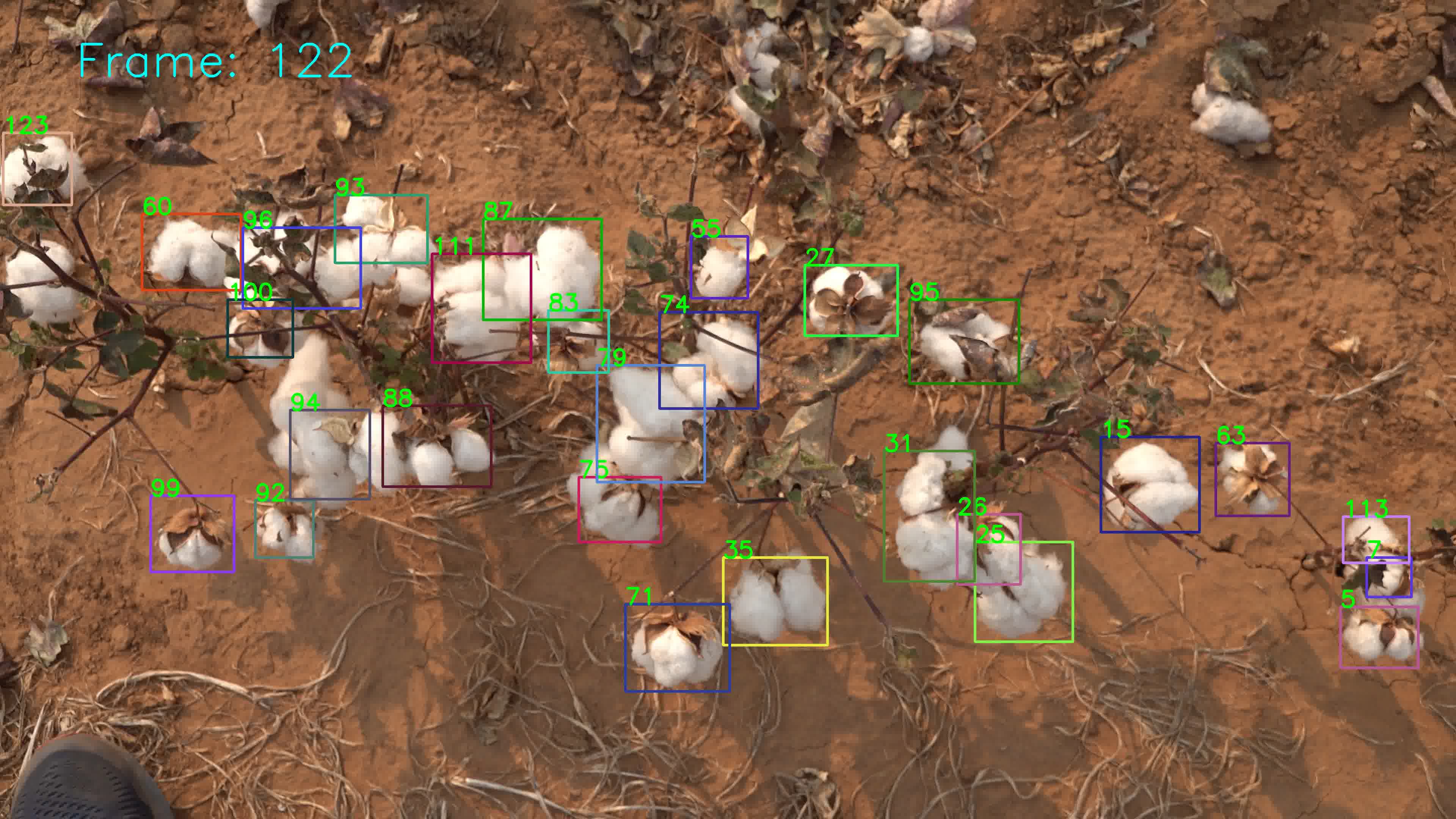}};
    \spy[every spy on node/.append style={ultra thick}] on (-.85,.18) in node [right] at  (.2,-0.7);
  \end{tikzpicture}
\end{subfigure}
\begin{subfigure}[b]{0.32\textwidth}
  \begin{tikzpicture}[spy using outlines={circle, magnification=2.2, size=2.7cm, height=2cm, spy_col, connect spies}]
    \node {\includegraphics[trim=625  200 650 50,clip, width=.99\textwidth]{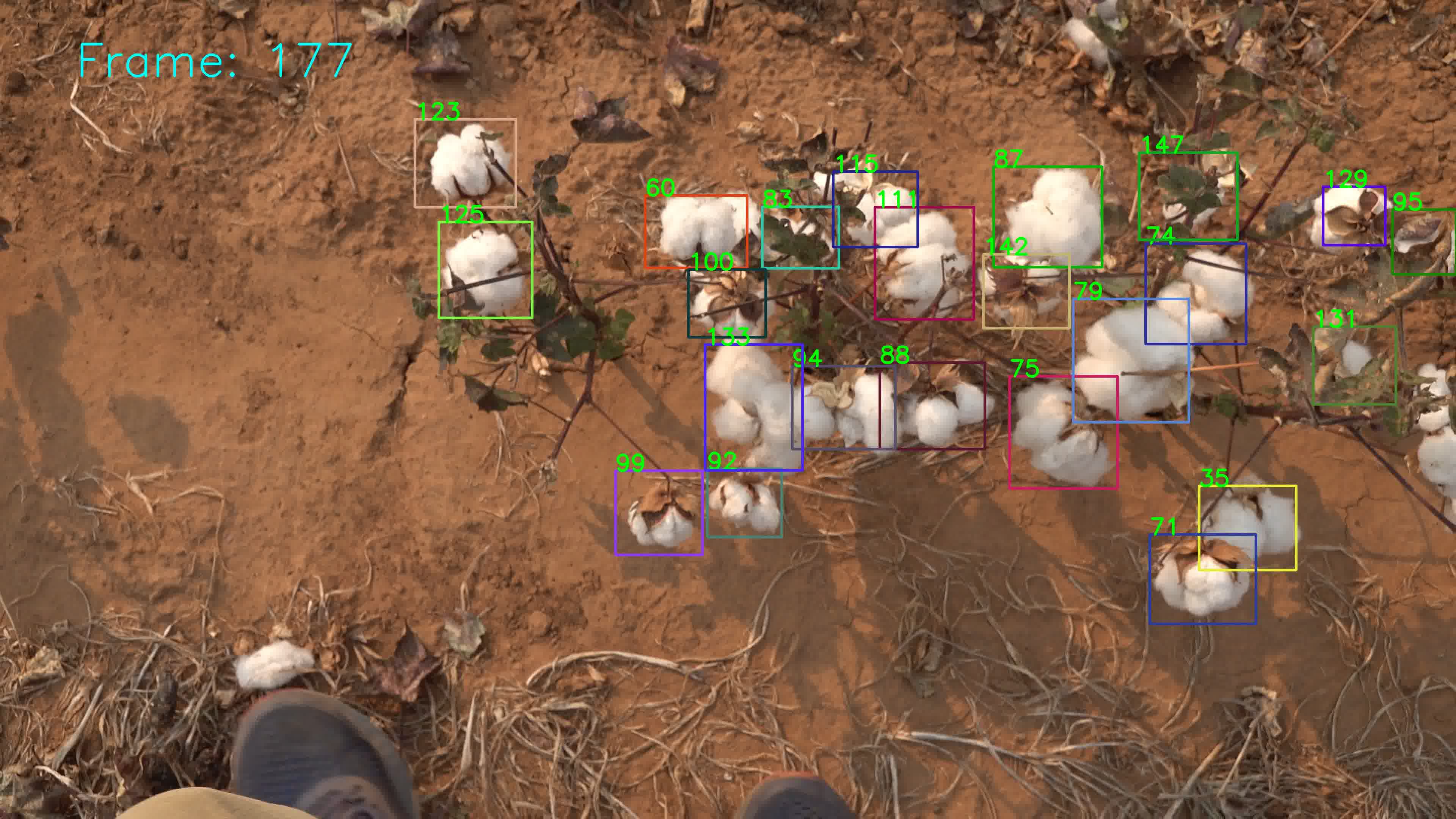}};
    \spy [every spy on node/.append style={ultra thick}] on (1.8,.55) in node [right] at (-1,-0.7);
  \end{tikzpicture}
\end{subfigure}

\begin{subfigure}[b]{0.32\textwidth}
  \begin{tikzpicture}[spy using outlines={ circle, magnification=2.5, size=2.7cm, height=2cm, spy_col, connect spies}]
    \node (image) {\includegraphics[trim=625  200 650 50,clip, width=.99\textwidth]{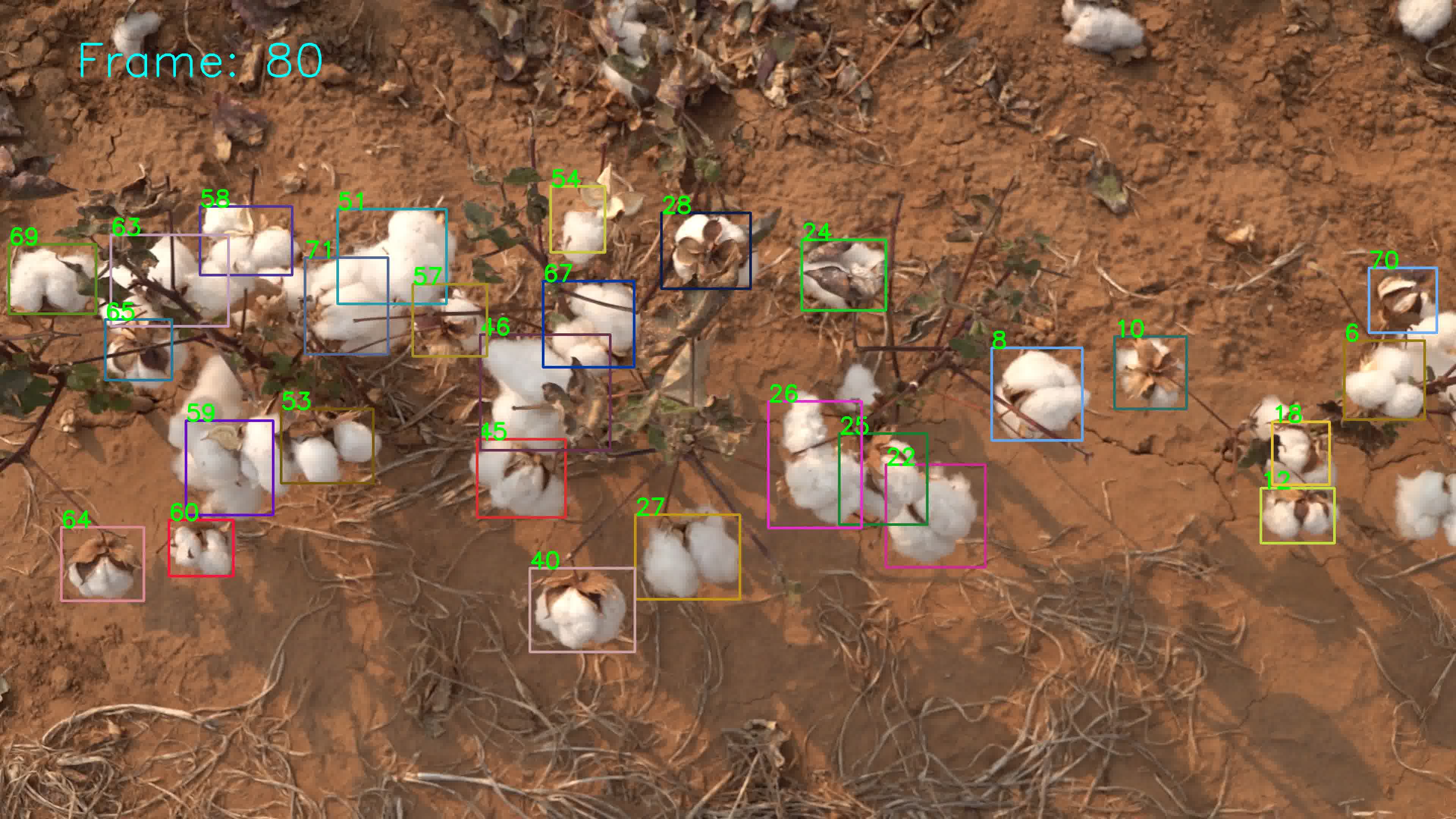}};
    \spy[every spy on node/.append style={ultra thick}] on (-1.5,.32)  in node [right] at  (0,-.7);
  \end{tikzpicture}
\end{subfigure}
\begin{subfigure}[b]{0.32\textwidth}
  \begin{tikzpicture}[spy using outlines={ circle, magnification=2.5, size=2.7cm, height=2cm, spy_col, connect spies}]
    \node {\includegraphics[trim=625  200 650 50,clip, width=.99\textwidth]{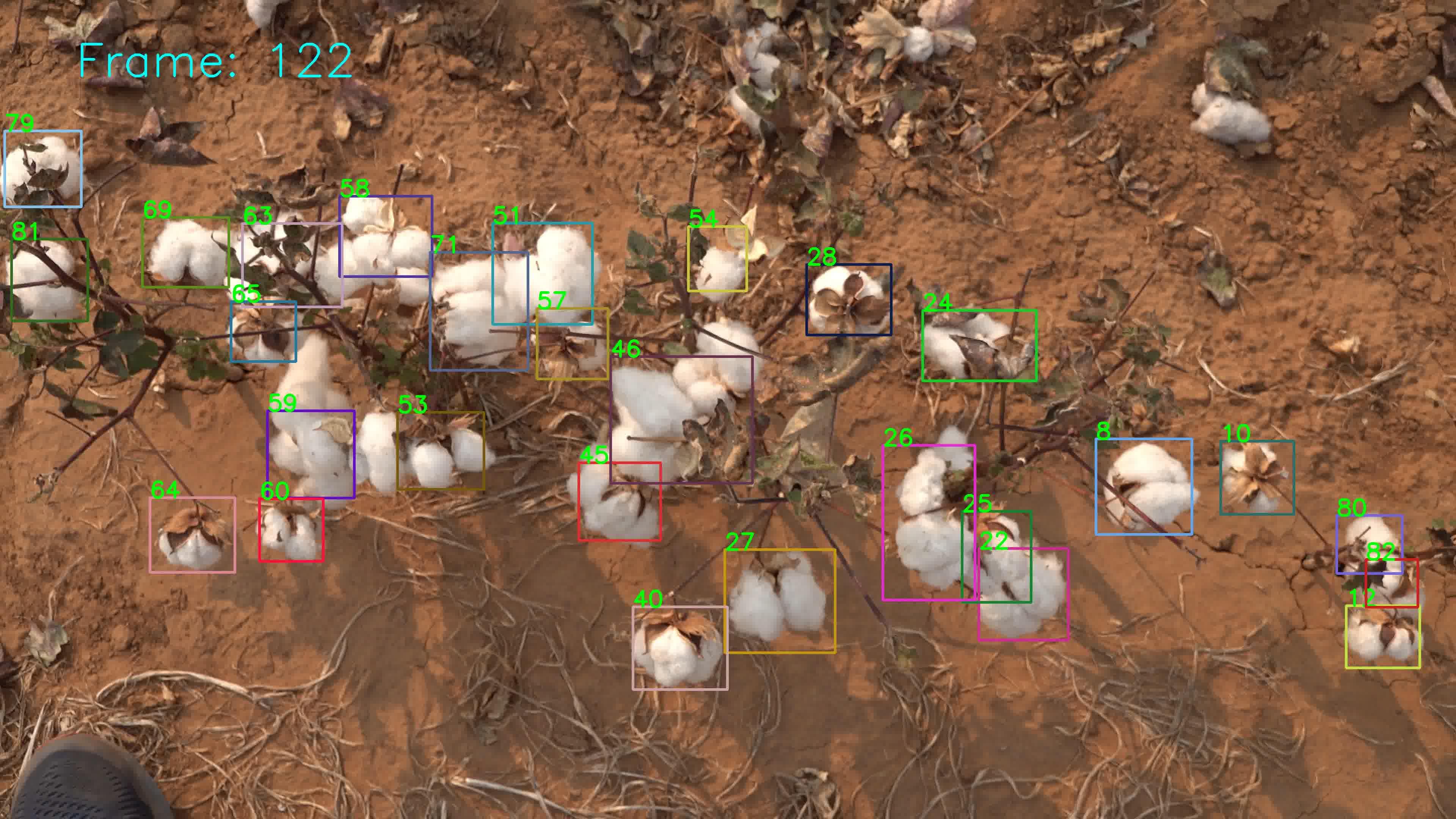}};
    \spy[every spy on node/.append style={ultra thick}] on (-.75,.18) in node [right] at  (.2,-0.7);
  \end{tikzpicture}
\end{subfigure}
\begin{subfigure}[b]{0.32\textwidth}
  \begin{tikzpicture}[spy using outlines={circle, magnification=2.2, size=2.7cm, height=2cm, spy_col, connect spies}]
    \node {\includegraphics[trim=625  200 650 50,clip, width=.99\textwidth]{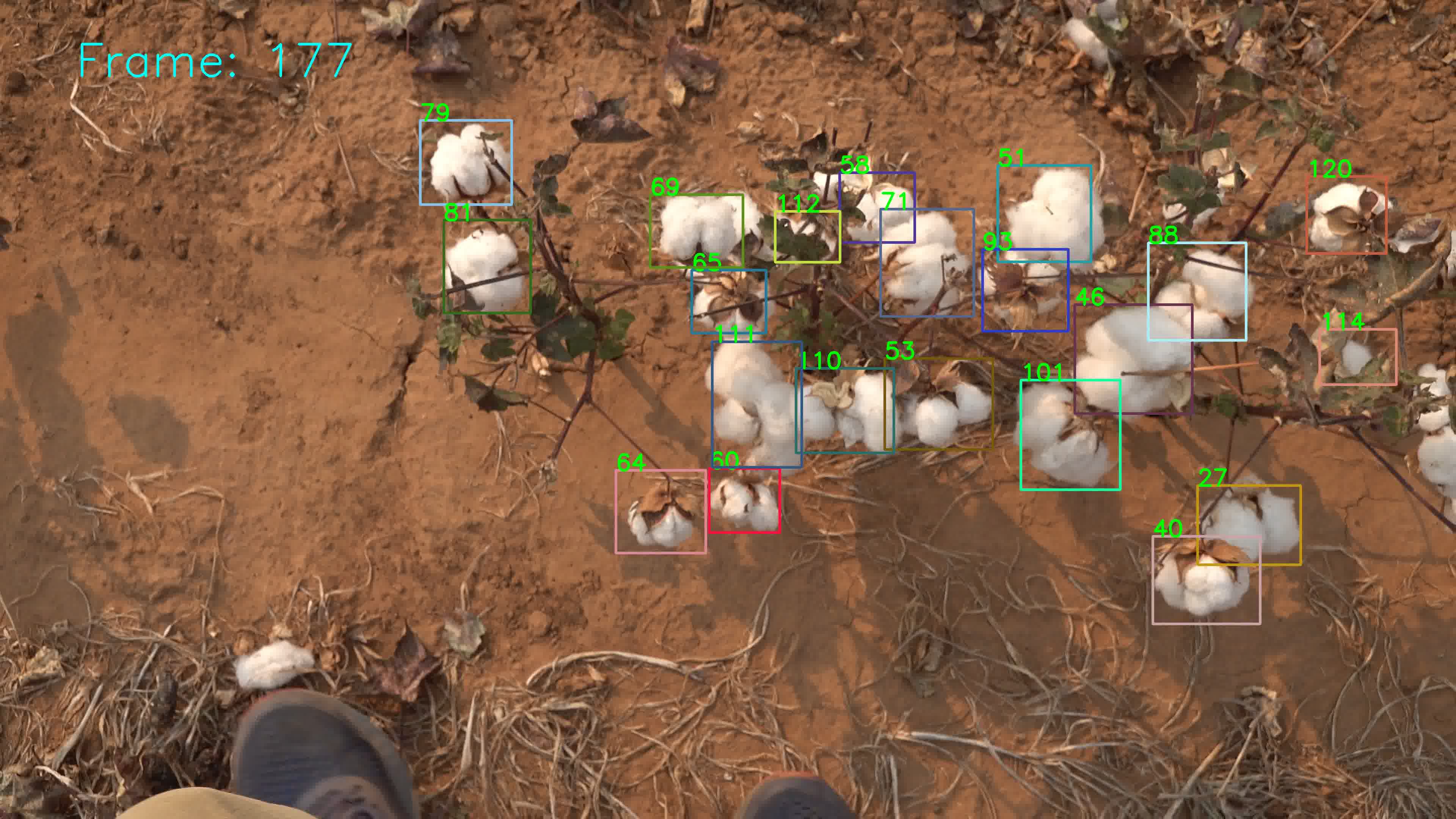}};
    \spy [every spy on node/.append style={ultra thick}] on (1.95,.55) in node [right] at (-1.2,-0.7);
  \end{tikzpicture}
\end{subfigure}

\begin{subfigure}[b]{0.32\textwidth}
  \begin{tikzpicture}[spy using outlines={ circle, magnification=2.5, size=2.7cm, height=2cm, spy_col, connect spies}]
    \node (image) {\includegraphics[trim=625  200 650 50,clip, width=.99\textwidth]{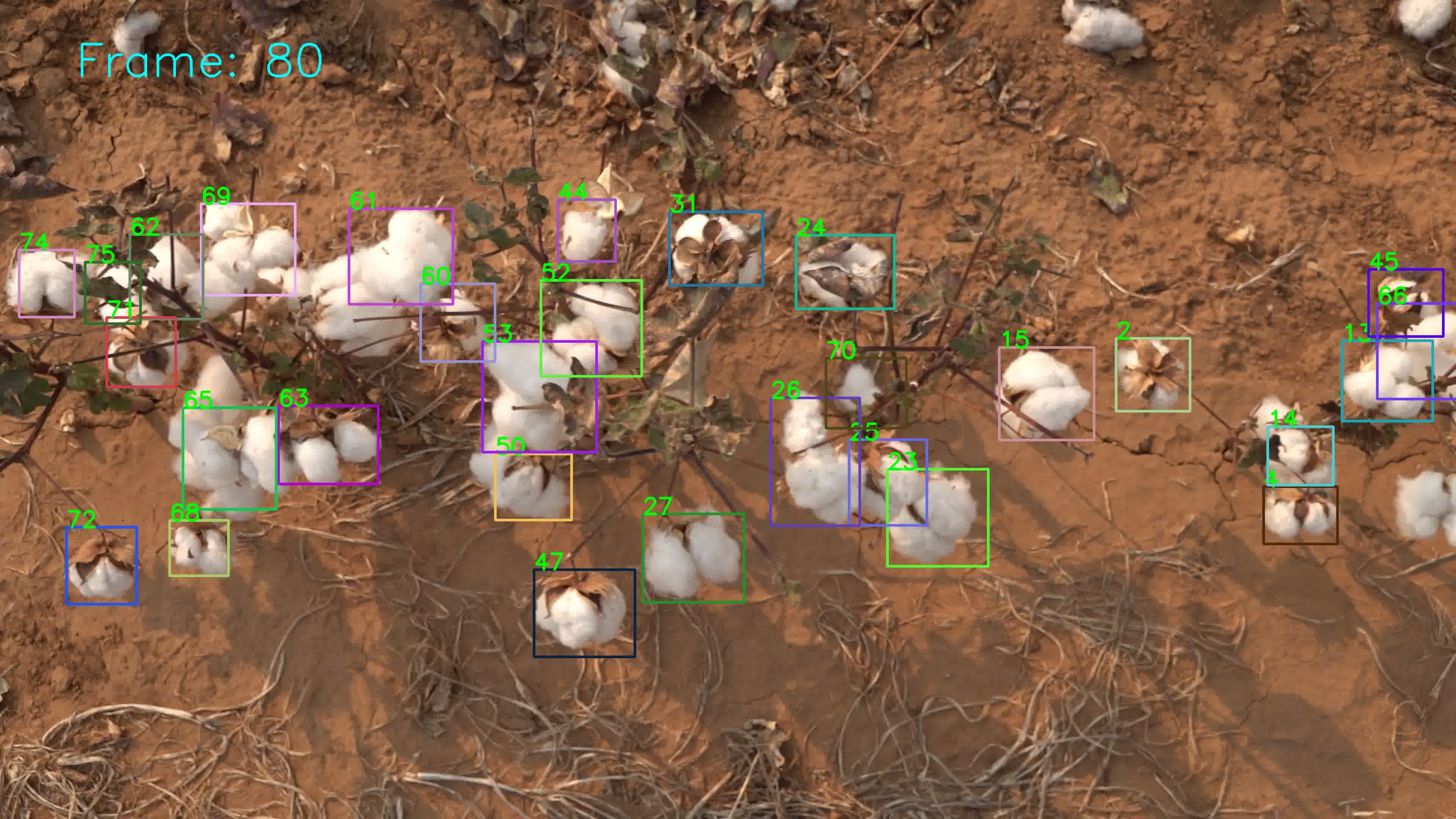}};
    \spy[every spy on node/.append style={ultra thick}] on (-1.65,.32)  in node [right] at  (0,-0.7);

  \end{tikzpicture}
\end{subfigure}
\begin{subfigure}[b]{0.32\textwidth}
  \begin{tikzpicture}[spy using outlines={ circle, magnification=2.5, size=2.7cm, height=2cm, spy_col, connect spies}]
    \node {\includegraphics[trim=625  200 650 50,clip, width=.99\textwidth]{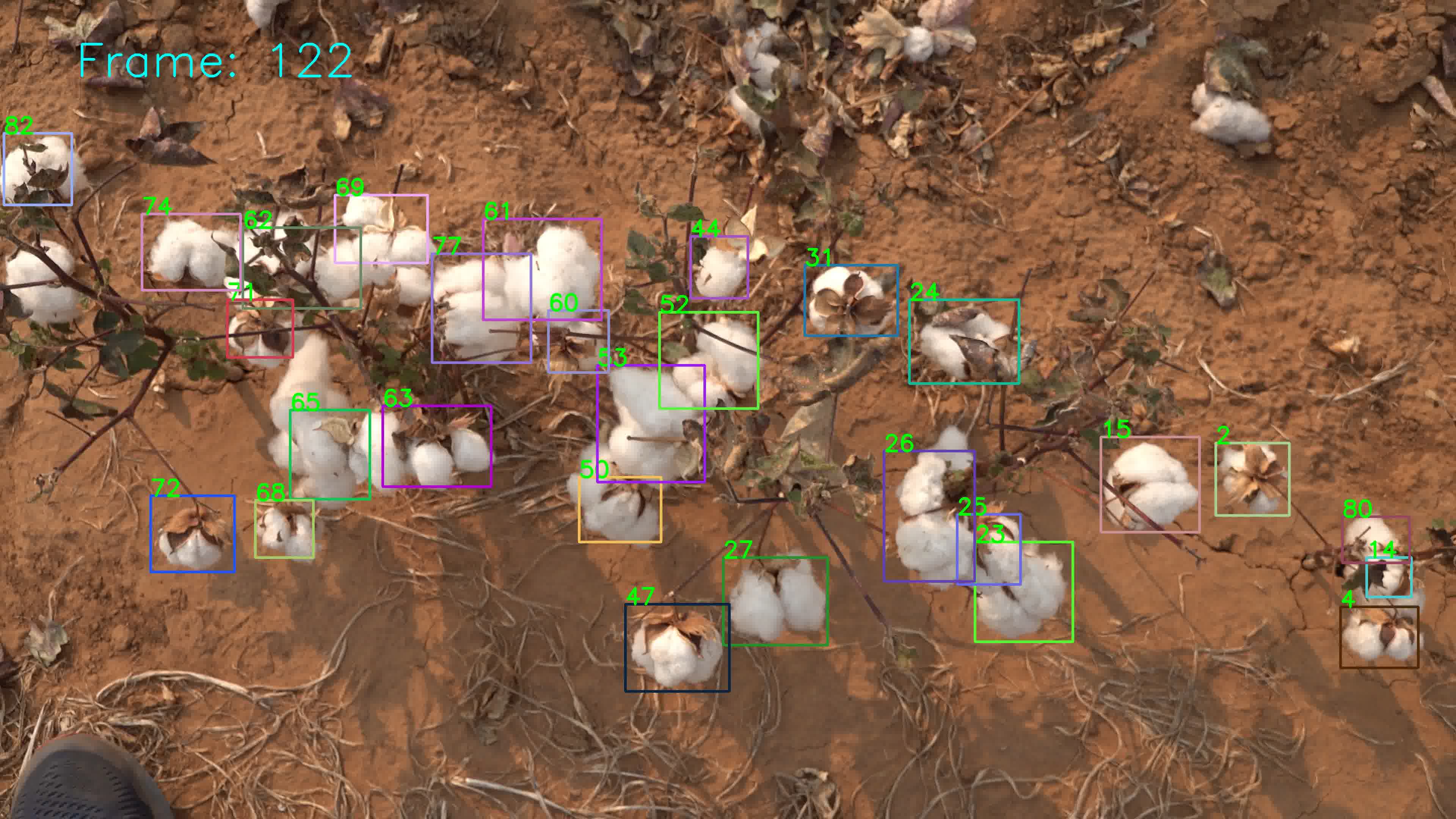}};
    \spy[every spy on node/.append style={ultra thick}] on (-.85,.18) in node [right] at  (.2,-.7);
  \end{tikzpicture}
\end{subfigure}
\begin{subfigure}[b]{0.32\textwidth}
  \begin{tikzpicture}[spy using outlines={circle, magnification=2.2, size=2.7cm, height=2cm, spy_col, connect spies}]
    \node {\includegraphics[trim=625  200 650 50,clip, width=.99\textwidth]{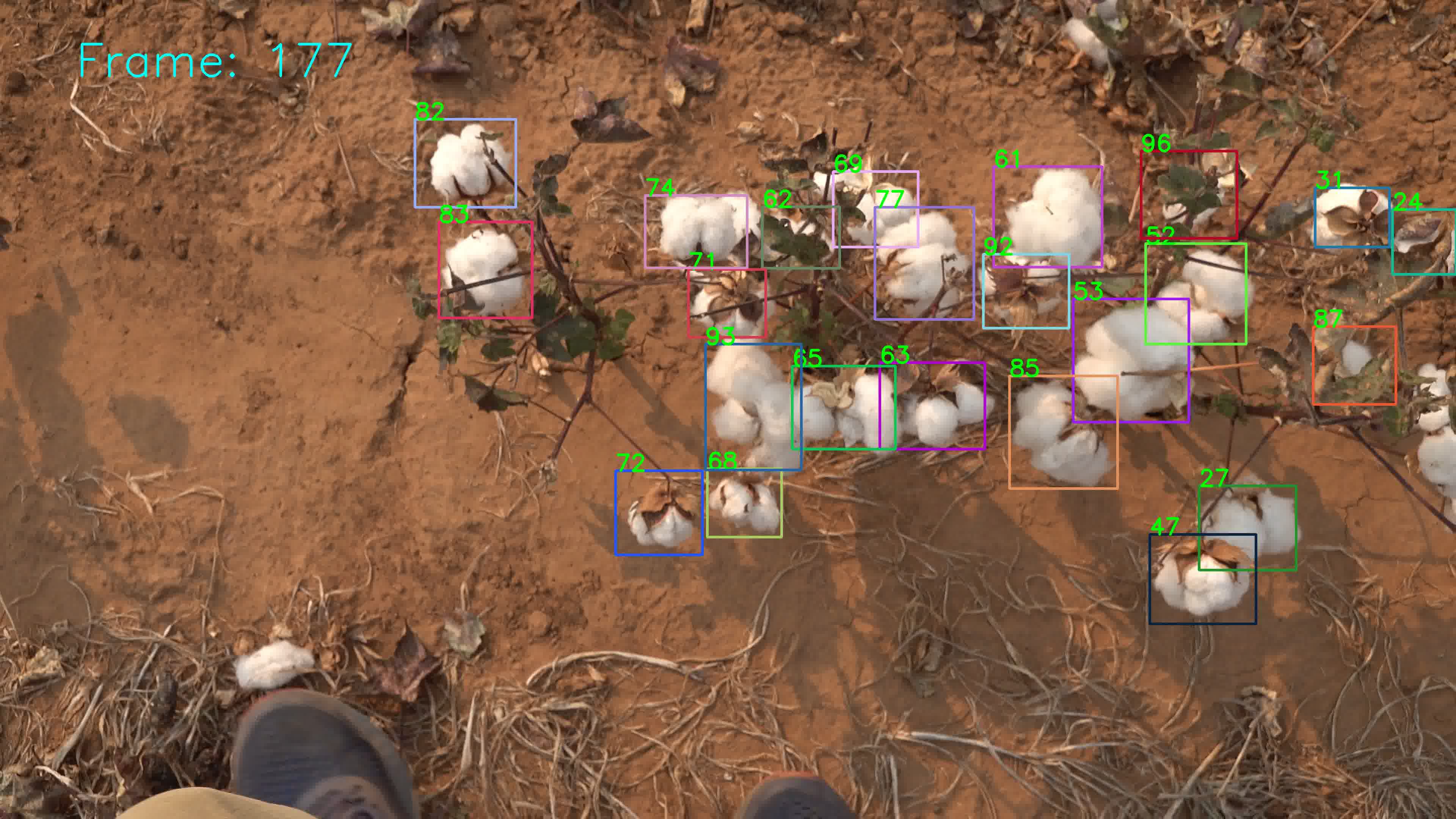}};
    \spy [every spy on node/.append style={ultra thick}] on (1.8,.55) in node [right] at (-1.4,-.7);
  \end{tikzpicture}
\end{subfigure}

\begin{subfigure}[b]{0.32\textwidth}
  \begin{tikzpicture}[spy using outlines={circle, magnification=2.2, size=2.7cm, height=2cm, spy_col, connect spies}]
    \node {\includegraphics[trim=625  200 650 50,clip, width=.99\textwidth]{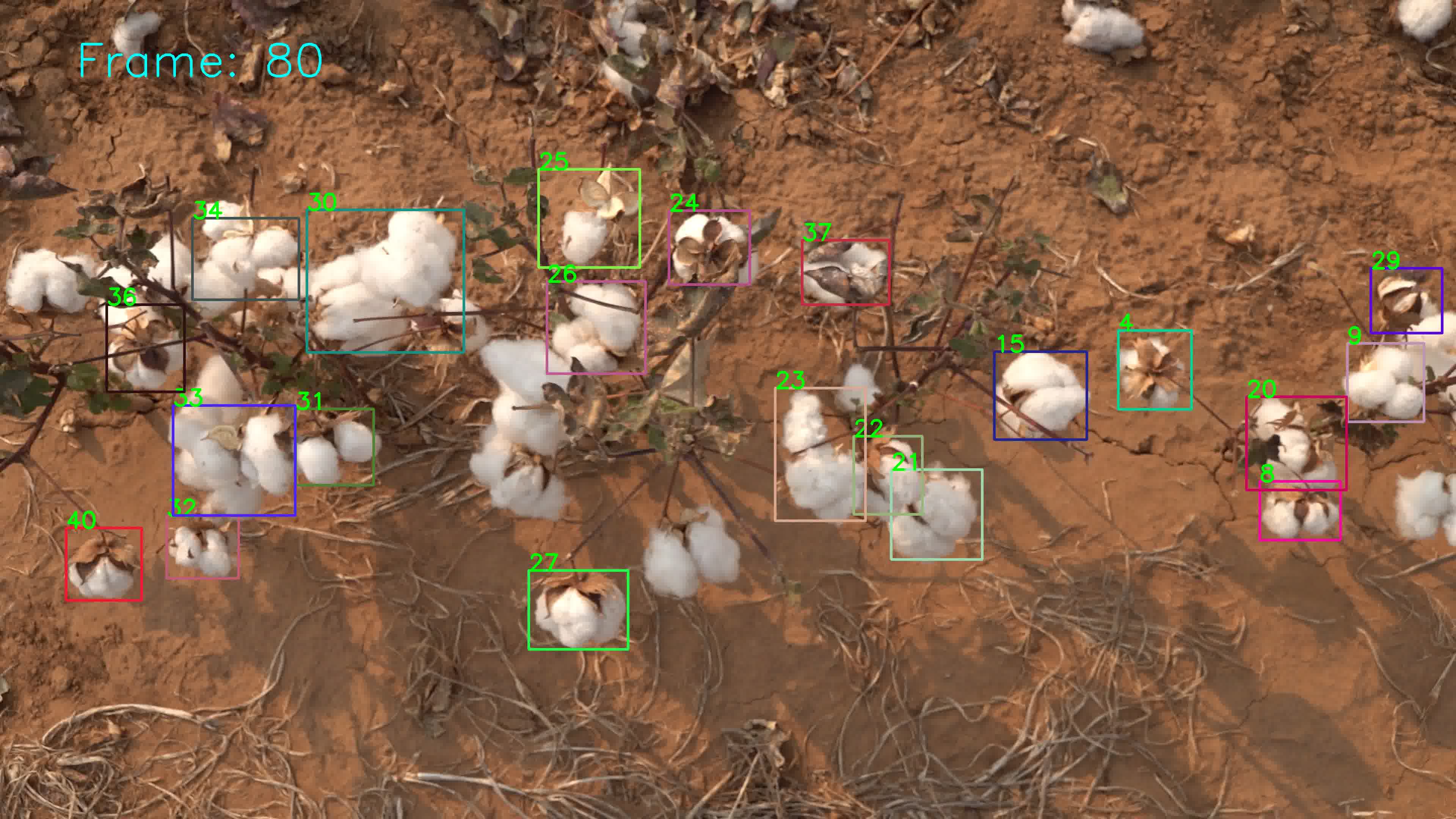}};
      \spy[every spy on node/.append style={ultra thick}] on (-1.65,.32)  in node [right] at  (0,-0.7);
  \end{tikzpicture}
\end{subfigure}
\begin{subfigure}[b]{0.32\textwidth}
  \begin{tikzpicture}[spy using outlines={ circle, magnification=2.5, size=2.7cm, height=2cm, spy_col, connect spies}]
    \node {\includegraphics[trim=625  200 650 50,clip, width=.99\textwidth]{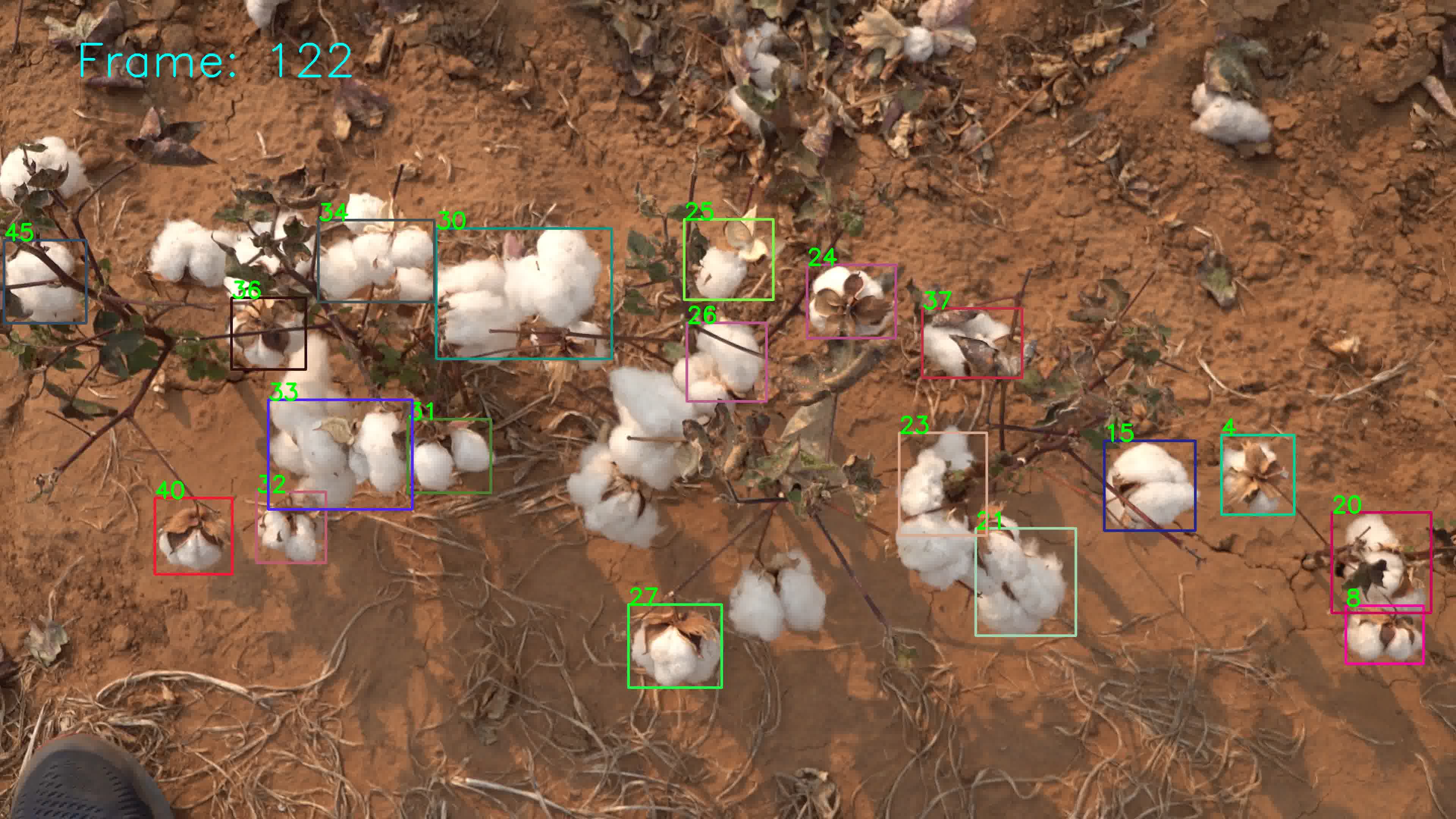}};
     \spy[every spy on node/.append style={ultra thick}] on (-.85,.18) in node [right] at  (.2,-.7);
  \end{tikzpicture}
\end{subfigure}
\begin{subfigure}[b]{0.32\textwidth}
  \begin{tikzpicture}[spy using outlines={ circle, magnification=2.5,size=2.7cm, height=2cm, spy_col, connect spies}]
    \node {\includegraphics[trim=625  200 650 50,clip, width=.99\textwidth]{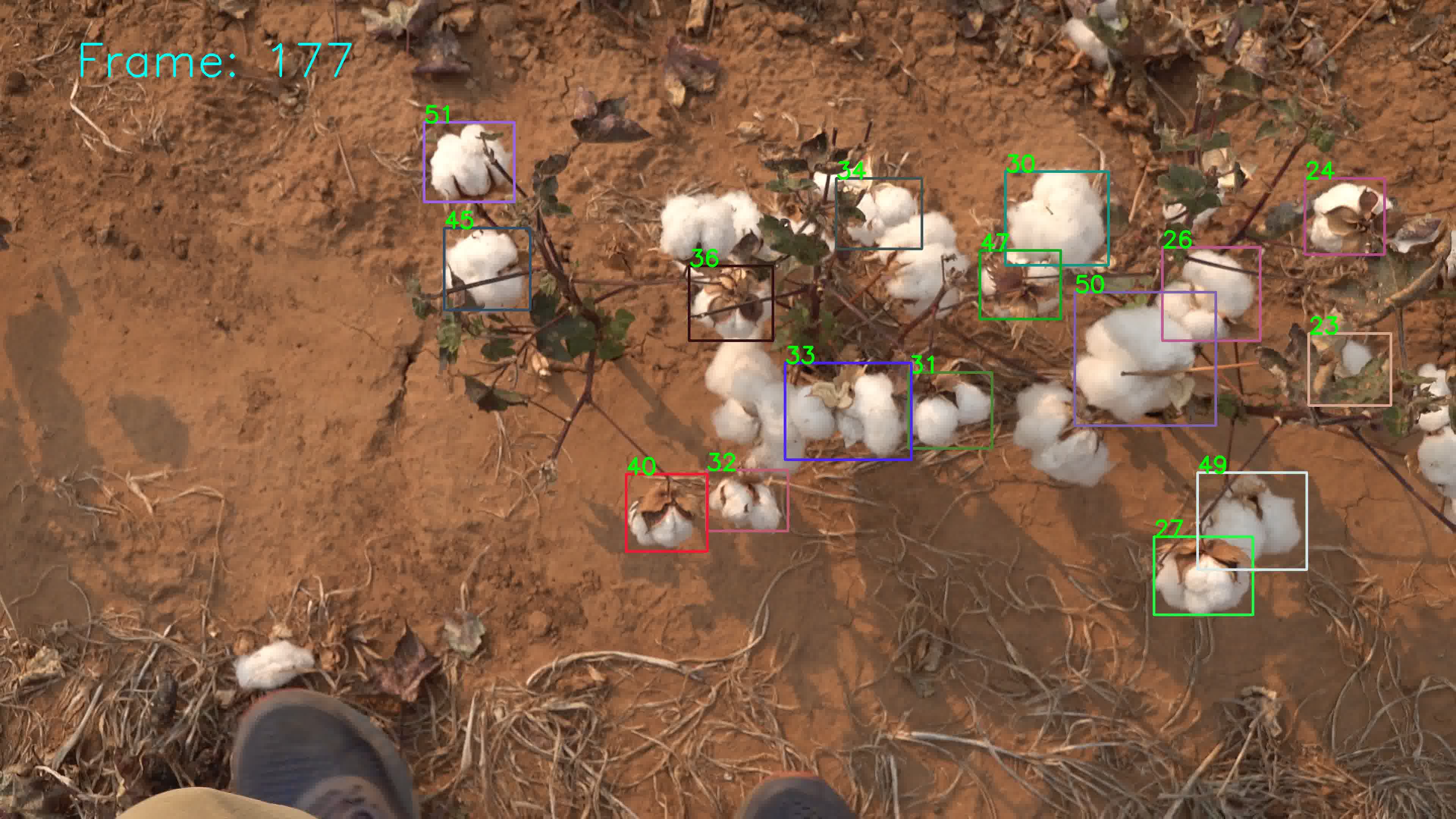}};
    \spy [every spy on node/.append style={ultra thick}] on (1.8,.65) in node [right] at (-1.4,-.7);
  \end{tikzpicture}
\end{subfigure}
\begin{tikzpicture}[overlay, remember picture]
  \draw[green,ultra thick,->] (-16.0, 20.8) to [out=30,in=150] (-10, 20.8); 
  \draw[green,ultra thick,->] (-16.2, 16.4) to [out=30,in=150] (-10, 16.3); 
  \draw[green,ultra thick,->] (-16.2, 12.1) to [out=30,in=150] (-9.8, 12.0); 
  \draw[green,ultra thick,->] (-16.2, 7.5) to [out=30,in=150] (-9.8, 7.4); 
  \draw[red,ultra thick,->] (-16.2, 3) to [out=30,in=150] (-9.8, 2.85); 

   \draw[green,ultra thick,->] (-9.35, 20.8) to [out=30,in=150] (-1.35, 21.1); 
  \draw[red,ultra thick,->] (-9.55, 16.4) to [out=30,in=150] (-1.35, 16.5); 
  \draw[red,ultra thick,->] (-9.55, 12.0) to [out=30,in=150] (-1.35, 12.2);
  \draw[red,ultra thick,->] (-9.55, 7.4) to [out=30,in=150] (-1.35,7.8); 
  \draw[red,ultra thick,->] (-9.55,2.8) to [out=30,in=150] (-1.40,2.95); 
\end{tikzpicture}
\caption{The first scenario qualitative comparison between NTrack and the
competing methods on the \textbf{TexCot22} \cite{muzaddid2023texcot22} dataset.
From top to bottom, each row (NTrack, DeepSORT \cite{wojke2017simple}, Tracktor
\cite{bergmann2019tracking}, ByteTrack\cite{zhang2022bytetrack}, and
Trackformer\cite{meinhardt2022trackformer}) shows the tracking performance of
the different techniques on the same video sequence. The numbers (80, 122, 177)
at the top-left corner of each image portray the frame number in the
corresponding video sequence. Correct and incorrect associations between cotton
bolls are illustrated by the green and red arrows, respectively. The numbers at
the top-left corner of each bounding box report the identity of the associated
cotton boll assigned by the tracker.}
\label{fig:ntrack_vs_other_qualitative_results_1}
\end{figure*}

We conducted a qualitative analysis of NTrack against the trackers in the
evaluation set. The first two rows, from the top of
Fig.~\ref{fig:ntrack_vs_other_qualitative_results_1}, show the tracking
outcomes of NTrack and DeepSORT on a test video sequence. In row 2, the arrow
highlights the track of a specific cotton boll with an ID of 83. DeepSORT fails
to track this boll between frame 122 and frame 177 as indicated by the red
arrow. At frame 177, DeepSORT cannot reidentify the previously seen cotton boll
(ID 83) and it erroneously assigns a new ID (ID 142). Similarly, Tracktor (row
3, ID 57), and ByteTrack (row 4, ID 60) also fail to track the same cotton
boll. TrackFormer (row 5) does not even detect the cotton boll in frames 80
and 122. However, NTrack (row 1) successfully tracks this boll (ID 33) and
assigns the same ID in frame 177.

\begin{figure*}
\centering
\begin{subfigure}[b]{0.32\textwidth}
  \begin{tikzpicture}[spy using outlines={ circle, magnification=2.5, size=2.7cm, spy_col, connect spies}]
    \node {\includegraphics[trim=625  550 850 50,clip, width=.99\textwidth]{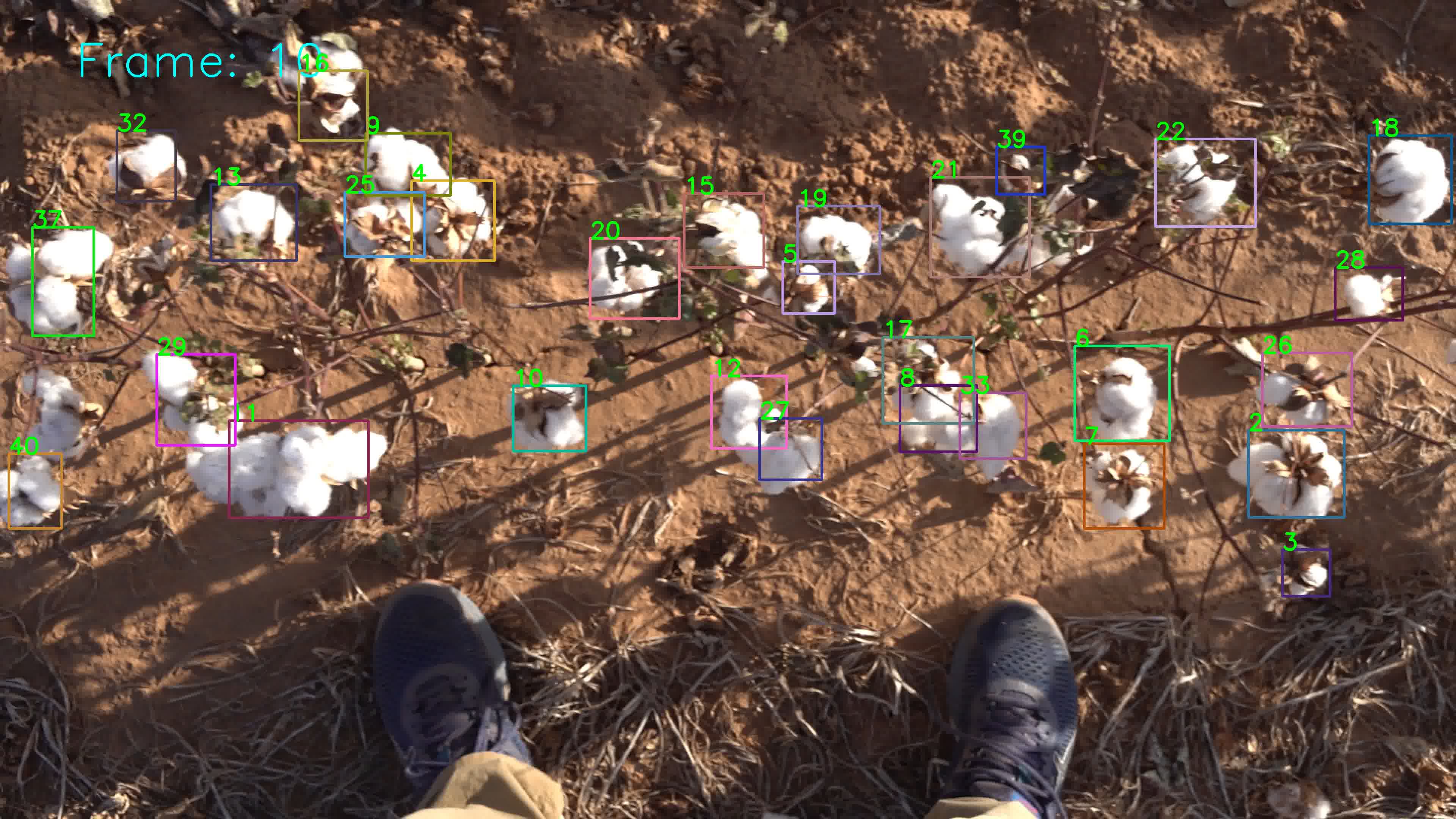}};
    \spy[every spy on node/.append style={ultra thick}] on (-1.7,.9)  in node [right] at  (0,-.4);
  \end{tikzpicture}
\end{subfigure}
\begin{subfigure}[b]{0.32\textwidth}
  \begin{tikzpicture}[spy using outlines={ circle, magnification=2.5, size=2.7cm, spy_col, connect spies}]
    \node {\includegraphics[trim=625  550 850 50,clip, width=.99\textwidth]{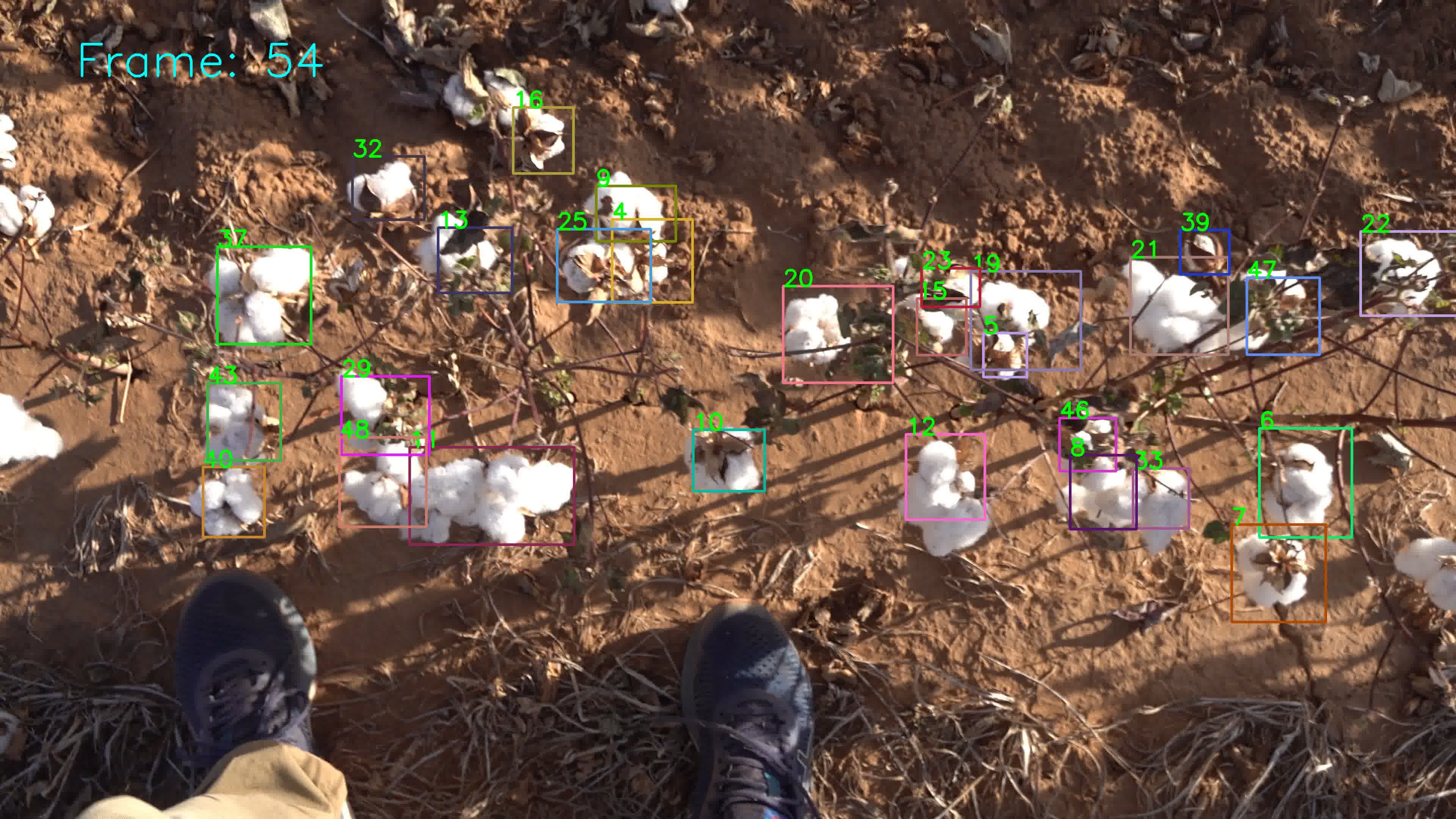}};
    \spy[every spy on node/.append style={ultra thick}] on (-.4,.55) in node [right] at  (0,-.4);
  \end{tikzpicture}
\end{subfigure}
\begin{subfigure}[b]{0.32\textwidth}
  \begin{tikzpicture}[spy using outlines={ circle, magnification=2.5, size=2.7cm, height=2cm, spy_col, connect spies}]
    \node {\includegraphics[trim=625  550 850 50,clip, width=.99\textwidth]{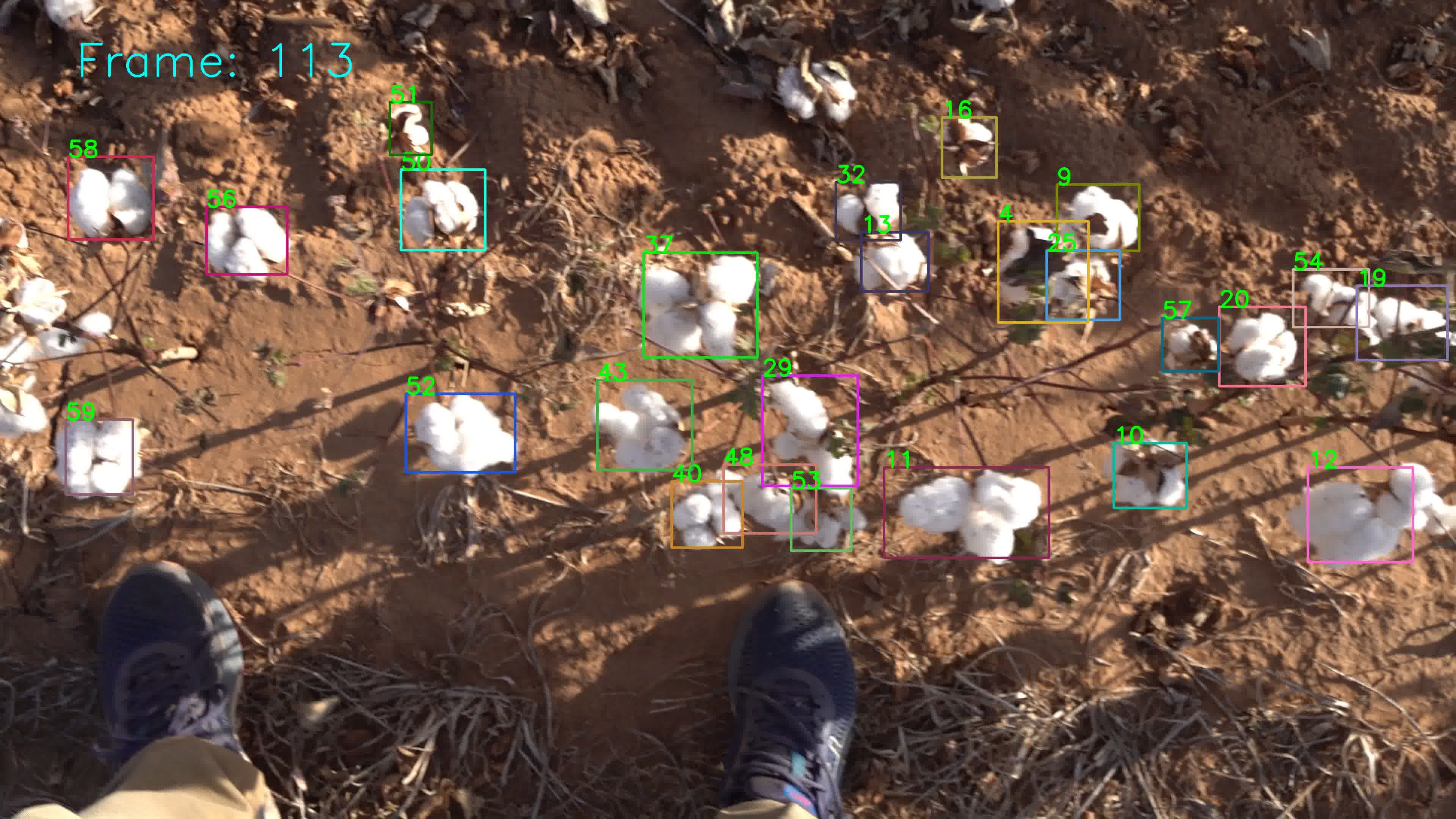}};
    \spy [every spy on node/.append style={ultra thick}] on (2.3,.5) in node [right] at (-1,-.4);
  \end{tikzpicture}
\end{subfigure}

\begin{subfigure}[b]{0.32\textwidth}
  \begin{tikzpicture}[spy using outlines={ circle, magnification=2.5, size=2.7cm, spy_col, connect spies}]
    \node {\includegraphics[trim=625  550 850 50,clip, width=.99\textwidth]{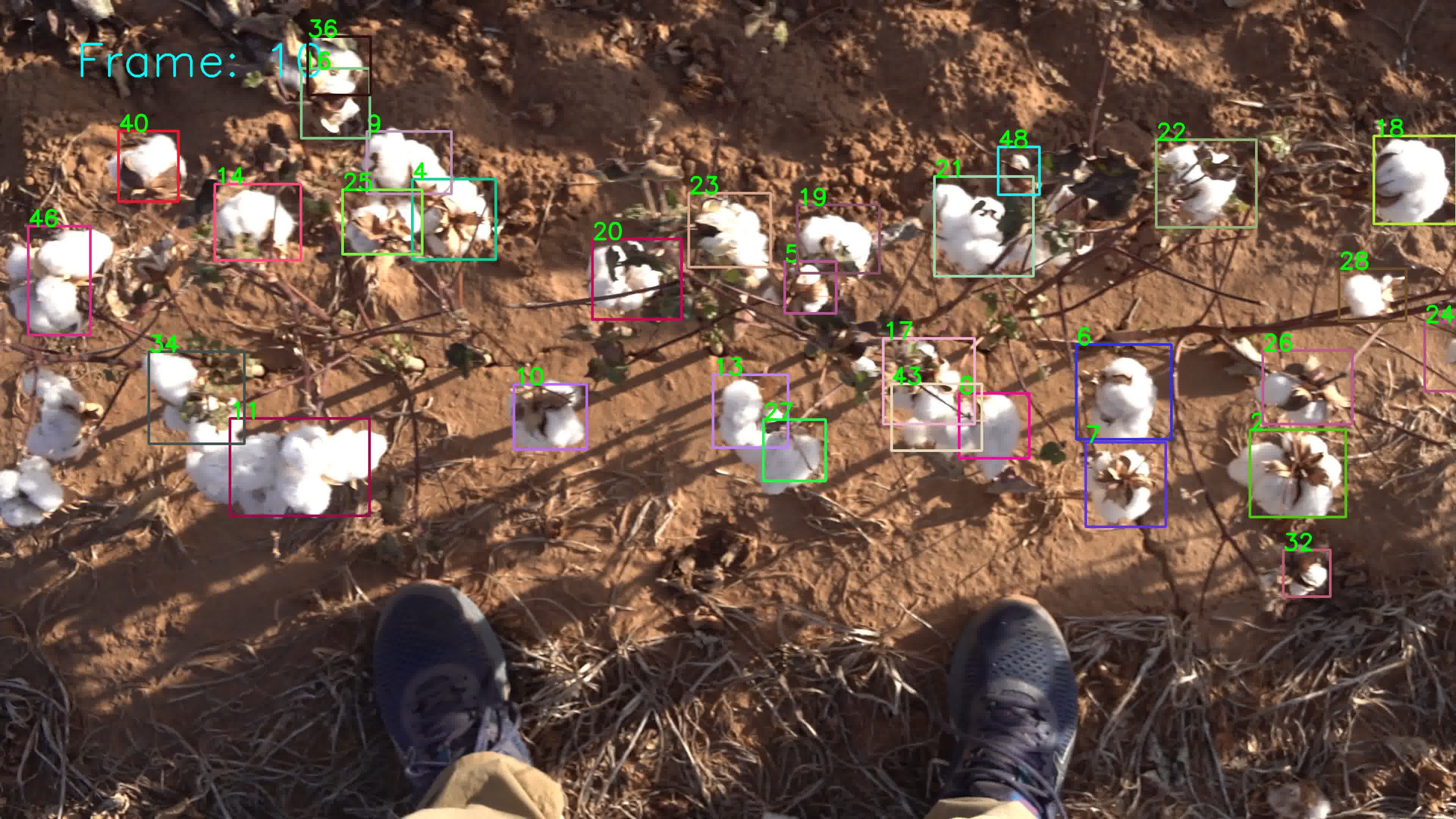}};
    \spy[every spy on node/.append style={ultra thick}] on (-1.7,.9)  in node [right] at  (0,-.4);
  \end{tikzpicture}
\end{subfigure}
\begin{subfigure}[b]{0.32\textwidth}
  \begin{tikzpicture}[spy using outlines={ circle, magnification=2.5, size=2.7cm, spy_col, connect spies}]
    \node {\includegraphics[trim=625  550 850 50,clip, width=.99\textwidth]{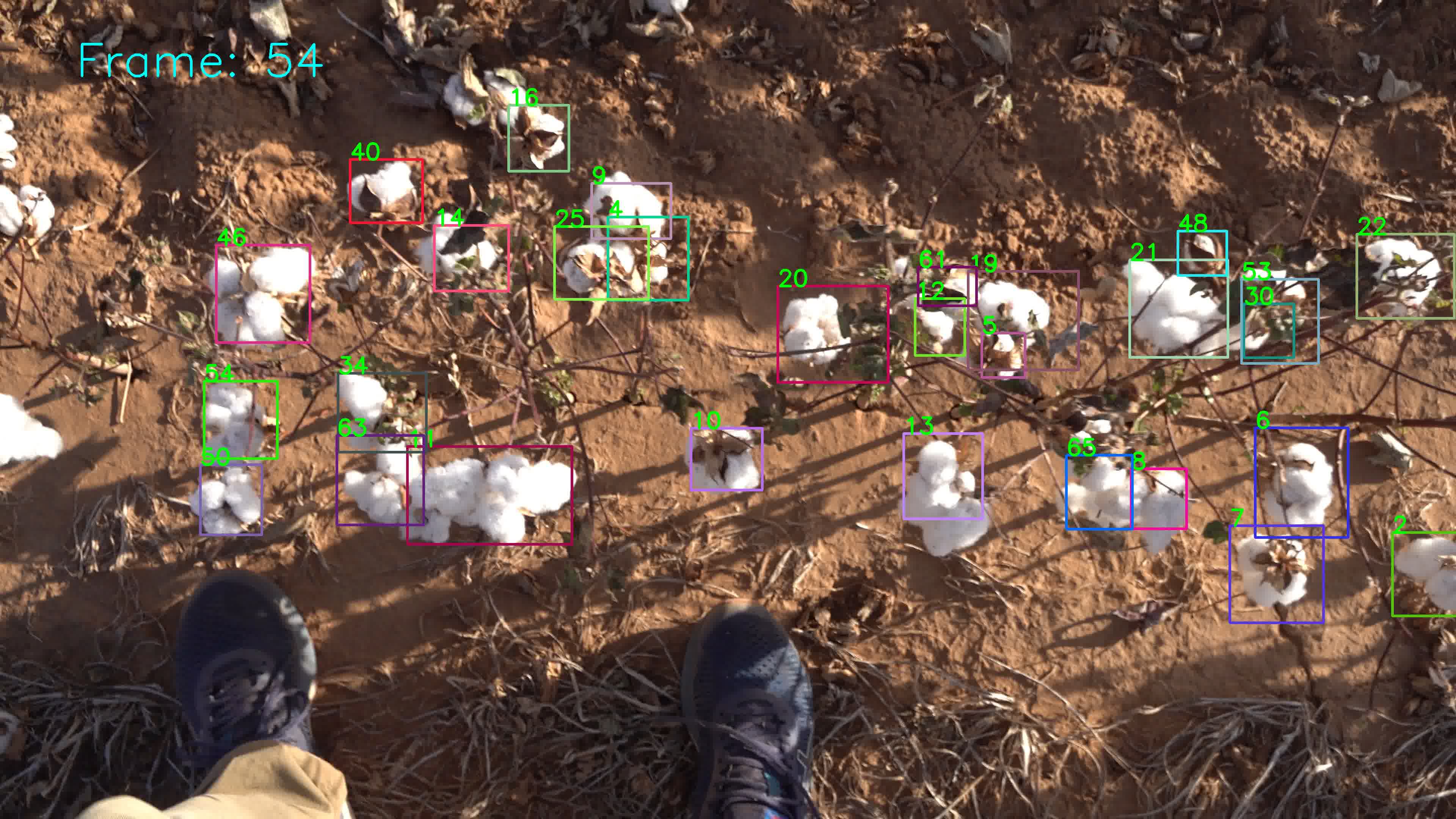}};
    \spy[every spy on node/.append style={ultra thick}] on (-.4,.55) in node [right] at  (0,-.4);
  \end{tikzpicture}
\end{subfigure}
\begin{subfigure}[b]{0.32\textwidth}
  \begin{tikzpicture}[spy using outlines={ circle, magnification=2.5, size=2.7cm, height=2cm, spy_col, connect spies}]
    \node {\includegraphics[trim=625  550 850 50,clip, width=.99\textwidth]{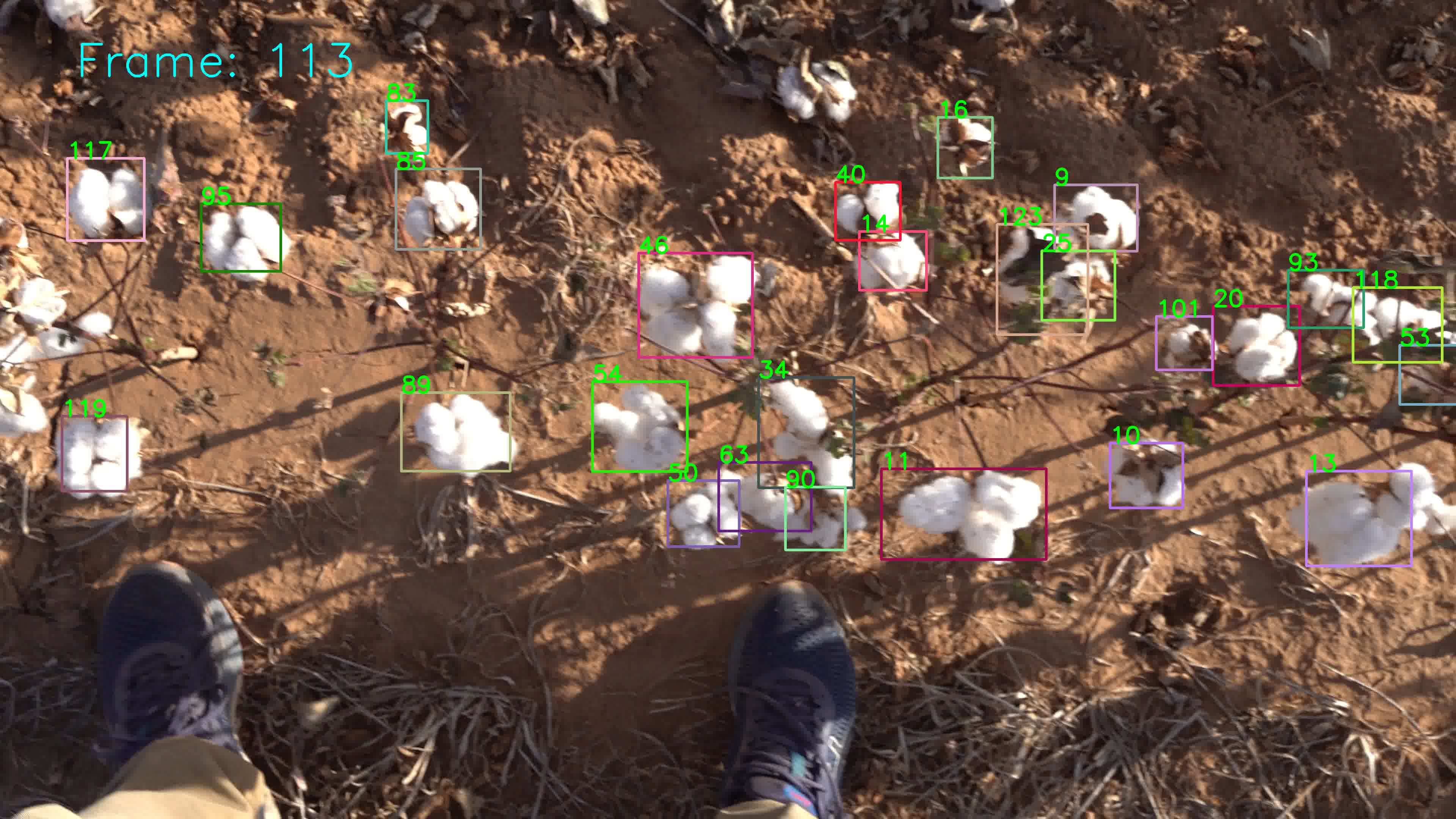}};
    \spy [every spy on node/.append style={ultra thick}] on (2.3,.5) in node [right] at (-1,-.4);
  \end{tikzpicture}
\end{subfigure}

\begin{subfigure}[b]{0.32\textwidth}
  \begin{tikzpicture}[spy using outlines={ circle, magnification=2.5, size=2.7cm, spy_col, connect spies}]
    \node {\includegraphics[trim=625  550 850 50,clip, width=.99\textwidth]{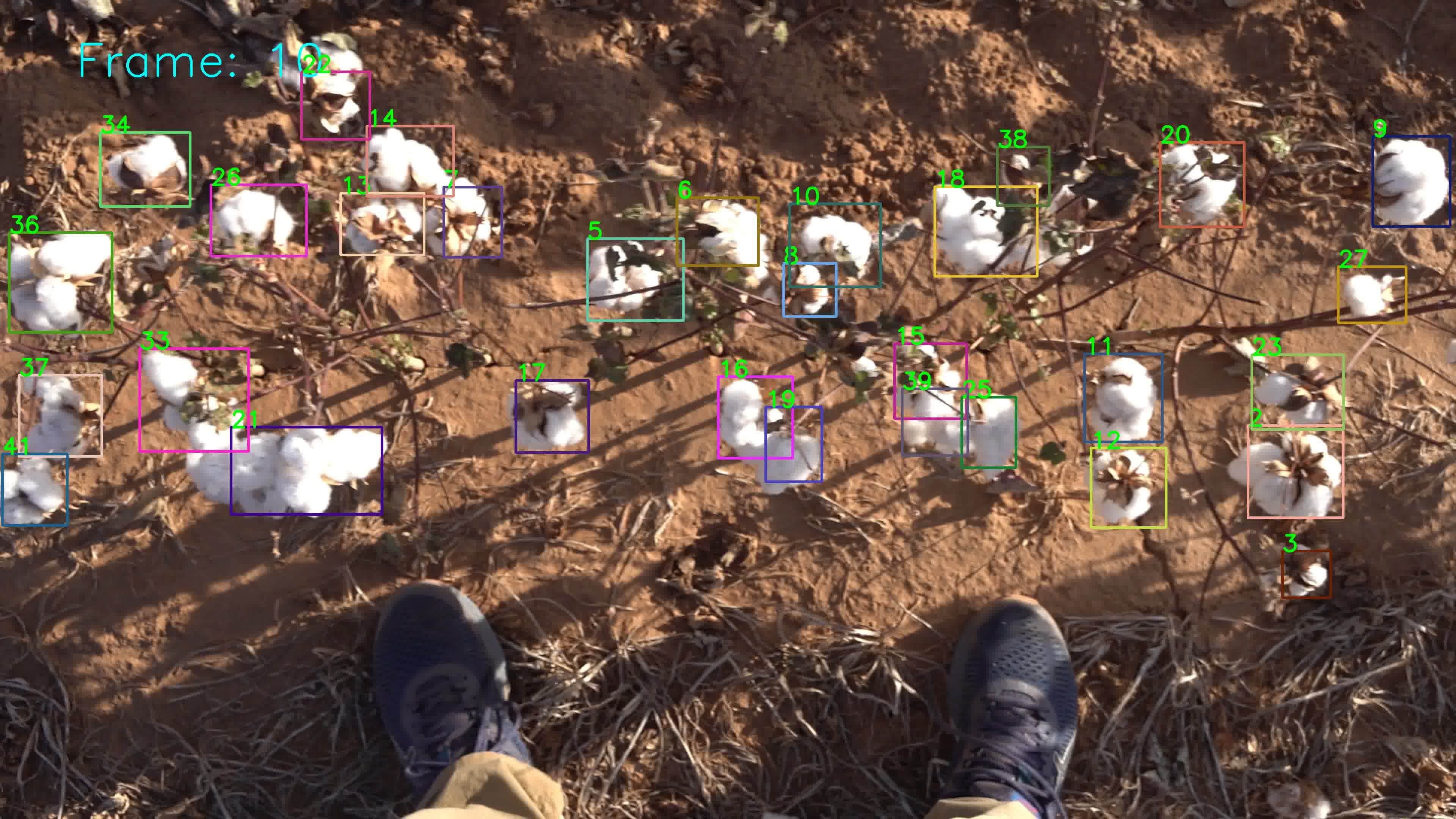}};
    \spy[every spy on node/.append style={ultra thick}] on (-1.7,.9)  in node [right] at  (0,-.4);
  \end{tikzpicture}
\end{subfigure}
\begin{subfigure}[b]{0.32\textwidth}
  \begin{tikzpicture}[spy using outlines={ circle, magnification=2.5, size=2.7cm, spy_col, connect spies}]
    \node {\includegraphics[trim=625  550 850 50,clip, width=.99\textwidth]{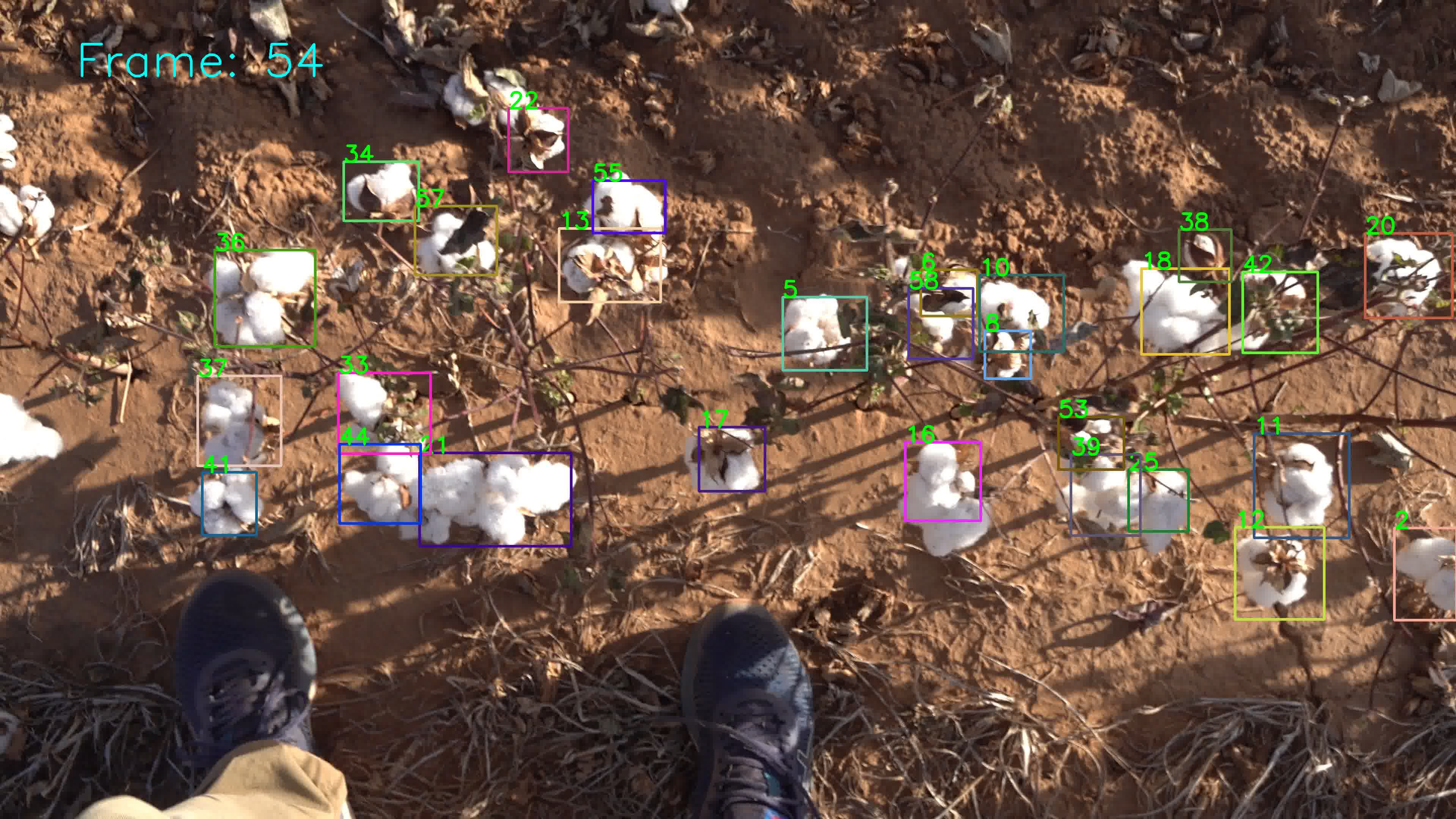}};
    \spy[every spy on node/.append style={ultra thick}] on (-.4,.55) in node [right] at  (0,-.4);
  \end{tikzpicture}
\end{subfigure}
\begin{subfigure}[b]{0.32\textwidth}
  \begin{tikzpicture}[spy using outlines={ circle, magnification=2.5, size=2.7cm, height=2cm, spy_col, connect spies}]
    \node {\includegraphics[trim=625  550 850 50,clip, width=.99\textwidth]{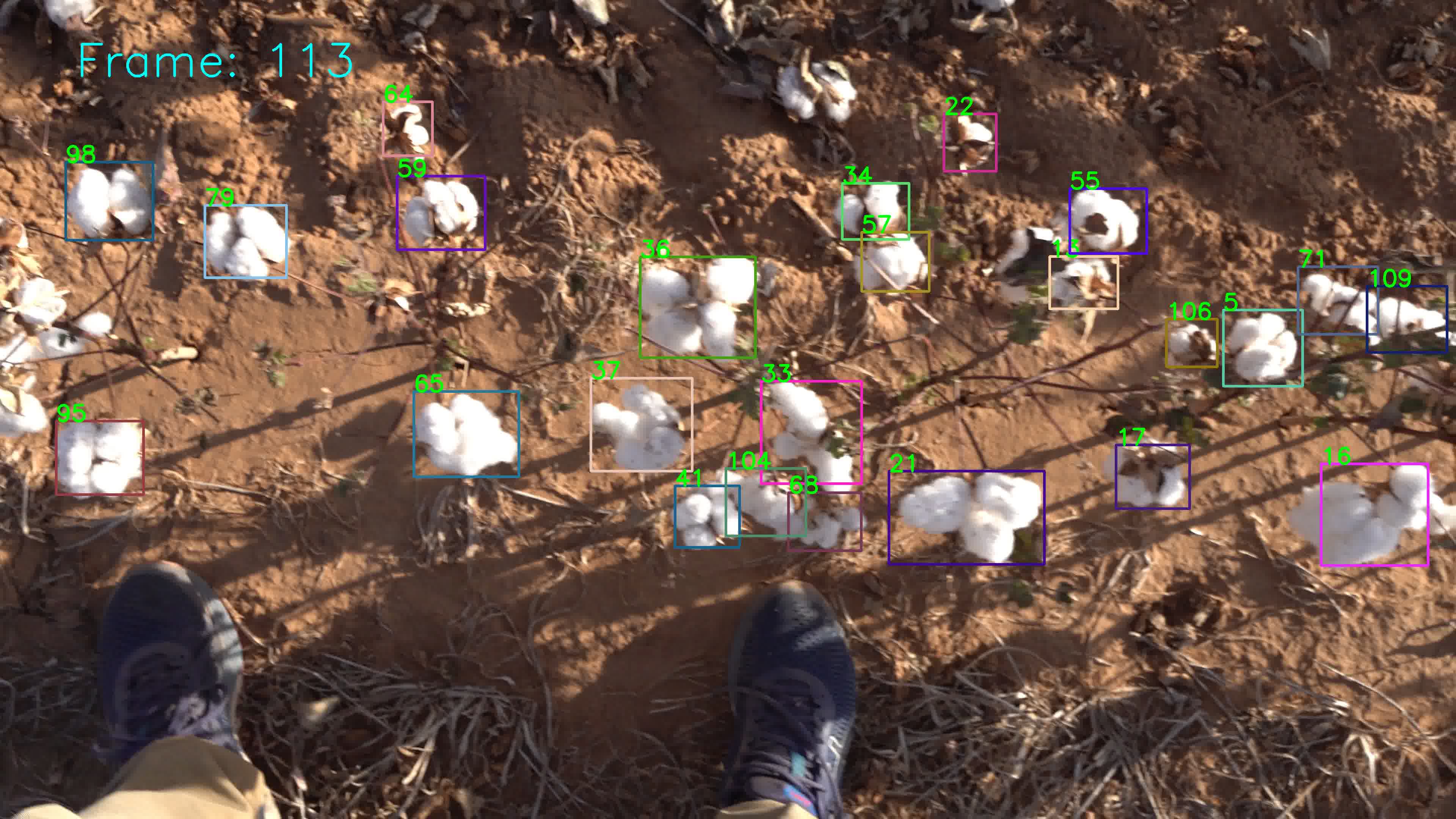}};
    \spy [every spy on node/.append style={ultra thick}] on (2.3,.5) in node [right] at (-1,-.4);
  \end{tikzpicture}
\end{subfigure}

\begin{subfigure}[b]{0.32\textwidth}
  \begin{tikzpicture}[spy using outlines={ circle, magnification=2.5, size=2.7cm, spy_col, connect spies}]
    \node {\includegraphics[trim=625  550 850 50,clip, width=.99\textwidth]{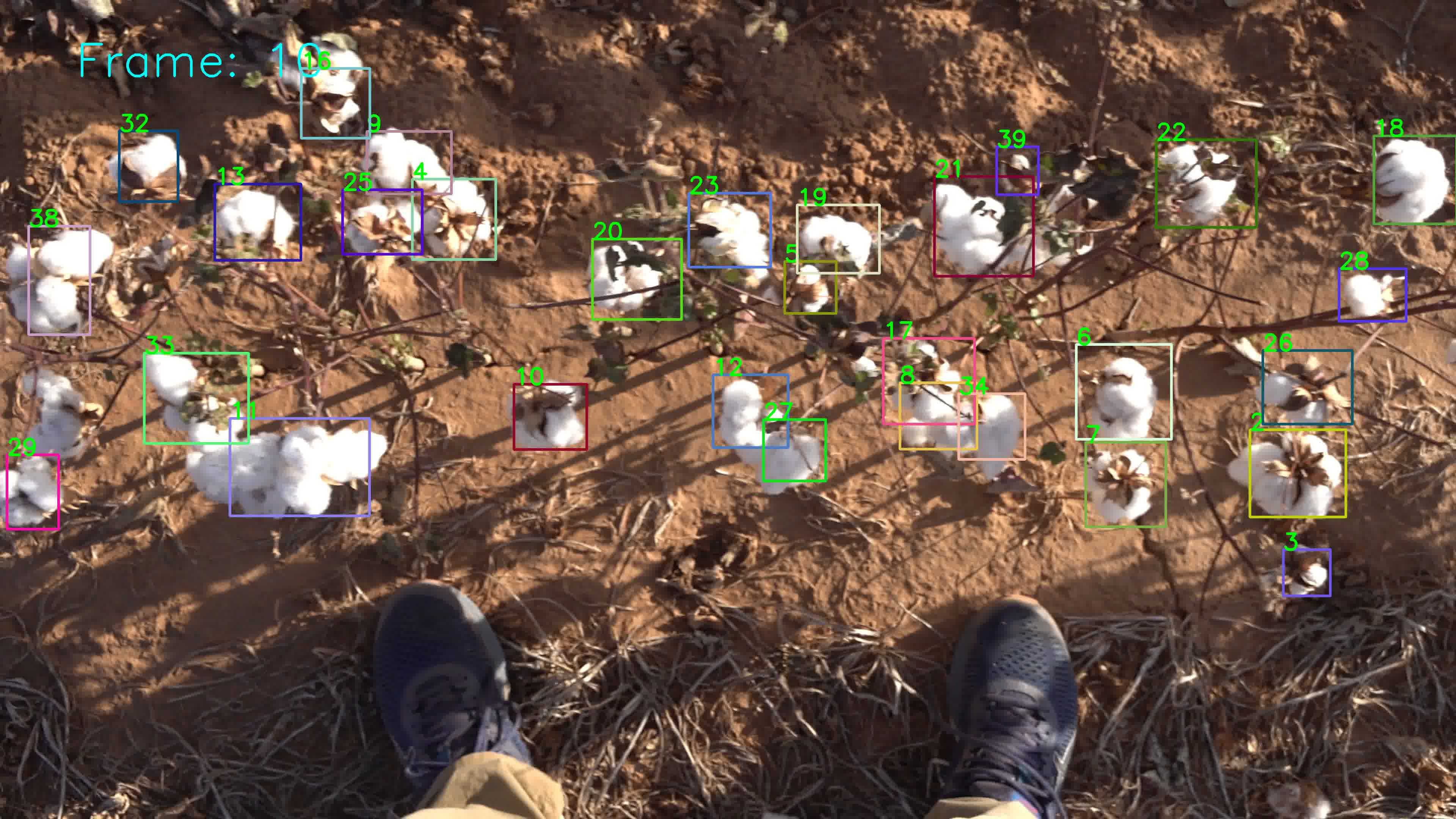}};
    \spy[every spy on node/.append style={ultra thick}] on (-1.7,.85)  in node [right] at  (0,-.4);
  \end{tikzpicture}
\end{subfigure}
\begin{subfigure}[b]{0.32\textwidth}
  \begin{tikzpicture}[spy using outlines={ circle, magnification=2.5, size=2.7cm, spy_col, connect spies}]
    \node {\includegraphics[trim=625  550 850 50,clip, width=.99\textwidth]{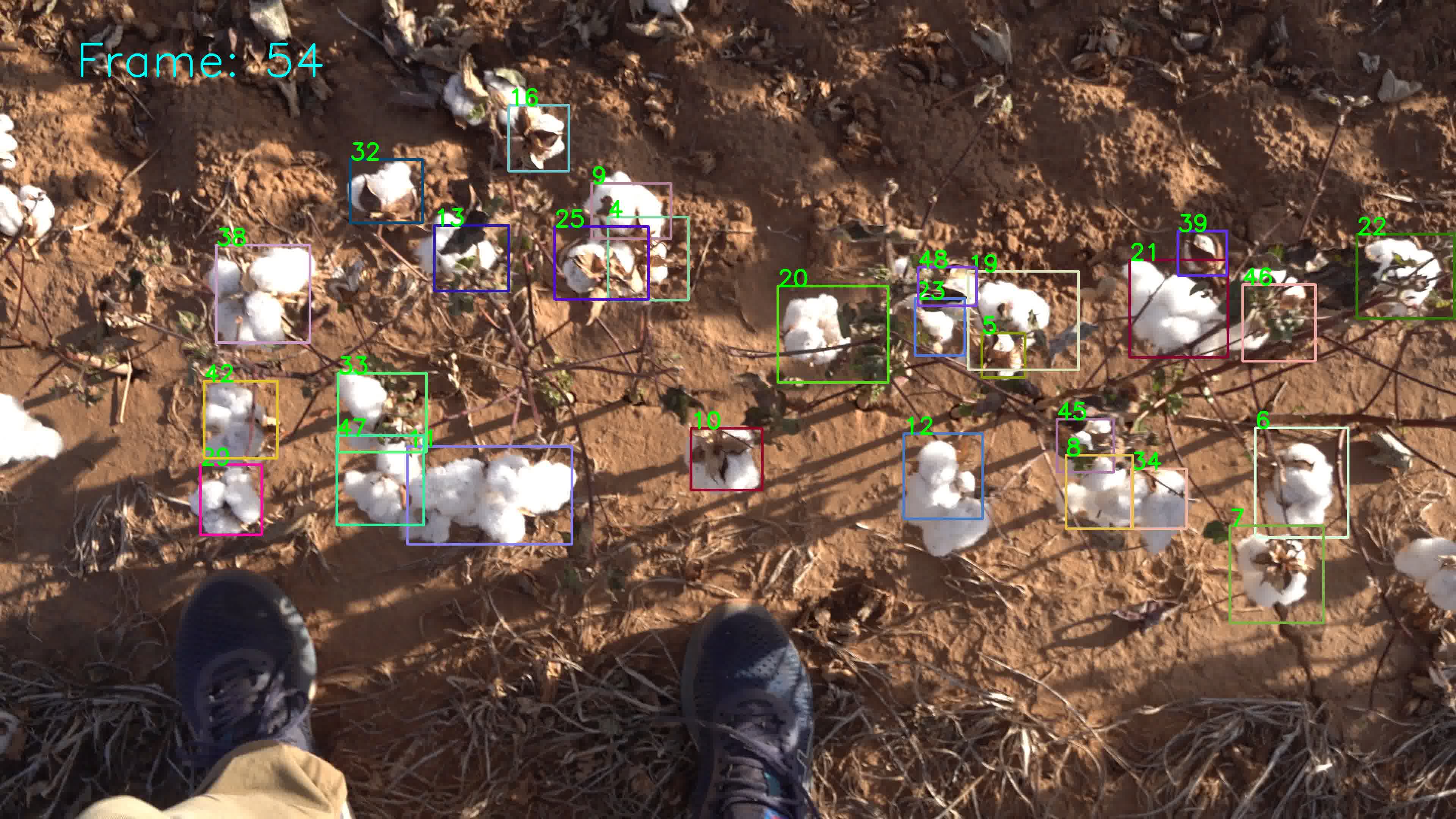}};
    \spy[every spy on node/.append style={ultra thick}] on (-.5,.57) in node [right] at  (0,-.4);
  \end{tikzpicture}
\end{subfigure}
\begin{subfigure}[b]{0.32\textwidth}
  \begin{tikzpicture}[spy using outlines={ circle, magnification=2.5, size=2.7cm, height=2cm, spy_col, connect spies}]
    \node {\includegraphics[trim=625  550 850 50,clip, width=.99\textwidth]{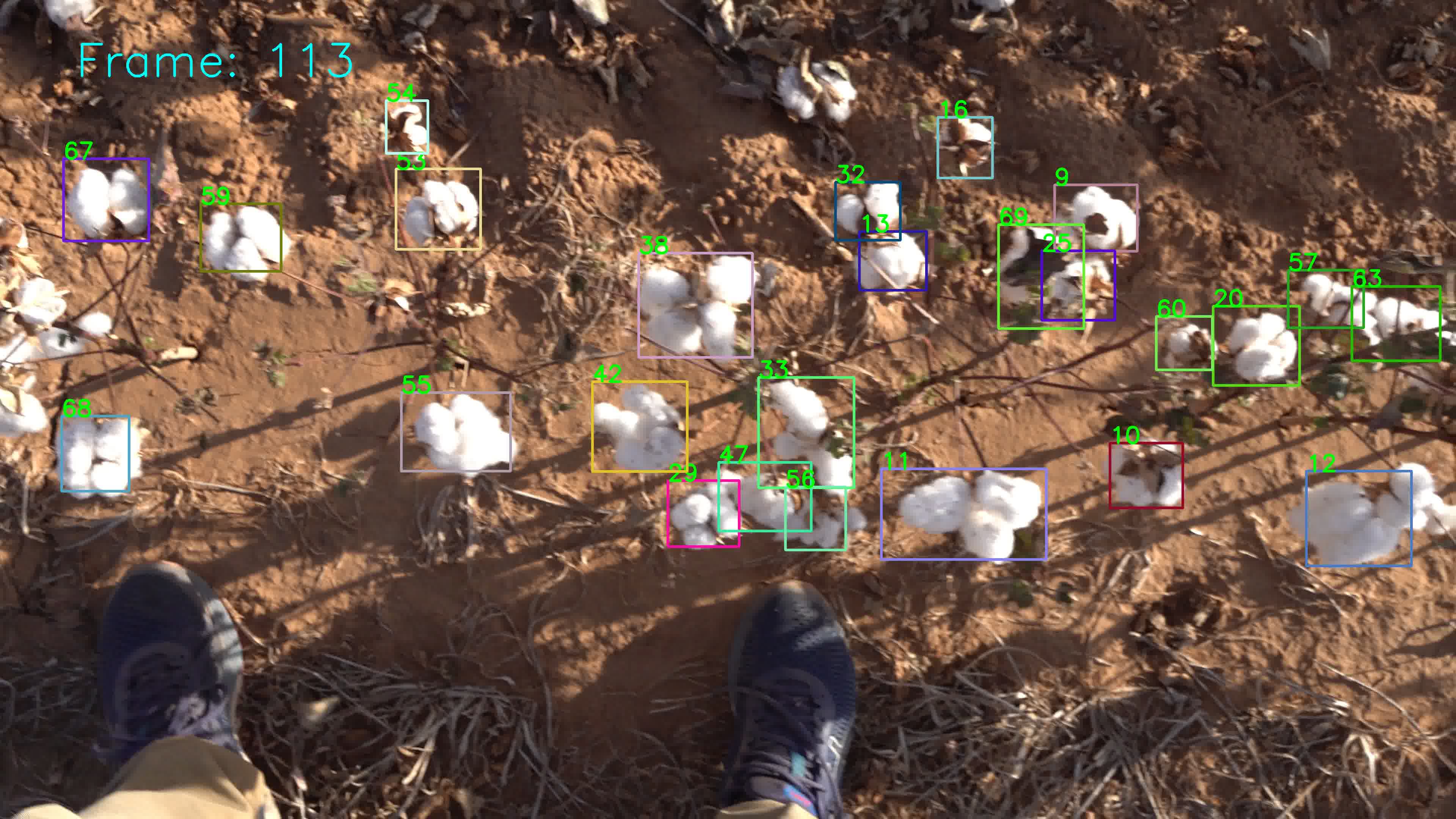}};
    \spy [every spy on node/.append style={ultra thick}] on (2.3,.5) in node [right] at (-1,-.4);
  \end{tikzpicture}
\end{subfigure}

\begin{subfigure}[b]{0.32\textwidth}
  \begin{tikzpicture}[spy using outlines={ circle, magnification=2.5, size=2.7cm, spy_col, connect spies}]
    \node {\includegraphics[trim=625  550 850 50,clip, width=.99\textwidth]{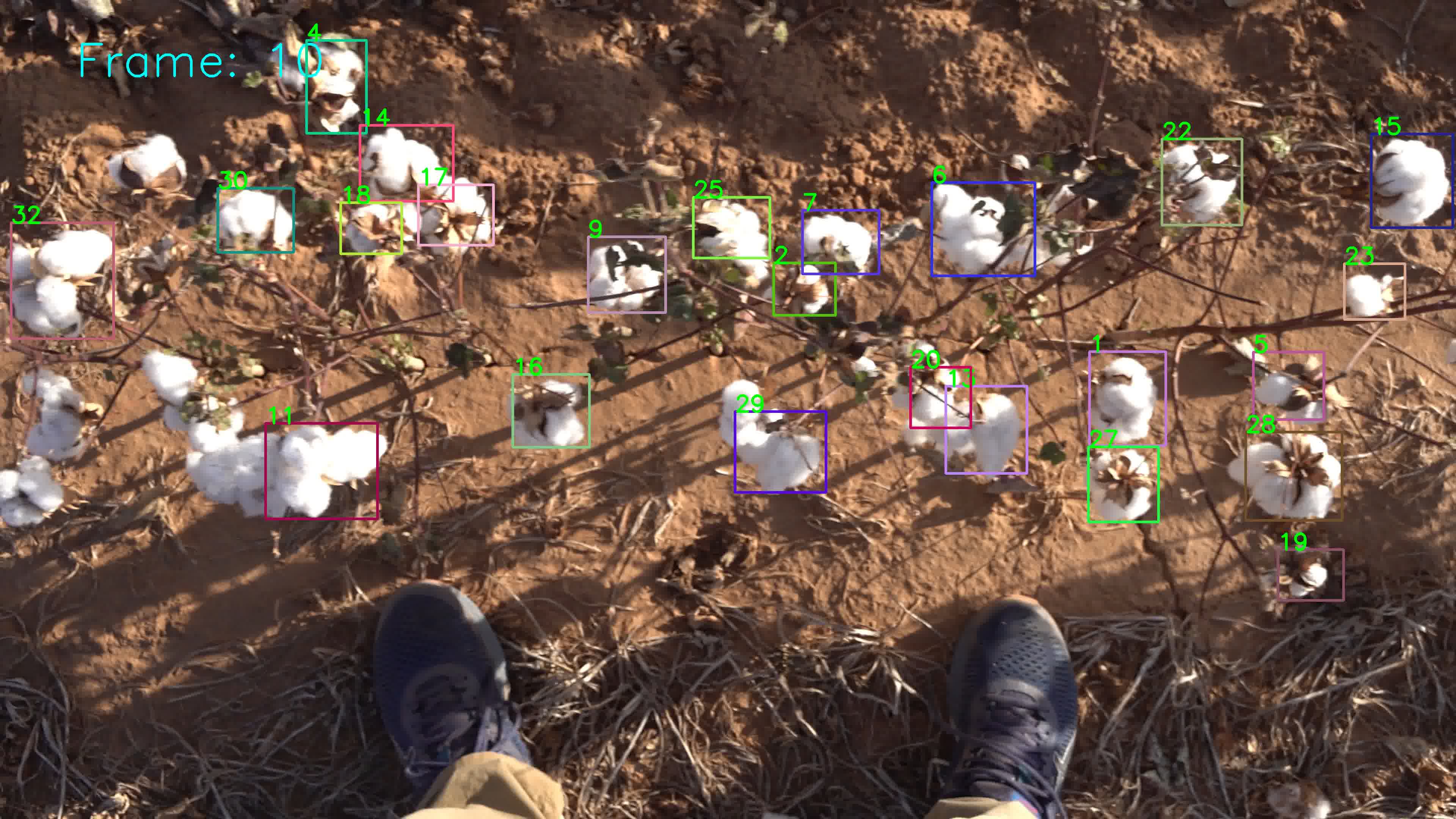}};
        \spy[every spy on node/.append style={ultra thick}] on (-1.8,.9)  in node [right] at  (0,-.4);
  \end{tikzpicture}
\end{subfigure}
\begin{subfigure}[b]{0.32\textwidth}
  \begin{tikzpicture}[spy using outlines={ circle, magnification=2.5, size=2.7cm, spy_col, connect spies}]
    \node {\includegraphics[trim=625  550 850 50,clip, width=.99\textwidth]{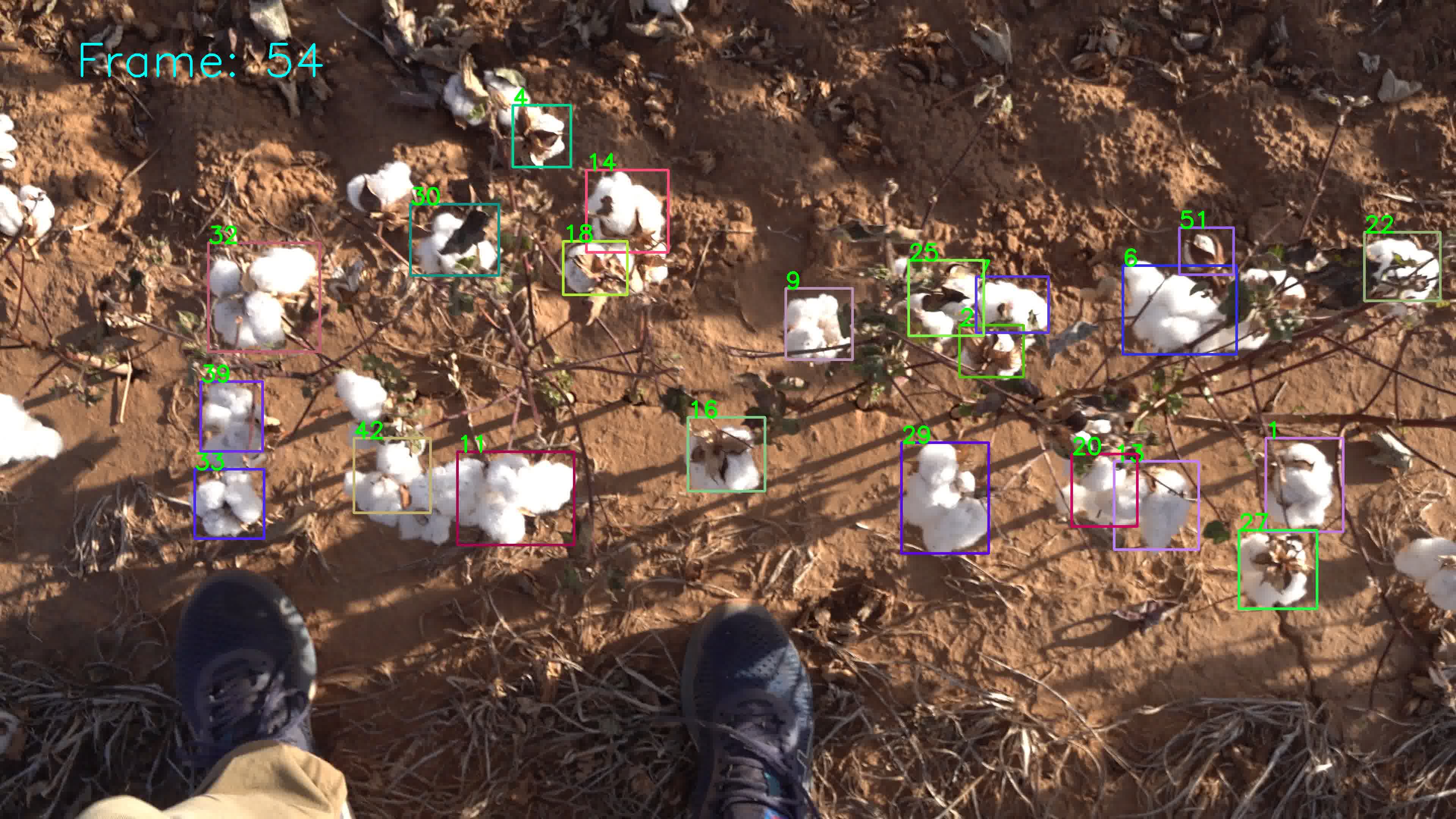}};
     \spy[every spy on node/.append style={ultra thick}] on (-.5,.57) in node [right] at  (0,-.4);
  \end{tikzpicture}
\end{subfigure}
\begin{subfigure}[b]{0.32\textwidth}
  \begin{tikzpicture}[spy using outlines={ circle, magnification=2.5, size=2.7cm, spy_col, connect spies}]
    \node {\includegraphics[trim=625  550 850 50,clip, width=.99\textwidth]{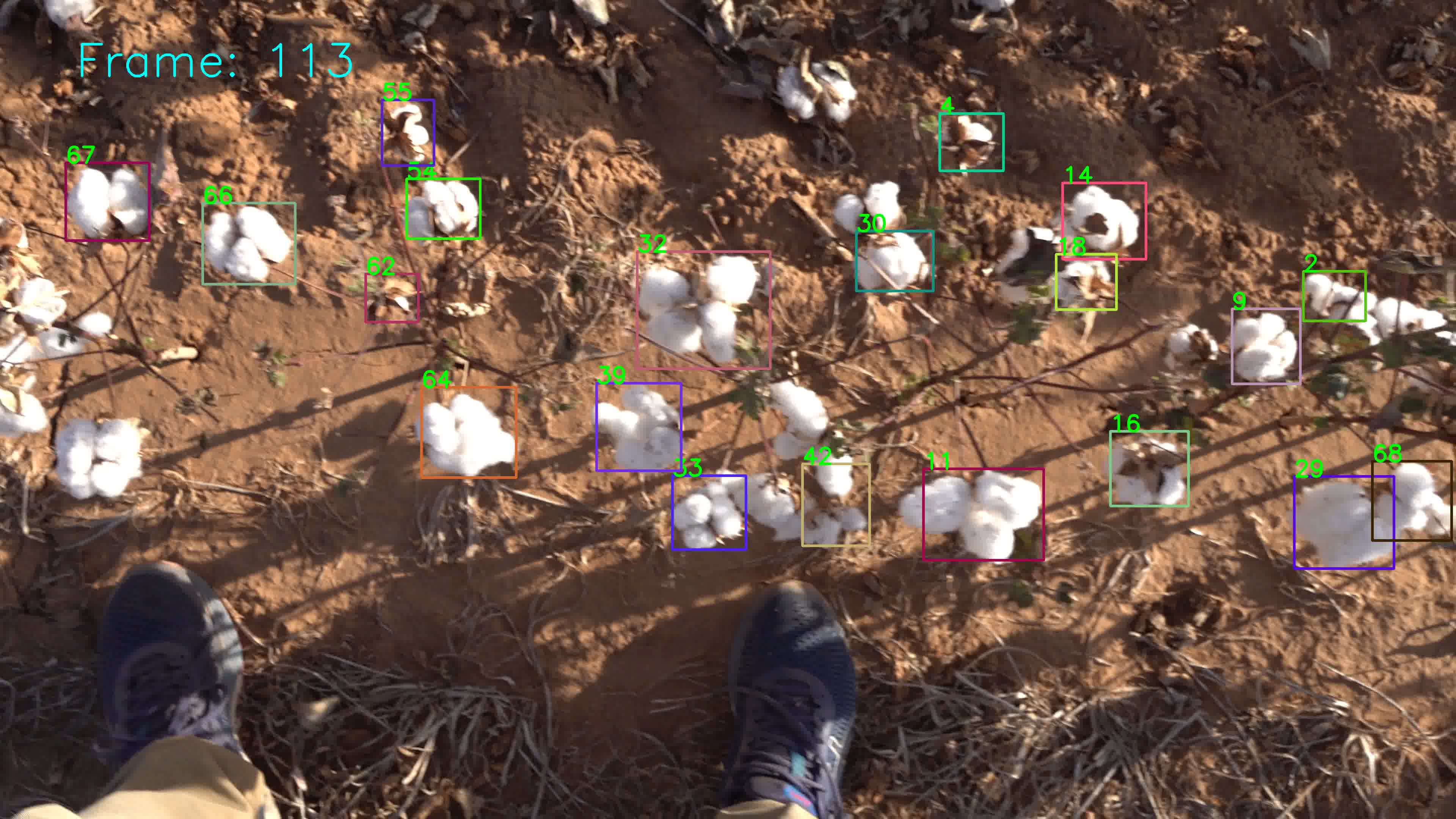}};
    \spy [every spy on node/.append style={ultra thick}] on (2.3,.5) in node [right] at (-1,-.4);
  \end{tikzpicture}
\end{subfigure}
\begin{tikzpicture}[overlay, remember picture]
  \draw[green,ultra thick,->] (-16.1, 19.3) to [out=20,in=160] (-9.6, 19.0);
  \draw[green,ultra thick,->] (-16.1, 15.3) to [out=20,in=160] (-9.6, 15.0); 
  \draw[red,ultra thick,->] (-16.1, 11.3) to [out=20,in=160] (-9.6, 10.9); 
  \draw[green,ultra thick,->] (-16.2, 7.2) to [out=20,in=160] (-9.6, 6.7); 
  \draw[red,ultra thick,->] (-16.2, 3.2) to [out=20,in=160] (-9.6, 2.95); 

  \draw[green,ultra thick,->] (-9.0, 19.1) to [out=20,in=160] (-.9,19.0); 
   \draw[red,ultra thick,->] (-9.0, 15.0) to [out=20,in=160] (-.9,15.0); 
  \draw[red,ultra thick,->] (-9.0, 10.9) to [out=20,in=160] (-.9,10.9); 
  \draw[red,ultra thick,->] (-9.0, 6.7) to [out=20,in=160] (-.9,6.7); 
  \draw[red,ultra thick,->] (-9.1, 3.0) to [out=20,in=160] (-1.0,2.9); 
\end{tikzpicture}
\caption{The second scenario qualitative comparison between NTrack and the
competing methods on the \textbf{TexCot22} \cite{muzaddid2023texcot22} dataset.
From top to bottom, each row (NTrack, DeepSORT \cite{wojke2017simple}, Tracktor
\cite{bergmann2019tracking}, ByteTrack\cite{zhang2022bytetrack}, and
Trackformer\cite{meinhardt2022trackformer}) shows the tracking performance of
the different techniques on the same video sequence. The numbers (10, 54, 133)
at the top-left corner of each image portray the frame number in the
corresponding video sequence. Correct and incorrect associations between cotton
bolls are illustrated by the green and red arrows, respectively.  The numbers
at the top-left corner of each bounding box report the identity of the
associated cotton boll assigned by the tracker.}
\label{fig:ntrack_vs_other_qualitative_results_2}
\end{figure*}

Fig.~\ref{fig:ntrack_vs_other_qualitative_results_2} shows additional
qualitative tracking results in a more challenging scenario involving wind. In
this scene, a cotton boll (ID 4 in rows 1, 2, 4, ID 7 in row 3, ID 17 in row 5)
is occluded for an extended period after frame 54. When the cotton boll
reappears in frame 113, it is successfully reidentified by NTrack (row 1). In
contrast, DeepSORT and ByteTrack assign a new ID (ID 123 and ID 69 in rows 2
and 4, respectively) to the boll. Tracktor and TrackFormer are unable to detect
the cotton boll (rows 3 and 5, respecitively). These observations demonstrate
the effectiveness of our tracking system in reidentifying cotton bolls,
especially in the field under harsh environmental conditions.

\begin{figure*}
\centering
\begin{subfigure}[b]{0.24\textwidth}
  \includegraphics[width=.99\columnwidth, height=22mm]{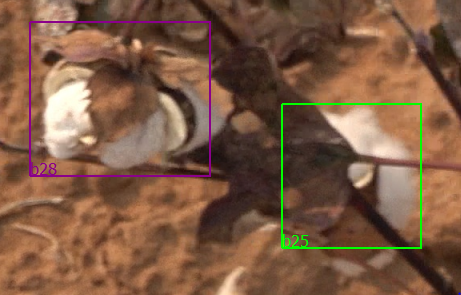}
\end{subfigure}
\vspace{3px}
\begin{subfigure}[b]{0.24\textwidth}
  \includegraphics[width=.99\columnwidth, height=22mm]{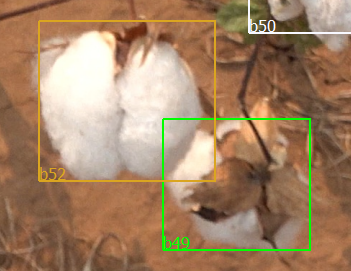}
\end{subfigure}
\begin{subfigure}[b]{0.24\textwidth}
  \includegraphics[width=.99\columnwidth, height=22mm]{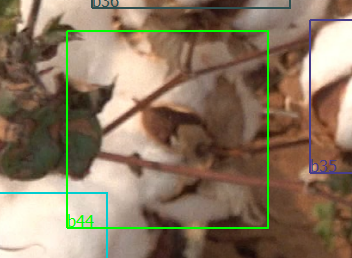}
\end{subfigure}
\begin{subfigure}[b]{0.24\textwidth}
  \includegraphics[width=.99\columnwidth, height=22mm]{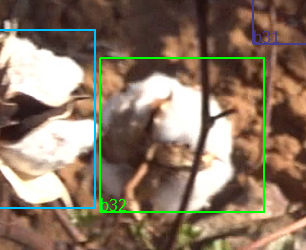}
\end{subfigure}
\begin{subfigure}[b]{0.24\textwidth}
  \includegraphics[width=.99\columnwidth, height=22mm]{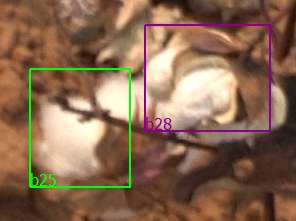}
  \caption{Similarity score: 0.015}
\end{subfigure}
\begin{subfigure}[b]{0.24\textwidth}
  \includegraphics[width=.99\columnwidth, height=22mm]{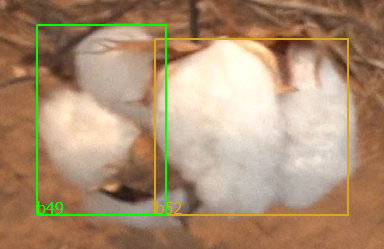}
  \caption{Similarity score: 0.016}
\end{subfigure}
\begin{subfigure}[b]{0.24\textwidth}
  \includegraphics[width=.99\columnwidth, height=22mm]{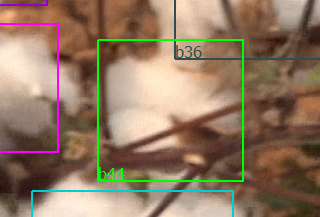}
  \caption{Similarity score: 0.011}
\end{subfigure}
\begin{subfigure}[b]{0.24\textwidth}
  \includegraphics[width=.99\columnwidth, height=22mm]{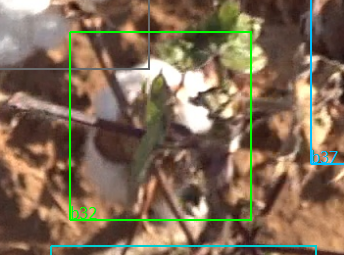}
  \caption{Similarity score: 0.021}
\end{subfigure}
\caption{NTrack re-identification results. Each column shows the same set of
cotton bolls. The bottom row shows the bolls after being occluded for several
frames. The bounding box colors represent the \textit{true} identity of the
objects and the IDs (e.g., b25, b49, etc.) at the bottom left corner of the
bounding boxes are \textit{assigned} by NTtrack. To highlight the poor
performance of visual similarity, each column is accompanied with a cosine
similarity score calculated between the \textit{green} bounding boxes. The
correspondences between the bounding box colors and IDs illustrates that NTrack
can successfully reidentify cotton bolls without visual similarity.} 
\label{fig:reindentification}
\end{figure*}

We also qualitatively evaluated NTrack by way of visual appearance-based
re-identification of cotton bolls across a video sequence. Examples where
appearance-based re-identification fails are shown in
Fig.~\ref{fig:reindentification}. The similarity scores in
Fig.~\ref{fig:reindentification} are calculated using the cosine similarity
between a pair of appearance descriptors ($r_i, r_j$) \cite{wojke2017simple}.
Concretely, 
\begin{equation} 
  \text{cosine similarity}(r_i,r_j) = \frac{r_i^\top r_j}{||r_i||\cdot||r_j||}, 
  \label{eq:cosine_similarity}
\end{equation} 
where $||r_i|| = ||r_j|| = 1$. \eqref{eq:cosine_similarity} outputs a real
number in the range $[0,1]$. Identical cotton bolls have a cosine similarity
score of 1, while distinct bolls have a score of 0.

In the \textbf{TexCot22} dataset, we hypothesize that all tracking methods that
depend on visual appearance to reidentify objects after reappearing will fail
to maintain true identity. This is based on the observation that
re-identification techniques exploited by many popular tracking methods (e.g.,
\cite{wojke2018deep}) tend to differentiate objects using color and shape as
distinctive features. However, individual cotton bolls have a homogeneous color
and shape distribution. Furthermore, the appearance of a cotton boll changes
drastically due to the shift in perspective after being occluded for a few
frames. Even for an experienced human, it is challenging to reidentify these
reappearing cotton bolls. Nevertheless, NTrack can successfully reidentify the
bolls (i.e., assign the same ID).

\subsection{Ablation Study}
\begin{table}
\centering
\begin{adjustbox}{max width=\linewidth}
\begin{tabular}{|l|p{1cm}p{1cm}p{1cm}|c|c|c|}
\hline
\textbf{Method} & \textbf{Linear\newline motion} & \textbf{Dynamic \newline motion} & \textbf{RLA}
& \textbf{IDF$_1$\textcolor{teal}{$\uparrow$}} & \textbf{MOTA\textcolor{teal}{$\uparrow$}} 
& \textbf{IDsw\textcolor{teal}{$\downarrow$}} \\\hline
Baseline        & \checkmark &            &            & 90.3  & 88.9 & 3682 \\ \hline
NTrack\_motion  &            & \checkmark &            & 92.2  & 89.3 & 264 \\ \hline
NTrack\_N1      &            & \checkmark & \checkmark & 92.5  & 89.5 & 101 \\ \hline
NTrack\_N3      &            & \checkmark & \checkmark & 92.8  & 89.5 & 85  \\ \hline
NTrack\_N5      &            & \checkmark & \checkmark & 92.8  & 89.4 & 85  \\ \hline
\end{tabular}%
\end{adjustbox}
\caption{A comparison of the impact of integrating different modules into NTrack
against a ByteTrack \cite{zhang2022bytetrack} baseline. NTrack\_N1, NTrack\_N3,
NTrack\_N5 model use 1, 3, and 5 neighbors respectively, in the RLA module.}
\label{tab:ablation_study}
\end{table}

To analyze our system design choices, we decoupled and validated the
performance impact of each of NTrack's modules. Table~\ref{tab:ablation_study}
shows the contributions of the various modules when compared against a
ByteTrack \cite{zhang2022bytetrack} baseline. The overall performance of NTrack
gradually improved upon integrating each module. The baseline system uses a
Kalman filter for motion prediction and an effective association method to
reidentify objects. When compared to the baseline, our dynamic motion model
improved the IDF$_1$ score by $2\%$. The combined model,
\textit{NTrack\_motion} (baseline + dynamic motion model), also reduced the
counting error by a large amount. This supports the hypothesis that for small
objects (e.g., fruits, flowers, etc.) that move irregularly in the wild, a
dynamic motion model is preferable for tracking.

We designed the RLA module specifically for identifying occluded cotton bolls.
Our experiments show that the RLA module serves its purpose very well. In
particular, NTrack achieved the highest scores on the IDF$_1$, MOTA, and IDsw
metrics when the RLA module was combined with the dynamic motion model.
Although there was only a 0.6\% gain in the IDF$_1$ score due to the addition
of the RLA module, the improvement is significant since there are few occluded
bolls in a video sequence when compared to the total number of bolls. The
number of neighbors (e.g, 1, 3, or 5) in the RLA module was empirically
selected for these experiments.

\section{Conclusion}
\label{sec:conclusion}
In this paper we described NTrack, a relative location-based MOT system that
enables accurate tracking of cotton bolls in outdoor field environments. NTrack
is able to robustly maintain object identity and it can reidentify objects
after long periods of occlusion, which is a common scenario in agricultural
applications. What's more, we introduced \textbf{TexCot22}, the \textit{first}
infield cotton boll video dataset. Using this dataset, NTrack was evaluated
against other contemporary MOT techniques and shown to significantly outperform
all of them in identity preserving metrics. In future work, we will extend our
framework to incorporate multiple view tracking data in order to further
improve the automated counting of infield cotton bolls.

\section*{Acknowledgments}
The data collected for the research results reported within this paper was done
in collaboration with Texas Tech University and the Texas A\&M AgriLife
Research and Extension Center's Halfway Station. The authors acknowledge the
Texas Advanced Computing Center (TACC) at The University of Texas at Austin for
providing software, computational, and storage resources that have contributed
to these results.

\bibliographystyle{IEEEtran}
\bibliography{IEEEabrv,ntrack}

\vspace{-4mm}
\begin{IEEEbiography}[{\includegraphics[width=25mm,height=32mm,clip]{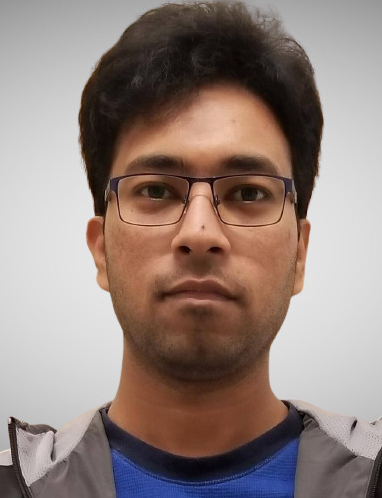}}]
{Md Ahmed Al Muzaddid} received the B.S. degree in computer science and
engineering from the Bangladesh University of Engineering and Technology. He is
currently pursing the Ph.D. degree with the Department of Computer Science and
Engineering, The University of Arlington at Texas. His research interests
include computer vision, machine learning, and agricultural automation.
\end{IEEEbiography}

\begin{IEEEbiography}[{\includegraphics[width=25mm,height=32mm,clip]{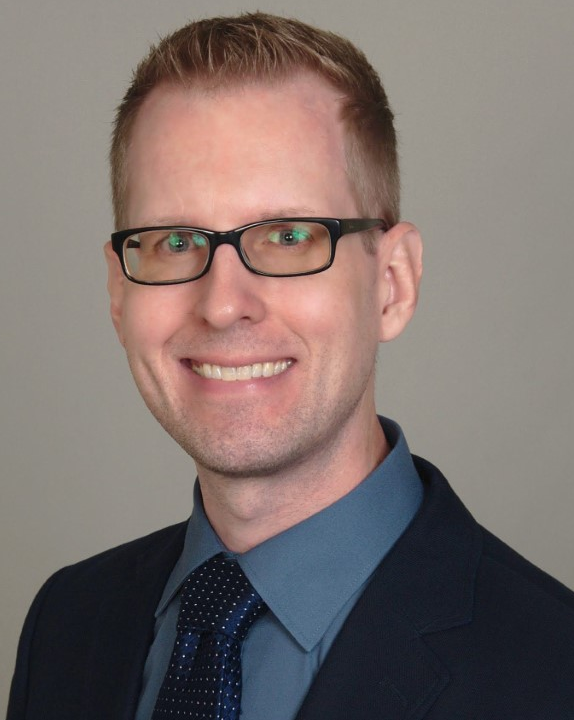}}]
{William J. Beksi} (Member, IEEE) received the B.S. degree in mathematics and
computer science from Stevens Institute of Technology in 2002, and the M.S. and
Ph.D. degrees from the University of Minnesota in 2016 and 2018, respectively.
He is currently an Assistant Professor in the Department of Computer Science
and Engineering at The University of Arlington at Texas, where he also leads
the Robotic Vision Laboratory. His research interests include robot perception,
human-robot interaction, and autonomous systems.
\end{IEEEbiography}
\vfill

\end{document}

%% file: figures/model.tikz
\usetikzlibrary{positioning}
\usetikzlibrary{chains,fit,shapes,calc}
\usetikzlibrary{decorations.text}

\definecolor{myblue}{RGB}{80,80,160}
\definecolor{mygreen}{RGB}{80,160,80}
\definecolor{myyellow}{RGB}{255,160,80}
\definecolor{mylime}{RGB}{199,234,70}
\definecolor{rlacolor}{RGB}{210,235,230}
\definecolor{labelcol}{RGB}{107,107,107}
\begin{tikzpicture}[thick,
  fsnode/.style={fill=myblue, inner sep=0pt,minimum size=4pt},
  ssnode/.style={fill=mygreen, inner sep=0pt,minimum size=4pt},
  every fit/.style={rectangle,draw,inner sep=2pt,text width=2cm},
 shorten >= 3pt,shorten <= 3pt
]
    \draw [rounded corners=10pt, thick, draw=gray, fill=orange, opacity=0.1](0,0) rectangle (12,4);
    \node (inpft1) at (.5, 3) [ thick] {$F_{l-1}$};
    \node (inpft) at (.5, 1) [ thick] {$F_{l}$};

    \node(optflow)[draw, shift={(1.2,0)}, align=center, minimum height= 1cm] at (inpft1.east) {\textbf{Optical}\\ \textbf{Flow}};
    \foreach \x in {0}
        \node(pf_pred)[draw, shift={(1.2,0)},align=center, minimum height= .5cm, fill = gray!30,  opacity=.9] at ($(optflow.east)+(.1*\x,-.1*\x)$) {\textbf{Predict}};
    \node(detector)[draw, shift={(1.5,0)}, minimum height= 1cm] at (inpft.east) {\textbf{Detector}};

    \node(asso)[draw, label=below:\textbf{Association}, shift={(1.2 ,-1)}, minimum height= 3cm, minimum width = 1.5cm] at (pf_pred.east) {\textbf{}};
    \begin{scope}[start chain=going right,node distance=1mm, shift={($(asso.west|-pf_pred.east)+(.24,.3)$)}]
    \foreach \i/\col in {1/mygreen, 2/mygreen, 3/mygreen, 4/myyellow, 5/myyellow}
      \node[ssnode,on chain, \col] (f\i)  {};
    \end{scope}
    \begin{scope}[start chain=going right,node distance=1mm, shift={($(asso.west|-detector.east)+(.35,-.3)$)}]
    \foreach \i/\col in {1/mygreen, 2/mygreen, 3/mygreen, 4/mylime}
      \node[fsnode,on chain, \col] (s\i)  {};
    \end{scope}
    \draw[thin, gray!80, dashed] ($(f1.south)+(0,.05)$) -- ($(s2.north)+(0,-.05)$);
    \draw[thin, gray!80, dashed] ($(f2.south)+(0,.05)$) -- ($(s3.north)+(0,-.05)$);;
    \draw[thin, gray!80, dashed] ($(f3.south)+(0,.05)$) -- ($(s1.north)+(0,-.05)$);;

    \node(pf_updt1)[draw, shift={(1.4,0)}, align=center,fill = gray!30,  opacity=.9, minimum height= .5cm] at (asso.east|-f1) {\textbf{Update}};

    \node(rla)[draw, fill=rlacolor, opacity=.4,circle,  minimum height= 1.7cm] at (asso.east-|pf_updt1) {}; 
    \node[ fill=mygreen, circle, shift={(.3,.2)}, inner sep=0pt,minimum size=2pt] (nn1)  at (rla.west) {};
    \node[fill=mygreen, circle, shift={(1.2,.45)}, inner sep=0pt,minimum size=2pt] (nn2)  at (rla.west) {};
    \node[fill=mygreen, circle, shift={(.8,-.6)}, inner sep=0pt,minimum size=2pt] (nn3)  at (rla.west) {};
    \node[fill=myyellow, circle, shift={(.6,-.1)}, inner sep=0pt,minimum size=2pt] (d1)  at (rla.west) {};
    \node[fill=myyellow, circle, shift={(.8,.6)}, inner sep=0pt,minimum size=2pt] (d2)  at (rla.west) {};
    \draw [gray,thin, ->] ($(nn1)+(-.03,.03)$) -- (d1.center);
    \draw [gray, thin, ->] ($(nn3)+(.02,-.03)$) -- (d1.center);
    \draw [gray,thin, ->] ($(nn1)+(-.02,-.01)$) -- (d2.center);
    \draw [gray, thin, ->] ($(nn2)+(.02,-.02)$) -- (d2.center);

    \node(pf_updt2)[draw, align=center, shift={(1.2,0)},fill = gray!30,  opacity=.9, minimum height= .5cm] at (rla.east) {\textbf{Update}};
    \node(init)[draw, minimum height= .5cm]  at (rla|-s1) {\textbf{Initialize}}; %
    \node(ref)[draw, shift={(.7,0)}, align=center, minimum height= 3cm] at (pf_updt2.east) {\rotatebox{90}{\textbf{Refiner}}};

    \draw [->] (inpft1.east) -- (optflow.west|-inpft1.east);
    \draw [->] (inpft.east) -- ([xshift=.4cm]inpft.east) -- ([xshift=.4cm]inpft.east |-  optflow.base west)--(optflow.base west) ;

    \draw[->] (optflow.east) -- (pf_pred.west);
    \draw[->] (pf_pred.east) --node [above,midway, labelcol] {$\mathcal{T}$} (pf_pred-|asso.west);
    \draw[->] (inpft.east) -- (detector) --node [above,midway, labelcol] {$\mathcal{D}^{f_l}$} (asso.west|-detector);

    \draw[->, mygreen] (asso.east|-f1) -- node [above,midway,labelcol] {$\mathcal{T}_a$} (pf_updt1) ;
    \draw[->, myyellow] (asso.east) --  node [above,midway, labelcol] {$\mathcal{T}_d$} (rla) ;
    \draw[->, mylime] (asso.east|-s1) -- node [above,midway, labelcol] {$\mathcal{D}_u$} (init) ;

    \draw[->, myyellow] (rla.east) -- (pf_updt2.west) ;
    \draw[->, mygreen] (pf_updt1.east|-f1) -- (ref.west|-pf_updt1.east) ;
    \draw[->, myyellow] (pf_updt2) -- (ref) ;
    \draw[->, mylime] (init.east)--(init-|ref.west) ;

    \path[postaction={decoration={
                    text along path,
                    text format delimiters={|}{|},
                    text={|\bf|RLA},
                    text align=center,
                    reverse path},
                    decorate}]
        ([shift=(10:.9)]rla)   arc (10:50:.9); 


\end{tikzpicture}